\documentclass[10pt]{article} % For LaTeX2e
\usepackage[preprint]{tmlr}

\usepackage{hyperref}
\usepackage{url}

\title{Cyclic and Randomized Stepsizes\\ Invoke Heavier Tails in SGD \mgrev{than Constant Stepsize}}

\author{%
 \name Mert G\"{u}rb\"{u}zbalaban \email mg1366@rutgers.edu\\
 \addr Department of Management Science
 and Information Systems\\ Rutgers Business School, Piscataway, NJ, USA. \& \\Center for Statistics and Machine Learning\\Princeton University, Princeton, NJ, USA.
 \AND
 \name Yuanhan Hu* \email yuanhan.hu@rutgers.edu \\
 \addr Department of Management Science
 and Information Systems\\ Rutgers Business School, Piscataway, NJ, USA
  \AND
 \name Umut \c{S}im\c{s}ekli  \email umut.simsekli@inria.fr \\
 \addr Inria, CNRS, Ecole Normale Sup\'{e}rieure\\ 
PSL Research University, Paris, France
  \AND
 \name Lingjiong Zhu  \email zhu@math.fsu.edu \\
 \addr Department of Mathematics\\ Florida State University, Tallahassee, FL, USA.\\
 \\
 * Most of this work was conducted while Y.H. was a Ph.D. student at Rutgers University.
}

\def\month{08}  % Insert correct month for camera-ready version
\def\year{2023} % Insert correct year for camera-ready version
\def\openreview{\url{https://openreview.net/forum?id=lNB5EHx8uC}} % Insert correct link to OpenReview for camera-ready version

\usepackage{mathtools}
\usepackage[textsize=tiny]{todonotes}

\usepackage{amsmath,amsthm,amsfonts,amssymb}
\usepackage{subfigure}
\usepackage{multirow}
\usepackage{hhline}

\usepackage{todonotes}

\newtheorem{theorem}{Theorem}
\newtheorem{corollary}{Corollary}

\newtheorem{lemma}{Lemma}
\newtheorem{proposition}{Proposition}
\newtheorem{remark}{Remark}

\def\mgrev#1{\textcolor{black}{#1}}

\usepackage{bbding}

\begin{document}

\maketitle

\begin{abstract}
Cyclic and randomized stepsizes are widely used in the deep learning practice and can often outperform standard stepsize choices such as constant stepsize in SGD. Despite their empirical success, not much is currently known about when and why they can theoretically improve the generalization performance. We consider a general class of Markovian stepsizes for learning, which contain i.i.d. random stepsize, cyclic stepsize as well as the constant stepsize as special cases, %While studying the generalization itself as a function of the stepsize sequence appears to be a difficult problem, 
and motivated by the literature which shows that heaviness of the tails (measured by
the so-called ``tail-index”) in the SGD iterates is correlated with generalization, we study tail-index %as a proxy to generalization performance 
and provide a number of theoretical results that demonstrate how the tail-index varies on the stepsize scheduling. Our results bring a new understanding of the benefits of cyclic
and randomized stepsizes compared to constant stepsize in terms of the tail behavior. We illustrate our theory on linear regression experiments and show through deep learning experiments that Markovian stepsizes can achieve even a heavier tail and be a viable alternative to cyclic and i.i.d. randomized stepsize rules.%, leading to the best performance in some cases. 
\end{abstract}

\section{Introduction}
Stochastic optimization problems are ubiquitous in supervised learning. In particular, the learning problem in neural networks can be expressed as an instance of the stochastic optimization problem
\begin{equation} \min\nolimits_{x\in\mathbb{R}^{d}} \mathbb{E}_{z \sim \mathcal{D}} [f(x,z)],
\label{eq-risk-min}
\end{equation}
where $x\in \mathbb{R}^d$ are the parameters, $z\in\mathcal{Z}$ is the random data which is assumed to obey an unknown probability distribution $\mathcal{D}$, and $f$ is the loss of misprediction with parameters $x$ corresponding to data $z$. Since the population distribution $\mathcal{D}$ is unknown, in practice, we often approximately solve the empirical version of \eqref{eq-risk-min}: 
\begin{equation} \min\nolimits_{x\in\mathbb{R}^{d}} F(x):= \frac{1}{n}\sum\nolimits_{i=1}^n \mgrev{f_i(x), \quad\text{where}\quad f_i(x):= f(x,z_i)},
\label{eq-emp-risk}
\end{equation}
based on a dataset $(z_1, z_2, \dots, z_n) \in \mathcal{Z}^n$ which consists of independent identically distributed (i.i.d.) samples from data.
Stochastic gradient descent (SGD) methods are workhorse methods for solving such problems due to their scalability properties and their favorable performance in practice \citep{bottou2010large,bottou2012stochastic}. SGD with batch-size $b$ consists of the updates %\todo{maybe it's better define online SGD directly if we are not using offline SGD?}
\begin{equation} 
x_{k+1} = x_k - \eta_{k+1} \tilde \nabla f \left(x_k\right)\,,
\label{eq-stoc-grad}
\end{equation}  
where $\eta_{k+1}$ is the stepsize and $\tilde \nabla f(x_k) := \frac{1}{b}\sum_{i \in \Omega_{k}}\nabla f_i(x_k)$
is the stochastic gradient at $k$-th iterate
with $\Omega_{k}$ being a set of data points with cardinality $|\Omega_{k}|=b$ (see Sec.~\ref{sec-least-squares} for details). \looseness=-1
%When $b\ll n$, it is much cheaper to compute
%the stochastic gradient $\tilde \nabla f_{k}(x)$
%than the full gradient $f(x)$ given in \eqref{f:eqn}
%and we also notice that $\tilde \nabla f_{k}(x)$
%is a conditionally unbiased estimator
%of the full gradient $f(x)$, i.e. $\mathbb{E}[\tilde \nabla f_{k}(x)|x]=f(x)$.

Understanding the generalization behavior of SGD, i.e. how a solution found by SGD performs on unseen data has been a major question of research in the last decades (see e.g. \cite{keskar2017improving,hardt2016train,lei2020fine,lin2016generalization}). %There have been several approaches. 
Recent empirical and theoretical studies have revealed an interesting phenomenon in this direction. Heavy tails in fact can arise in SGD iterates due to multiplicative noise and the amount of heavy tails (measured by the so-called ``tail-index") is significantly correlated with the generalization performance in deep learning practice \citep{HTP2021, hodgkinson2020multiplicative}. This phenomenon is not specific to deep learning in the sense that it arises even in surprisingly simple settings such as linear regression (when the loss is a quadratic) with Gaussian input data. This raises the natural question of how the choice of the stepsize sequence $\{\eta_k\}$ affects the tail-index and generalization properties which will be the main topic of study in this work.

The stepsize sequence $\{\eta_k\}$ can be chosen in various ways in a deterministic or a randomized fashion. Constant stepsize as well as varying stepsizes are proposed in the literature \citep{robbins1951stochastic, kiefer1952stochastic,Smith2017,bottou2018optimization}. %{\color{red}More references and older references here}
For constant stepsize and for (least squares) linear regression problems where the loss $f$ is a quadratic, \cite{HTP2021} showed that the tails are monotonic with respect to the stepsize and the batch-size and there exists a range of stepsizes for which there exists a stationary distribution with an infinite variance. In a very recent work, \cite{gurbuzbalaban2022heavy} showed that even heavier tails can arise in the decentralized stochastic gradient descent with constant stepsize and this correlates with generalization. However, other (non-constant) choices of the stepsize can often outperform constant stepsize. \cite{musso2020stochastic} suggests that in the small learning rate regime, uniformly-distributed random learning rate yields better regularization without extra computational cost compared to constant stepsize. Also, cyclic stepsizes where the stepsize changes in a cyclic fashion (obeying lower and upper bounds) have been numerically demonstrated to be very useful for some problems. Among this line of work, \cite{Smith2017} argued that cyclic stepsizes require less tuning in deep learning and can lead to often better performance.  \cite{Smith2015} suggests that experiments with sinusoidal learning rates versus cyclic stepsize with equal grid points lead to similar results. \cite{smith2017exploring} demonstrates that the cyclical learning rate can produce greater testing accuracy than traditional training despite using large learning rates for different network structures and datasets. 
\cite{Huang2017} argues that cyclic stepsize allows to train better models compared to constant stepsize. \cite{Kalousek2017} studies a steepest descent method with random stepsizes and shows that it can achieve faster asymptotic rate than gradient descent without knowing the details of the Hessian information. \cite{ZhangCyclical2020} suggests cosine cyclical stepsize for Bayesian deep learning. \cite{gulde2020deep} investigates cyclical learning rate and proposes a method for defining a general cyclical learning rate for various deep reinforcement learning problems. They show that using cyclical learning rate achieves similar or even better results than highly tuned fixed learning rates. \cite{wang2022empirical} applies the cyclical learning rate to train transformer-based neural networks for neural machine translation. They show that the choice of optimizers and the associated cyclical learning rate policy can have a significant impact on performance. \cite{alyafi2018cyclical}  argues that with a cyclical learning rate, the neural networks will have better results with similar or even smaller number of epochs on unbalanced datasets without extra computational cost compared to constant stepsize. In addition, due to their popularity, cyclic stepsizes are also part of popular packages such as PyTorch.\footnote{See e.g. the ``CyclicLR package" available at the website \url{https://pytorch.org/docs/stable/generated/torch.optim.lr_scheduler.CyclicLR.html}.} 

Despite the empirical success of randomized stepsize and cyclical stepsize rules, to our knowledge, the effect of the stepsize scheduling (cyclic, randomized versus constant etc.) on the generalization performance is not well studied from a theoretical standpoint except numerical results that highlight the benefits of randomized and cyclic stepsize rules compared to a constant stepsize choice. While studying the generalization itself as a function of the stepsize sequence appears to be a difficult problem; building on the literature which demonstrates that tail-index is correlated with generalization (see e.g. \cite{simsekli2020hausdorff,barsbey2021heavy,csimcsekli2019heavy,HTP2021}), we study
tail-index as a proxy to generalization performance and demonstrate how the tail-index depends on the stepsize scheduling, highlighting the benefits of using cyclic and randomized stepsizes compared to constant stepsize in terms of the tail behavior. %\looseness=-1%Our contributions are as follows: 

\textbf{Contributions.} We propose a general class of Markovian stepsizes for learning, where the stepsize sequence $\eta_k$ evolves according to a finite-state Markov chain where transitions between the states happen with a certain probability $p\in [0,1]$. This class recovers %w
    i.i.d. random stepsize, cyclic stepsize as well as the constant stepsize as special cases. For losses that have quadratic growth outside a compact set in dimension one, one can infer from the results of \cite{mirek2011heavy} that SGD iterates with Markovian stepsizes will be heavy-tailed (see Thm.~\ref{thm:nonlinear:cyclic} and Thm. \ref{thm:nonlinear:Markovian}). However, the assumptions involved are highly nontrivial to check and do not hold for dimension two or higher even for linear regression.\footnote{In \cite{mirek2011heavy}, it is required that the Hessians of the component functions $f(x,z_i)$ have certain orthogonality properties which does not typically hold in linear regression or more generally in statistical learning.} Furthermore, these results do neither specify how the tail-index depends on quantities of interest such as the batch-size, dimension nor compare the tails with that of the constant stepsize. %This motivates us to study of least square problems (when $f$ is a quadratic) where more precise characterizations of the tail-index can be obtained from which practical insights to deep learning can be extracted \citep{HTP2021,raj-quadratic,raj-general,gurbuzbalaban2022heavy}. 
    Hence, to get finer characterizations of the tail-index and further intuition about the tails, we focus on linear regression as a case study throughout this paper and obtain explicit characterizations that provides further insights as well as theoretical support for the use of cyclic and randomized stepsizes in deep learning practice. We find that even in the linear regression setting, there are significant technical challenges. Here, 
    %This class is general recovering random stepsizes and cyclic stepsizes as special cases
    viewing SGD as an iterated random recursion, we make novel technical contributions in analyzing the iterated random recursions with cyclic and Markovian structure with finite state space. We borrow the idea of regeneration times from probability theory, and consider iterates at those regeneration times to utilize the hidden renewal structure of the underlying iterated random recursions (see Sec.~\ref{sec:main:results} for further details). As a result, we are able to study and characterize the tail-index of the SGD iterates with cyclic stepsizes (%Thm.~\ref{thm:nonlinear:cyclic},
    Thm.~\ref{thm:cyclic}), Markovian stepsizes (Thm.~\ref{thm:Markov:regeneration}) %and %Thm.~\ref{thm:nonlinear:Markovian} in the Appendix) 
    as well as i.i.d. random stepsizes (Thm.~\ref{thm:iid}). %(Thm.~\ref{thm:nonlinear:iid} and Thm.~\ref{thm:iid} in the Appendix).

For cyclic, Markovian and i.i.d. stepsizes, under Gaussian data assumptions for linear regression, we also give a sharp characterization of the tail-index and provide a formula for the range of stepsizes in which SGD iterates admit a stationary distribution with an infinite variance (Lem.~\ref{lem:alternative:expression:cyclic}, Prop.~\ref{alpha:2:cyclic}, Lem.~\ref{lem:alternative:expression:Markov}, Prop.~\ref{alpha:2:Markov}, Lem.~\ref{lem:alternative:expression}, Prop.~\ref{alpha:2} in the Appendix). We provide non-asymptotic bounds moment bounds for SGD iterates and provide non-asymptotic convergence rates to the stationary distribution in the Wasserstein metric (Thm.~\ref{thm:convergence:cyclic}, Thm.~\ref{thm:convergence:Markov:Gaussian}, Thm.~\ref{thm:convergence:iid} in the Appendix). These results are obtained by using various technical inequalities, synchronous coupling, and spherical symmetry of Gaussian distributions and are deferred to the Appendix due to space considerations. 

In Sec.~\ref{sec:comparison}, we compare the tail-indices among these stepsize rules theoretically for linear regression. We show that i.i.d. random stepsizes (where $\eta_k$ are i.i.d. with the same distribution as $\eta$) centered around an expected value $\hat{\eta}$ and a positive range $R$ has heavier tail compared to constant stepsize with $\hat{\eta}$ (Prop.~\ref{prop:iid:constant:comparison}) as well as cyclic stepsize centered around $\hat{\eta}$ on a uniform grid with the same range (Prop. \ref{prop:compare:iid:cyclic}). Our results show that coarser the grid for the stepsize or wider the stepsize range $R$, heavier the tails become in a sense that we will precise (Thm.~\ref{thm:mono:Gaussian:range}, Thm.~\ref{thm:mono:Markovian:range}, Thm.~\ref{thm:mono:Markovian:range:2}). We prove that reducing the batch-size or increasing the dimension also leads to heavier tails, provided that the tail-index is not too small (Thm.~\ref{thm:mono:Gaussian:cyclic}, Thm.~\ref{thm:mono:Gaussian:Markov:regeneration}, Thm.~\ref{thm:mono:Gaussian}). 
These results are obtained by employing delicate analysis 
leveraging convexity and Jensen's inequality.
When the transition probability $p \in (0,1/2)$ and there are two states, we completely work out the distribution of the regeneration time and the sample path of the Markov chain till the regeneration time to show that Markovian stepsizes can achieve even heavier tail with respect to cyclic and i.i.d. stepsizes and the tail gets heavier if $p$ gets smaller (Prop.~\ref{cor:comparison}). We also discuss in Appendix~\ref{subsec-appendix-markovian} how our comparisons can be extended to general state spaces. Since smaller tail-indices are correlated with better generalization in deep learning practice, our results shed further light into why random and cyclic stepsizes could perform better than constant stepsize in some deep learning settings. In Sec.~\ref{sec:numerical}, we illustrate our theory on linear regression experiments and show through deep learning experiments that our proposed Markovian stepsizes can be a viable alternative to other standard choices in the literature, leading to the best performance in some cases. 
\section{Technical Background and Literature Review}\label{sec-least-squares}
\textbf{Heavy-tailed power-law distributions.} A real-valued random variable $X$ is said to be \emph{heavy-tailed}
if the right tail or the left tail of the distribution  decays slower than any exponential distribution. 
A real-valued random variable $X$ is said to have heavy right tail
if $\lim_{x\rightarrow\infty}\mathbb{P}(X\geq x)e^{cx}=\infty$
for any $c>0$, and a real-valued random variable $X$ is said to have heavy left tail 
if $\lim_{x\rightarrow\infty}\mathbb{P}(X\leq-x)e^{c|x|}=\infty$
for any $c>0$. Similarly, an $\mathbb{R}^{d}$-valued random vector $X$
is said to have heavy tail if $u^{T}X$ has heavy right tail 
for some vector $u\in\mathbb{S}^{d-1}$, 
where $\mathbb{S}^{d-1}:=\{u\in\mathbb{R}^{d}:\Vert u\Vert=1\}$ is the unit sphere in $\mathbb{R}^{d}$.
%%%%%%%%%%%%%%%%%%%%%%%%%%%%%%%%%%%%%%%%%%%%%%%%%%%%%%%%%%%%%%%%%%%%%%
Heavy tail distributions include $\alpha$-stable distributions,
Pareto distribution, log-normal distribution and Weilbull distribution. 
An important class of the heavy-tailed distributions
is the distributions with \emph{power-law} decay, 
which is the focus of our paper. That is,
$\mathbb{P}(X\geq x)\sim c_{0}x^{-\alpha}$
as $x\rightarrow\infty$ for some $c_{0}>0$ and $\alpha>0$,
where $\alpha$ is known as the \emph{tail-index},
which determines the tail thickness of the distribution.
Similarly, the random vector $X$ is said to have power-law decay
with tail-index $\alpha$ if for some $u\in\mathbb{S}^{d-1}$,
we have $\mathbb{P}(u^{T}X\geq x)\sim \mgrev{c_{0}(u)}x^{-\alpha}$,
for some constant \mgrev{$c_{0}(u)$ that may depend on the direction $u$} and for some $\alpha>0$.

\noindent \textbf{Tail-index in SGD with constant stepsize.} 
It is possible to show that SGD iterates are heavy-tailed with polynomially-decaying tails admitting a unique stationary distribution, when the loss is strongly convex outside a compact set \citep{hodgkinson2020multiplicative} or when the loss has linear growth for large $x$ \citep{mirek2011heavy,HTP2021} (see Thm.~\ref{thm-mirek} in the Appendix for details), however in these results verifying the assumptions behind are highly non-trivial; %for such losses verifying the assumptions of the existing results are highly nontrivial.  
%However, much less is known about the tail behavior of the limiting distribution $x_\infty$ except when the map $\Psi_\Omega(x)$ has a linear growth for large $x$ \cite{HTP2021} or when the map is strongly convex outside a compact set \cn.  
and the dependence of the tail-index to the parameters of SGD such as the stepsize is not explicitly given. This motivates the study of least square problems (when $f$ is a quadratic) where more precise characterizations of the tail-index can be obtained %from which practical insights to deep learning can be extracted 
(see e.g. \citep{HTP2021,raj-quadratic,raj-general,gurbuzbalaban2022heavy}). \mgrev{It is also known that slow algorithms that can generalize well subject to heavier tails can be slower for optimization purposes.} \mgrev{For example, heavier tails in the gradients and iterates can often result in slower convergence rates in the training process in different optimization settings \cite{csimcsekli2019heavy,wang2021convergence}, where clipping the gradients and robust versions of stochastic gradient descent have been proposed for achieving faster convergence \cite{zhang2019gradient,gorbunov2022clipped}. Therefore, we believe understanding the heavy-tailedness of the iterates can be beneficial for understanding both optimization and generalization performance of an algorithm, and least square problems serve as a fundamental case where stronger results can be obtained due to their special structure.} \looseness=-1

Consider the least squares problem
%it is highly non-trivial to
%verify such assumptions in practice, and furthermore, the literature does not provide any rigorous connections
$\min_{x\in\mathbb{R}^d} \frac{1}{2n}\|Ax -y \|^2$,
where $A$ is an $n\times d$ matrix, with i.i.d. entries, $y \in \mathbb{R}^n$ and $x\in \mathbb{R}^d$. This problem can be written in the form of \eqref{eq-emp-risk} as
\begin{equation}\label{f:eqn} 
F(x) = \frac{1}{n}\sum\nolimits_{i=1}^n f_i(x), \quad\text{where}\quad f_i(x):= \frac{1}{2}\left(a_i^Tx - y_i\right)^2\,,
\end{equation}
where $a_i^T$ is the $i$-th row of the data matrix $A$. Using the fact that 
$\nabla f_i(x)  = a_i\left(a_i^T x - y_i\right)$, 
we can rewrite SGD iterations \eqref{eq-stoc-grad} as 
\begin{equation} 
x_{k+1} = \left(I - (\eta_{k+1}/b) H_{k+1}\right) x_k  + q_{k+1}\,,
\label{eq-stoc-grad-2}
\end{equation}  
where 
$H_k := \sum_{i\in \Omega_{k} } a_i a_i^T$ and $q_k := \frac{\eta_{k}}{b} \sum_{i\in \Omega_{k} } y_i$ with $\Omega_k := \{b(k-1)+1, b(k-1)+2, \dots, bk\}$ and $|\Omega_k| = b$. Here, for simplicity, throughout the paper, we assume that we are in the streaming regime (also called the one-pass setting \citep{frostig2015competing,jain2017accelerating,gao2022global}) where each sample is used once and is not recycled. We also make the following assumptions on the data throughout the paper:
\begin{itemize}
    \item [\textbf{(A1)}] 
    $a_i$'s are i.i.d. with a continuous density supported on $\mathbb{R}^d$ with all the moments being finite. 
    \item [\textbf{(A2)}] $y_i$ are i.i.d. with a continuous density whose support is $\mathbb{R}$ with all the moments finite. 
\end{itemize}
Assumptions $\textbf{(A1)}$ and $\textbf{(A2)}$ are satisfied in a large variety of cases, for instance when $a_{i}$ and $y_i$ are Gaussian distributed. When the stepsize $\eta_{k}\equiv\eta$ is constant, and $H_{k}$ are i.i.d., $x_{k}$ converges
to $x_{\infty}$, where $x_{\infty}$ has a heavy tail distribution with a tail-index
that can be characterized \citep{HTP2021}. However, to the best of our knowledge, cyclic stepsizes or Markovian stepsizes are not studied in the literature in terms of the tail behavior that they result in the SGD iterates \mgrev{which will be the subject of this work. We also note that decaying stepsize rules where $\eta_k \to 0$ are also widely used and they can ensure convergence of the SGD iterates \cite{sebbouh2021almost,gower2019sgd}. However, in this case, the iterates often converge to a limit point almost surely and therefore the stationary distribution of the iterates will typically be degenerate as a Dirac mass where characterizing the tail-index of the iterates will no longer be meaningful at stationarity. Therefore, in our analysis, we will assume that the stepsize is bounded away from zero, often lying on a finite grid. }
%froThat being said, comparing decaying deterministic stepsizes with decaying Markovian stepsizes at finite time in terms of the nature of limit points they converge to would also be an interesting problem left as future work.
%The latter result (Theorem~\ref{thm:main}) is given in the Appendix for the sake of completeness.
\looseness=-1 %However, such arguments do not directly work other stepsizes such as random stepsizes or cyclic stepsizes.
%%%%%%%%%%%%%%%%%%%%%%%%%%%%%%%%%%%%%%%%%%%%%%%%%%%%%%%%%%%%%%%%%%%%%%%
% \begin{theorem}[Theorem~2 in \cite{HTP2021}]\label{thm:main}
% Consider the SGD iterations (\ref{eq-stoc-grad-2}).
% If $\rho_{c}<0$ and there exists a unique positive $\alpha_{c}$ such that $h_{c}(\alpha_{c})=1$, 
% where $h_{c}$ and $\rho_{c}$ are defined in \eqref{def-hs} and \eqref{def-rho},
% then \eqref{eq-stoc-grad-2} admits a unique stationary solution $x_{\infty}$
% and the SGD iterations converge to $x_{\infty}$ in distribution,
% where the distribution of $x_{\infty}$ satisfies
% \begin{equation} 
% \lim\nolimits_{t\to\infty} t^{\alpha_{c}} \mathbb{P}\left(u^T x_{\infty} > t \right)= e_{\alpha_{c}}(u)\,, \quad u\in\mathbb{S}^{d-1}\,,\label{eq-heavy-tail}
% \end{equation}
% for some positive and continuous function $e_\alpha$ on 
% $\mathbb{S}^{d-1}$. 
% \end{theorem}
%%%%%%%%%%%%%%%%%%%%%%%%%%%%%%%%%%%%%%%%%%%%%%%%%%%%%%%%%%%%%%%%%%%%%% 
%In general, the tail-index $\alpha_{c}$ does not have a simple formula
%since $h_{c}(s)$ function lacks a simple expression. 
%A lower bound $\hat{\alpha}_{c}\leq\alpha_{c}$ holds
%where $\hat{\alpha}_{c}$ is the unique positive
%solution to $\hat{h}_{c}\left(\hat{\alpha}_{c}\right)=1$, 
%where $\hat{h}_{c}(s):=\mathbb{E}\left[\left\Vert I-\frac{\eta}{b}H_{1}\right\Vert^{s}\right]$, 
%provided that $\hat{\rho}_{c}:=\mathbb{E}\log\left\Vert I-\frac{\eta}{b}H_{1}\right\Vert<0$.

%%%%%%%%%%%%%%%%%%%%%%%%%%%%%%%%%%

%%%%%%%%%%%%%%%%%%%%%%%%%%%%%%%%%%%%%%%%%%%%%%%%%%%%%%%%%%
%\vspace{-0.1in}
\section{Main Results}\label{sec:main:results}

%%%%%%%%%%%%%%%%%%%%%%%%%%%%%%%%%%%%%%%%%%%%%%%%%%%%%%%%%%
\textbf{Stochastic Gradient Descent with Markovian Stepsizes.}
We consider Markovian stepsizes with finite state spaces.
The discussions with general state spaces will be provided in Section~\ref{sec:general:state:Markov} in the Appendix. 
%%%%%%%%%%%%%%%%%%%%%%%%%%%%%%%%%%%%%%%%%%%%%%%%%%%%%%%%%%%%%%%%
Let us consider the finite state space \begin{equation}\label{state:space:m}
{\color{black}\{\eta_{1},\eta_{2},\ldots,\eta_{m},\eta_{m+1}\}
=\{c_{1},c_{2},\ldots,c_{K-1},c_{K},c_{K-1},\ldots,c_{2},c_{1}\},}
\end{equation}
where $m=2K-2$ \mgrev{and $(c_1, c_2, \dots, c_K)$ is the stepsize grid.} \mgrev{We assume} the stepsize goes from $\eta_{1}$ to $\eta_{2}$ with probability $1$
and it goes from $\eta_{K}$ to $\eta_{K-1}$ with probability $1$. 
In between, for any $i=2,3,\ldots,K-1,K+1,\ldots,m$,
the stepsize goes from $\eta_{i}$ to $\eta_{i+1}$ with probability $p$
and from $\eta_{i}$ to $\eta_{i-1}$ with probability $1-p$
with the understanding that $\eta_{m+1}:=\eta_{1}$.
Therefore, $p=1$ reduces to the case of cyclic stepsizes.
\mgrev{We assume $c_i$ is not the same for every $i$}; %$\eta_{i}$ is not a constant; 
otherwise \mgrev{this setting} reduces to the case of constant stepsizes.

%%%%%%%%%%%%%%%%%%%%%%%%%%%%%%%%%%%%%%%%%%%%%%%%%%%%%%%

We first observe that SGD  \eqref{eq-stoc-grad} is an iterated random recursion of the form
\begin{equation}\label{Markovian:nonlinear} 
x_k = \Psi_{k}(x_{k-1},\eta_{k}),
\end{equation} 
where the map $\Psi_{k}:\mathbb{R}^d\times\mathbb{R}_{+} \to \mathbb{R}^d$, {\color{black}hides the dependence on  $\Omega_k$ which are random and i.i.d.} 
and $\eta_{k}$ are Markovian with the finite state space \eqref{state:space:m}.
To the best of our knowledge, there is no general stochastic linear recursion theory
for Markovian coefficients, except for some special cases, e.g. with heavy-tailed coefficients \citep{hay2011multivariate}. However, we will show that in the case of finite-state space,
it is possible to use the idea of regeneration times from probability theory to analyze 
the iterated random recursion.
The key idea is to introduce the regeneration times $r_{k}$, 
which are defined as $r_{0}=0$ and for any $k\geq 1$:
\begin{equation}\label{defn:regeneration:time}
r_{k}:=\inf\left\{j>r_{k-1}:\eta_{j}=\eta_{0}\right\}.
\end{equation}
That is, $r_{k}$ are the random times that the stepsizes start at $\eta_{0}$ at $k=0$. 
It is easy to see that $\{r_{k}-r_{k-1}\}_{k\in\mathbb{N}}$ are i.i.d. with the same distribution as $r_{1}$.
%%%%%%%%%%%%%%%%%%%%%%%%%%%%%%%%%%%%%%%%%%%%%%%%%%
{\color{black}By iterating \eqref{Markovian:nonlinear}, we have
\begin{equation}\label{Markovian:dynamics:nonlinear}
x_{r_{k+1}}=\Psi^{(r)}_{k+1}(x_{r_{k}}),
\end{equation}
with
$\Psi^{(r)}_{k+1}(\cdot):=\Psi_{r_{k+1}}\left(\cdots\Psi_{r_{k}+2}(\Psi_{r_{k}+1}(\cdot,\eta_{r_{k}+1}),\eta_{r_{k}+2}),\ldots,\eta_{r_{k+1}}\right)$, where the superscript $(r)$ refers to the random choice of stepsizes and it also happens to be the first letter of ``regeneration''.}
Since $\Omega_{k}$ are i.i.d. over $k$, by the definition of the regeneration time, 
it follows that $\Psi^{(r)}_{k}$ are i.i.d. over $k$.
We denote that $\Psi_{k}^{(r)}$ has the common distribution as $\Psi^{(r)}$.
If we assume that the random map $\Psi^{(r)}$ is Lipschitz on average, i.e.  $\mathbb{E}[L^{(r)}] < \infty$ with $L^{(r)} := \sup\nolimits_{x,y\in\mathbb{R}^d} \frac{\|\Psi^{(r)}(x) - \Psi^{(r)}(y) \|}{\|x-y\|}$, and is mean-contractive, i.e. if $\mathbb{E}\log(L^{(r)})<0$ then it can be shown under further technical assumptions that the distribution of the iterates converges to a unique stationary distribution $x_\infty$ geometrically fast \citep{diaconis1999iterated}. 

The following result can be obtained in 
dimension $d=1$, which follows directly from \citet{mirek2011heavy} by adapting it to our setting (see also \citet{buraczewski2016stochastic}). \citet{mirek2011heavy} also considers higher dimensions; but for $d>1$, the assumptions required are not satisfied for quadratic losses nor for least square problems.  %and we refer the readers to \citet{mirek2011heavy} for general $d$).

\begin{theorem}[Adaptation of \citet{mirek2011heavy} ]\label{thm:nonlinear:Markovian}
Assume stationary solution to \eqref{cyclic:dynamics:nonlinear} exists, \mgrev{$d=1$}, and: %. We further assume 
\begin{itemize}
    \item [(i)] There exists a random \mgrev{variable} %matrix 
$M^{(r)}$ and a random variable $B^{(r)}>0$ such that a.s. 
$|\Psi^{(r)}(x)-M^{(r)} x | \leq B^{(r)}$ for every $x$ \mgrev{where $|\cdot|$ denotes the absolute value};
    \item [(ii)] The conditional law of $\log |M^{(r)}|$ given $M^{(r)}\neq 0$ is non-arithmetic; i.e. \mgrev{its support is not equal to $a\mathbb{Z}$ for any scalar $a$ where $\mathbb{Z}$ is the set of integers};
    \item 
[(iii)] There exists $\alpha^{(r)}>0$ such that $\mathbb{E} [|M^{(r)}|^{\alpha^{(r)}}] = 1$,  $\mathbb{E} [|B^{(r)}|^{\alpha^{(r)}}] <\infty$ and 
$\mathbb{E}[|M^{(r)}|^{\alpha^{(r)}} \log^+ |M^{(r)}|]< \infty$,
where $\log^+(x) := \max(\log(x),0)$.
\end{itemize}
Then, there exists some constant $c_0^{(r)}>0$ such that 
$\lim_{t \to \infty } t^{\alpha^{(r)}} \mathbb{P}(|x_\infty|>t) = c_0^{(r)}.$  
\end{theorem}

 For linear regression, $\Psi^{(r)}$ is a composition of affine maps and stays affine (see our discussion in the proof of Thm.~\ref{thm:Markov:regeneration}); therefore Thm.~\ref{thm:nonlinear:Markovian} is applicable \mgrev{in dimension one}. More generally, Thm.~\ref{thm:nonlinear:Markovian} can be applicable to the restricted class of smooth losses that can be non-convex on a compact while having a quadratic structure outside the compact (so that the gradient is affine up to a constant), and says that heavy tails arises in SGD in this setting but does not precise the tail-index $\alpha^{(r)}$ and this result works only in dimension $d=1$. 
This motivates the study of more structured losses in high dimensional settings where more insights can be obtained. When the objective is a quadratic, we will provide a more detailed analysis.
%%%%%%%%%%%%%%%%%%%%%%%%%%%%%%%%%%%%%%%%%

\mgrev{For Markovian stepsizes, we observe that the SGD iterates are given by \eqref{eq-stoc-grad-2}} where 
% \begin{equation}\label{eq-stoc-grad-2-Markov} 
% x_{k+1} = \left(I - (\eta_{k+1}/b) H_{k+1}\right) x_k  + q_{k+1}\,.
% \end{equation} 
% where $H_{k}:= \sum_{i\in \Omega_{k} } a_i a_i^T$ 
% $q_k := \frac{\eta_{k}}{b} \sum_{i\in \Omega_{k} } y_i$ such that
$(H_{k},q_{k})$ is an i.i.d. sequence
and $\eta_{k}$ is a stationary Markov chain with 
finite state space independent of $(H_{k},q_{k})_{k\in\mathbb{N}}$.
%%%%%%%%%%%%%%%%%%%%%%%%%%%%%%%%%%%%%%%%%%%%%%%%%%%%%%%%%%%%%%%%%
Next, we will show that one can fully characterize the tail-index
when the stepsizes follow a Markov chain with a finite state space
using a renewal argument based on regeneration times. 
%%%%%%%%%%%%%%%%%%%%%%%%%%%%%%%%%%%%%%%%%%%%%%%%%%%%%%%
Let us introduce
\begin{equation}
h^{(r)}(s) := \lim\nolimits_{k\to\infty}\left(\mathbb{E}\left\| M_k^{(r)} M_{k-1}^{(r)}\dots M_1^{(r)}\right\|^s\right)^{1/k}\,,
\label{def-hs-Markov}
\end{equation}
where $$M_{k+1}^{(r)}:=\left(I - (\eta_{r_{k+1}}/b) H_{r_{k+1}}\right)
\left(I - (\eta_{r_{k+1}-1}/b) H_{r_{k+1}-1}\right)\cdots \left(I - (\eta_{r_{k}+1}/b) H_{r_{k}+1}\right)$$ and $r_{k}$'s are regeneration times defined in \eqref{defn:regeneration:time}.
We also define $$\Pi_k^{(r)} := M_k^{(r)} M_{k-1}^{(r)}\dots M_1^{(r)},$$
and $$\rho^{(r)} := \lim\nolimits_{k\to\infty} (2k)^{-1} \log\left(\mbox{largest eigenvalue of }\left(\Pi_k^{(r)}\right)^T \left(\Pi_k^{(r)}\right)\right).$$
We have the following result that characterizes the tail-index for the SGD with Markovian stepsizes.

%%%%%%%%%%%%%%%%%%%%%%%%%%%%%%%%%%%%%%%%%%%%%%%%%%%%%%%%%%%%%%%%
%In the Appendix, we provide the convergence results for the SGD iterations with Markovian stepsizes (\ref{eq-stoc-grad-2-Markov}) in Theorem~\ref{thm:Markov:regeneration} using a renewal argument based on regeneration times such that $x_{r_{k}}$ converges to $x_{\infty}$ which has a heavy tail distribution
%whose tail-index $\alpha^{(r)}$ can be characterized as the unique positive value such that $h^{(r)}\left(\alpha^{(r)}\right)=1$.
%%%%%%%%%%%%%%%%%%%%%%%%%%%%%%%%%%%%%%%%%%%%%%%%%%%%%%%%%%%%%

\begin{theorem}\label{thm:Markov:regeneration}
Consider the SGD iterations \eqref{eq-stoc-grad-2} with Markovian stepsizes \mgrev{in a finite-state space.} %\mgrev{where $(H_{k},q_{k})$ is an i.i.d. sequence
%and $\eta_{k}$ is a stationary Markov chain with 
%finite state space independent of $(H_{k},q_{k})_{k\in\mathbb{N}}$}. %(\ref{eq-stoc-grad-2-Markov}).
If $\rho^{(r)}<0$ and there exists a unique positive $\alpha^{(r)}$ such that $h^{(r)}\left(\alpha^{(r)}\right)=1$,
then 
\mgrev{\eqref{eq-stoc-grad-2}} admits a unique stationary solution $x_{\infty}^{(r)}$
and the SGD iterations with Markovian stepsizes converge to $x_{\infty}^{(r)}$ in distribution,
where the distribution of $x_{\infty}^{(r)}$ satisfies
$\lim\nolimits_{t\to\infty} t^{\alpha^{(r)}} \mathbb{P}(u^T x_{\infty}^{(r)} > t )= e_{\alpha^{(r)}}(u)$, for any $u\in\mathbb{S}^{d-1}$,
for some positive and continuous function $e_{\alpha^{(r)}}$ on 
$\mathbb{S}^{d-1}$.
\end{theorem}

\textbf{Proof.}
We recall from \mgrev{\eqref{eq-stoc-grad-2}} that the SGD iterates are given by
$x_{k+1} = \left(I - \frac{\eta_{k+1}}{b} H_{k+1}\right) x_k  + q_{k+1}$,
where $(H_{k},q_{k})$ is an i.i.d. sequence
and $\eta_{k}$ is a stationary Markov chain with 
finite state space independent of $(H_{k},q_{k})_{k\in\mathbb{N}}$.
We recall from \eqref{defn:regeneration:time} the regeneration times $r_{k}$, 
such that $r_{0}=0$ and for any $k\geq 1$:
$r_{k}:=\inf\left\{j>r_{k-1}:\eta_{j}=\eta_{0}\right\}$.
That is $r_{k}$ are the random times that the stepsizes start at $\eta_{0}$ at $k=0$. 
It is easy to see that $\{r_{k}-r_{k-1}\}_{k\in\mathbb{N}}$ are i.i.d. with the same distribution as $r_{1}$.
It follows that
$x_{r_{k+1}}=M_{k+1}^{(r)}x_{r_{k}}+q_{k+1}^{(r)}$,
where $M_{k+1}^{(r)}$ and $q_{k+1}^{(r)}$ are defined as:
\begin{align*}
&M_{k+1}^{(r)}:=\left(I - (\eta_{r_{k+1}}/b) H_{r_{k+1}}\right)
\left(I - (\eta_{r_{k+1}-1}/b) H_{r_{k+1}-1}\right)\cdots \left(I - (\eta_{r_{k}+1}/b) H_{r_{k}+1}\right),
\\
&q_{k+1}^{(r)}:=\sum\nolimits_{i=r_{k}}^{r_{k+1}}
\left(I - (\eta_{r_{k+1}}/b) H_{r_{k+1}}\right)
\left(I - (\eta_{r_{k+1}-1}/b) H_{r_{k+1}-1}\right)\cdots \left(I - (\eta_{i+1}/b) H_{i+1}\right)q_{i}.
\end{align*}
Since $r_{k}$ are regeneration times, one can easily check that
$\left(M_{k}^{(r)},q_{k}^{(r)}\right)$ are i.i.d. in $k$.
%%%%%%%%%%%%%%%%%%%%%%%%%%%%%%%%%%%%%%%%%%%%%%%%%%%%%%%%%%%%%%%%
The rest of the proof follows from Theorem~4.4.15 in \cite{buraczewski2016stochastic} which goes back to Theorem~1.1 in \cite{alsmeyer2012tail} and Theorem~6 in \cite{kesten1973random}. 
See also \cite{goldie1991implicit,bdp2015}.
The proof is complete.
\hfill $\Box$
\mgrev{
\begin{remark} It is possible to extend some of our results given in Thm.~\ref{thm:Markov:regeneration} for linear regression to problems where the loss function is a convex quadratic up to an error term outside a compact set. Smoothed Lasso problems where the penalty term is a smooth version of the $\ell_1$ penalty would be an example. In this more general case, while it is not possible to provide a formula for the tail-index exactly, one can still provide lower and upper bounds on the tail-index. The details can be found in Appendix~\ref{appendix-non-quadratic-results}.
\end{remark}
}

%%%%%%%%%%%%%%%%%%%%%%%%%%%%%%%%%%%%%%%%%%%%%%%%%%%%%%%%%%%%%%%%%%%%%
In general, there is no simple explicit formula for $h^{(r)}(s)$ to evaluate $\alpha^{(r)}$. 
However, we can easily obtain the following bound
using the sub-multiplicativity of the norm of matrix products:
\begin{equation}
h^{(r)}(s)
\leq
\hat{h}^{(r)}(s)
:=\mathbb{E}\left[\prod\nolimits_{i=1}^{r_{1}}\left\Vert I-(\eta_{i}/b)H_{i}\right\Vert^{s}\right]
=\mathbb{E}\left[\prod\nolimits_{i=1}^{r_{1}}\mathbb{E}_{H}\left[\left\Vert I-(\eta_{i}/b)H\right\Vert^{s}\right]\right],
\end{equation}
where $\mathbb{E}_{H}$ denotes the expectation taken over $H_{i}$, which are i.i.d. distributed as $H$, \mgrev{and which are} independent of $(\eta_{k})_{k\in\mathbb{N}}$, \mgrev{where} $r_{1}$ is \mgrev{the regeneration time} defined in \eqref{defn:regeneration:time}.
We define the lower bound $\hat{\alpha}^{(r)}$ for the tail-index \mgrev{$\alpha^{(r)}$} as the unique
positive value such that $\hat{h}^{(r)}\left(\hat{\alpha}^{(r)}\right)=1$, provided that 
$\hat{\rho}^{(r)}:=\mathbb{E}\left[\sum_{i=1}^{r_{1}}\mathbb{E}_{H}\left[\log\left\Vert I-\frac{\eta_{i}}{b}H\right\Vert\right]\right]<0$.
%Similar to Thm.~\ref{thm:cyclic:index}, 
\mgrev{In the following,} we show that the lower bound $\hat{\alpha}^{(r)}$ for the tail-index is increasing in batch-size.\looseness=-1

\begin{theorem}\label{thm:Markov:regeneration:index}
$\hat{\alpha}^{(r)}$ is strictly increasing in batch-size $b$
provided that $\hat{\alpha}^{(r)}\geq 1$.
\end{theorem}
\mgrev{
Our proof of Theorem~\ref{thm:Markov:regeneration:index} is based on the fact that the function $h^{(r)}(s)$ is strictly decreasing in $b$ for $s\geq 1$ which follows from Jensen's inequality and convexity of the function $\|\cdot\|^s$ for $s\geq 1$. The function $\|\cdot\|^s$ is not convex for $s<1$; this is the reason why the condition $\hat{\alpha}^{(r)}\geq 1$ is needed in Theorem~\ref{thm:Markov:regeneration:index} within our analysis. Note also that} the tail-index $\alpha^{(r)}$ is the unique positive value 
such that $h^{(r)}\left(\alpha^{(r)}\right)=1$
provided that $\rho^{(r)}<0$. 
In general, there is no simple closed-form expression for $h^{(r)}(s)$ that is defined in \eqref{def-hs-Markov}.

However, when the input data $a_{i}$ are Gaussian, 
we are able to obtain more explicit expression for $h^{(r)}(s)$.

\begin{itemize}
    \item [\textbf{(A3)}] $a_{i}\sim\mathcal{N}(0,\sigma^{2}I_{d})$ are Gaussian distributed for every $i$.
\end{itemize}

Under \textbf{(A3)}, 
we can obtain a more explicit expression (see Lem.~\ref{lem:alternative:expression:Markov} and Lem.~\ref{lem:simplified:Markov:regeneration} in the Appendix) to characterize the tail-index $\alpha^{(r)}$.
Under \textbf{(A3)}, we obtain the following result
that shows the tail-index $\alpha^{(r)}$ is increasing in batch-size $b$
and decreasing in dimension $d$.

\begin{theorem}\label{thm:mono:Gaussian:Markov:regeneration}
Assume \textbf{(A3)} holds and $\rho^{(r)}<0$. Then:
(i) The tail-index $\alpha^{(r)}$ is strictly increasing in batch-size $b$
provided that $\alpha^{(r)}\geq 1$;
(ii) The tail-index $\alpha^{(r)}$ is strictly decreasing in dimension $d$.
\end{theorem}

%%%%%%%%%%%%%%%%%%%%%%%%%%%%%%%%%%%%%%%%%%%%%%%%%%%%%%%%%%
\textbf{Stochastic Gradient Descent with Cyclic Stepsizes.} 
{\color{black}As a special case of the Markovian stepsizes, we consider the stochastic gradient descent method with cyclic stepsizes.} More specifically, %by a slight abuse of the notation,\footnote{Here, we are abusing the notation in the sense that $\{\eta_j\}_{j\leq K}$ denote the grid points and the stepsizes simultaneously.}
\mgrev{we assume that $\eta_{k}$ takes values on a grid $(c_1, c_2, \dots, c_K)$ in a cyclic manner satisfying \eqref{state:space:m}} % $\eta_{1},\eta_{2},\ldots,\eta_{K}$, 
%so that 
%\mgrev{
%\begin{equation}
%(\eta_{1},\eta_{2},\ldots,\eta_{m},\eta_{m+1})
%=(c_{1},c_{2},\ldots,c_{K-1},c_{K},c_{K-1},c_{K-2},\ldots,c_{2},c_{1}),
%\end{equation}
%}
% \begin{equation}
% (\eta_{1},\eta_{2},\ldots,\eta_{m},\eta_{m+1})
% =(\eta_{1},\eta_{2},\ldots,\eta_{K-1},\eta_{K},\eta_{K-1},\eta_{K-2},\ldots,\eta_{2},\eta_{1}),
% \end{equation}
\mgrev{and} the length of the cycle is $m=2K-2$. %so that 
\mgrev{In other words} $\eta_{mk+i}=\eta_{i}$, $i=1,2,\ldots,m$,
for any $k=0,1,2,\ldots$. \mgrev{Note that we can view cyclic stepsizes as a special case of Markovian stepsizes with transition probability $p=1$}.
%%%%%%%%%%%%%%%%%%%%%%%%%%%%%%%%%%%%%%%%%%%

{\color{black}As a special case of the Markovian stepsizes, the SGD \eqref{eq-stoc-grad} can be iterated and we can consider:
\begin{equation}\label{cyclic:dynamics:nonlinear}
x_{(k+1)m}=\Psi^{(m)}_{k+1}(x_{km}),
\end{equation}
where
$\Psi^{(m)}_{k+1}(\cdot):=\Psi_{(k+1)m}\left(\cdots\Psi_{km+2}(\Psi_{km+1}(\cdot,\eta_{km+1}),\eta_{km+2}),\ldots,\eta_{(k+1)m}\right)$ are i.i.d. over $k$. \mgrev{This is basically the map that corresponds to consecutive $m$ iterations of SGD, which demonstrates an i.i.d. structure.} We denote $\Psi^{(m)}$ as the common distribution of $\Psi_{k}^{(m)}$ where the superscript $(m)$ highlights the dependence on the cycle length $m$.}
In this case, an analogue of Thm.~\ref{thm:nonlinear:Markovian} can be obtained in dimension one (see Thm.~\ref{thm:nonlinear:cyclic} in the Appendix); but as before in the rest of the discussion, we focus on the quadratic case to have finer results for the tail-index.
%%%%%%%%%%%%%%%%%%%%%%%%%%%%%%%%%%%%%%%%%%%%%%%%%%%%%%%%%%%%%%
We recall the \mgrev{SGD} iterates from \eqref{eq-stoc-grad-2}, \mgrev{where we consider the stepsize $\eta_{k}$ to be deterministic and cyclic with a cycle length $m$}.
%\begin{equation}\label{eq-stoc-grad-2-m} 
%$x_{k+1} = \left(I - (\eta_{k+1}/b)H_{k+1}\right) x_k  + q_{k+1}\,,
%\end{equation} 
%where $H_{k}:= \sum_{i\in \Omega_{k} } a_i a_i^T$ are i.i.d. Hessian matrices,
%$q_k := \frac{\eta_{k}}{b} \sum_{i\in \Omega_{k} } y_i$, 
%and $\eta_{k}$ are deterministic and cyclic.
%%%%%%%%%%%%%%%%%%%%%%%%%%%%%%%%%%%%%%%%%%%%%%%
Next, let us introduce
\begin{equation}
h^{(m)}(s) := \lim\nolimits_{k\to\infty}\left(\mathbb{E}\left\| M_k^{(m)} M_{k-1}^{(m)}\dots M_1^{(m)}\right\|^s\right)^{1/k}\,,
\label{def-hs-m}
\end{equation}
where
\begin{equation}\label{M:k:m:defn}
M_{k}^{(m)}:=\left(I - (\eta_{m}/b) H_{km}\right)
\left(I - (\eta_{m-1}/b) H_{km-1}\right)\cdots \left(I - (\eta_{1}/b) H_{(k-1)m+1}\right),
\end{equation}
\mgrev{is the product of consecutive $m$ iteration matrices}. We also define %the Lyapunov exponent of these matrix products as
$$\rho^{(m)} := \lim\nolimits_{k\to\infty} (2k)^{-1} \log\left(\mbox{largest eigenvalue of } \left(\Pi_k^{(m)}\right)^T \left(\Pi_k^{(m)}\right)\right), \quad \Pi_k^{(m)}:= M_k^{(m)} M_{k-1}^{(m)}\dots M_1^{(m)}. $$
%where $\Pi_k^{(m)}$ is defined as $\Pi_k^{(m)}:= M_k^{(m)} M_{k-1}^{(m)}\dots M_1^{(m)}$. %\mgrev{The latter quantity is a measure of the growth of matrix products that corresponds to $m$ consecutive SGD updates.}
{\color{black}We can iterate the SGD from \eqref{eq-stoc-grad-2} to obtain
$x_{(k+1)m}=M_{k+1}^{(m)}x_{km}+q_{k+1}^{(m)}$,
where $M_{k+1}^{(m)}$ is defined in \eqref{M:k:m:defn}
and $q_{k+1}^{(m)}:=\sum_{i=km+1}^{(k+1)m}
\left(I - \frac{\eta_{(k+1)m}}{b} H_{(k+1)m}\right)
\left(I - \frac{\eta_{(k+1)m-1}}{b} H_{(k+1)m-1}\right)\cdots \left(I - \frac{\eta_{i+1}}{b} H_{i+1}\right)q_{i}$.}
%where:
%\begin{align*}
%&M_{k+1}^{(m)}:=\left(I - \frac{\eta_{(k+1)m}}{b} H_{(k+1)m}\right)
%\left(I - \frac{\eta_{(k+1)m-1}}{b} H_{(k+1)m-1}\right)\cdots \left(I - \frac{\eta_{km+1}}{b} H_{km+1}\right),
%\\
%&q_{k+1}^{(m)}:=\sum_{i=km+1}^{(k+1)m}
%\left(I - \frac{\eta_{(k+1)m}}{b} H_{(k+1)m}\right)
%\left(I - \frac{\eta_{(k+1)m-1}}{b} H_{(k+1)m-1}\right)\cdots \left(I - \frac{\eta_{i+1}}{b} H_{i+1}\right)q_{i}.
%\end{align*}
We have the following result
that characterizes the tail-index.
%%%%%%%%%%%%%%%%%%%%%%%%%%%%%%%%%%%%%%%%%%%%%%%%%%%%%%%%%%%%%%%%%%%%%%%%%%%%
%In the Appendix, we state the convergence results in Theorem~\ref{thm:cyclic} which
%says $x_{km}$ converges to $x_{\infty}$ which has the tail-index $\alpha^{(m)}$ that is the unique
%positive value such that $h^{(m)}\left(\alpha^{(m)}\right)=1$.

\begin{theorem}\label{thm:cyclic}
Consider the SGD iterations (\ref{eq-stoc-grad-2}) with cyclic stepsizes \mgrev{$\{\eta_k\}$} \mgrev{where the length of the cycle is $m$}. If $\rho^{(m)}<0$ and there exists a unique positive $\alpha^{(m)}$ such that $h^{(m)}\left(\alpha^{(m)}\right)=1$,
then \mgrev{\eqref{eq-stoc-grad-2}} admits a unique stationary solution $x_{\infty}^{(m)}$
and the SGD iterations with cyclic stepsizes converge to $x_{\infty}^{(m)}$ in distribution,
where the distribution of $x_{\infty}^{(m)}$ satisfies
$\lim\nolimits_{t\to\infty} t^{\alpha^{(m)}} \mathbb{P}(u^T x_{\infty}^{(m)} > t )= e_{\alpha^{(m)}}(u)$, for any $u\in\mathbb{S}^{d-1}$,
for some positive and continuous function $e_{\alpha^{(m)}}$ on 
$\mathbb{S}^{d-1}$.
\end{theorem}

\textbf{Proof.}
This follows immediately from Theorem~\ref{thm:Markov:regeneration}, \mgrev{by noting that Markovian stepsizes reduce to the cyclic stepsizes in the special case when $p=1$}.
%By iterating the SGD iterates \mgrev{\eqref{eq-stoc-grad-2}} \mgrev{with cyclic stepsizes}, we obtain
%$x_{(k+1)m}=M_{k+1}^{(m)}x_{km}+q_{k+1}^{(m)}$,
%where:
%\begin{align*}
%&M_{k+1}^{(m)}:=\left(I - \frac{\eta_{(k+1)m}}{b} H_{(k+1)m}\right)
%\left(I - \frac{\eta_{(k+1)m-1}}{b} H_{(k+1)m-1}\right)\cdots \left(I - \frac{\eta_{km+1}}{b} H_{km+1}\right),
%\\
%&q_{k+1}^{(m)}:=\sum_{i=km+1}^{(k+1)m}
%\left(I - \frac{\eta_{(k+1)m}}{b} H_{(k+1)m}\right)
%\left(I - \frac{\eta_{(k+1)m-1}}{b} H_{(k+1)m-1}\right)\cdots \left(I - \frac{\eta_{i+1}}{b} H_{i+1}\right)q_{i}.
%\end{align*}
%Since $\eta_{k}$ are cyclic, 
%$M_{k+1}^{(m)}=\left(I - \frac{\eta_{m}}{b} H_{(k+1)m}\right)
%\left(I - \frac{\eta_{m-1}}{b} H_{(k+1)m-1}\right)\cdots \left(I - \frac{\eta_{1}}{b} H_{km+1}\right)$,
%and 
%$q_{k+1}^{(m)}=\sum_{i=km+1}^{(k+1)m}
%\left(I - \frac{\eta_{m}}{b} H_{(k+1)m}\right)
%\left(I - \frac{\eta_{m-1}}{b} H_{(k+1)m-1}\right)\cdots \left(I - \frac{\eta_{i-km+1}}{b} H_{i+1}\right)q_{i}$,
%which implies that
%$(M_{k}^{(m)},q_{k}^{(m)})$ are i.i.d. in $k$.
%%%%%%%%%%%%%%%%%%%%%%%%%%%%%%%%%%%%%%%%%%%%%%%%%%%%%%%%%%%%%%
%The rest of the proof follows from Theorem~4.4.15 in \cite{buraczewski2016stochastic} which goes back to Theorem~1.1 in \cite{alsmeyer2012tail} and Theorem~6 in \cite{kesten1973random}. 
%See also \cite{goldie1991implicit,bdp2015}.
\hfill $\Box$

%%%%%%%%%%%%%%%%%%%%%%%%%%%%%%%%%%%%%%%%%%%%%%%%%%%%%%%%%%%%%%%%%%%%%%%%%%%%%%%%%%%%%
%%%%%%%%%%%%%%%%%%%%%%%%%%%%%%%%%%%%%%
Determining the exact value of the tail-index $\alpha^{(m)}$ for the stationary distribution $x_\infty^{(m)}$ seems to be a hard problem; nevertheless, we can characterize a lower bound for the tail-index to control how heavy tailed SGD iterates can be \mgrev{by following a similar approach to our discussions for the Markovian stepsizes}. 
\mgrev{We start by noticing} that $h^{(m)}(s)=1$ if and only if $\left(h^{(m)}(s)\right)^{1/m}=1$. 
In general, there is no simple explicit formula for $h^{(m)}(s)$. 
However, we have the following bound
due to the sub-multiplicativity of the norm of matrix products:
$$\left(h^{(m)}(s)\right)^{1/m}
\leq
\hat{h}^{(m)}(s):=\left(\prod\nolimits_{i=1}^{m}\mathbb{E}\left[\left\Vert I-\frac{\eta_{i}}{b}H\right\Vert^{s}\right]\right)^{1/m}.$$Let $\hat{\alpha}^{(m)}$ be the unique
positive value such that $\hat{h}^{(m)}\left(\hat{\alpha}^{(m)}\right)=1$
provided that $\hat{\rho}^{(m)}:=\sum_{i=1}^{m}\mathbb{E}\left[\log\left\Vert I-\frac{\eta_{i}}{b}H\right\Vert\right]<0$. 
This provides a lower bound for the tail-index $\alpha^{(m)}$.   
Next, we show that the lower bound $\hat{\alpha}^{(m)}$ for tail-index is increasing in batch-size. \mgrev{We next state analogues of Thm.~\ref{thm:Markov:regeneration:index} and Thm.~\ref{thm:mono:Gaussian:Markov:regeneration} in the cyclic stepsize setting, the proofs are similar, with the only difference that regeneration times (the time it takes to revisit a particular stepsize) are random for Markovian stepsizes, whereas they are deterministic for cyclic stepsizes. The proofs are given in the appendix for the sake of completeness.}

\begin{theorem}\label{thm:cyclic:index}
$\hat{\alpha}^{(m)}$ is strictly increasing in batch-size $b$
provided that $\hat{\alpha}^{(m)}\geq 1$.
\end{theorem}

Under \textbf{(A3)}, 
we can get a more explicit formula for $h^{(m)}(s)$ and $\rho^{(m)}$ (see Lem.~\ref{lem:alternative:expression:cyclic}, Lem.~\ref{lem:simplified:cyclic} in the Appendix) and
hence can obtain further properties of the tail-index $\alpha^{(m)}$.
Under \textbf{(A3)}, we obtain the monotonic dependence
of the tail-index $\alpha^{(m)}$ on the batch-size $b$ and the dimension $d$.

\begin{theorem}\label{thm:mono:Gaussian:cyclic}
Assume \textbf{(A3)} holds and $\rho^{(m)}<0$. Then: (i) The tail-index $\alpha^{(m)}$ is strictly increasing in batch-size $b$
provided $\alpha^{(m)}\geq 1$. (ii) The tail-index $\alpha^{(m)}$ is strictly decreasing in dimension $d$.
\end{theorem}

\section{Comparisons of Tail-Indices}\label{sec:comparison}

In this section, we compare the tail-indices of SGD with constant, i.i.d., cyclic and Markovian stepsizes.
The i.i.d. stepsizes can be considered as a special case of the Markovian stepsizes
and in Sec.~\ref{sec:iid} in the Appendix, we study the SGD with i.i.d. stepsizes in detail.
Under Assumption~\textbf{(A3)}, we compare
the tail-index $\alpha$ of the SGD with i.i.d. stepsizes (where $\eta_{k}=\eta$ has the same distribution for every $k$) with the SGD with constant stepsize (where stepsize is fixed at $\mathbb{E}[\eta]$).

\begin{proposition}\label{prop:iid:constant:comparison}
Assume \textbf{(A3)} holds and $\rho<0$.
Then the tail-index $\alpha$ with i.i.d. stepsize is strictly less than the tail-index $\alpha_{c}$
with constant stepsize $\mathbb{E}[\eta]$ provided $\alpha\geq 1$.
\end{proposition}

In Prop.~\ref{prop:iid:constant:comparison}, 
we showed that the tail-index $\alpha$ is strictly less than the tail-index $\alpha_{c}$
with constant stepsize $\mathbb{E}[\eta]$.
%%%%%%%%%%%%%%%%%%%%%%%%%%%%%%%%%%%%%%%%%%%%
Under \textbf{(A3)}, we can also compare
the tail-index $\alpha^{(m)}$ of the SGD with cyclic stepsizes
and the tail-index $\alpha$ of the SGD with i.i.d. uniformly distributed stepsizes
such that
$\mathbb{P}(\eta=\eta_{i})=\frac{1}{m}$, for any $1\leq i\leq m$.
Then, we have the following result which says i.i.d. stepsizes have heavier tail.

\begin{proposition}\label{prop:compare:iid:cyclic}
Assume \textbf{(A3)} holds. 
The tail-index $\alpha$ for the SGD with i.i.d. stepsizes
such that $\mathbb{P}(\eta=\eta_{i})=\frac{1}{m}$ for any $1\leq i\leq m$
is smaller than the tail-index $\alpha^{(m)}$ for SGD with cyclic stepsizes.\looseness=-1
\end{proposition}

{\color{black}Next, we compare the the tail-index $\alpha^{(m)}$ with cyclic stepsizes and the tail-index $\alpha_{c}$ with constant stepsize $\frac{1}{m}\sum_{i=1}^{m}\eta_{i}$.
\mgrev{When the batch-size is not too large relative to the dimension (i.e. when $d\geq b+3$)}, we can show
that cyclic stepsizes lead to heavier tails under \textbf{(A3)}. \mgrev{The proof is based on exploiting log-convexity properties of the $h^{(m)}(\cdot)$ function when the batch-size is in this regime}.}
\begin{proposition}\label{prop:compare:cyclic:constant}
{\color{black}Assume \textbf{(A3)} holds and $d\geq b+3$. 
Then the tail-index $\alpha^{(m)}$ with cyclic stepsizes
is strictly smaller than the tail-index $\alpha_{c}$ 
with constant stepsize $\frac{1}{m}\sum_{i=1}^{m}\eta_{i}$ provided that $\eta_{i}$'s are sufficiently small.}
\end{proposition}

In general, for SGD with Markovian stepsizes, 
the regeneration time is hard to analyze. 
Next, we consider the simplest example of a Markov chain, 
i.e., a homogeneous Markov chain with two-state space $\{\eta_{l},\eta_{u}\}$
such that $\mathbb{P}(\eta_{1}=\eta_{u}|\eta_{0}=\eta_{l})=p$
and $\mathbb{P}(\eta_{1}=\eta_{l}|\eta_{0}=\eta_{u})=p$. 
This Markov chain exhibits a unique stationary distribution $\mathbb{P}(\eta_{0}=\eta_{\ell})=\mathbb{P}(\eta_{0}=\eta_{u})=\frac{1}{2}$. 
We notice that the special case $p=1$ reduces to the cyclic stepsizes. 
Indeed, we have the following monotonicity result
of the tail-index depending on the parameter $p$.

\begin{proposition}\label{cor:mono:p}
Consider the two-state Markov chain, i.e. $\mathbb{P}(\eta_{1}=\eta_{u}|\eta_{0}=\eta_{l})=p$
and $\mathbb{P}(\eta_{1}=\eta_{l}|\eta_{0}=\eta_{u})=p$ 
{\color{black}and assume that $p\in\mathcal{P}$, where 
\begin{equation}\label{P:set}
\mathcal{P}:=\left\{p\in[0,1]:(1-p)\max\left(\mathbb{E}_{H}\left[\left\Vert \left(I-\frac{\eta_{l}}{b}H\right)e_{1}\right\Vert^{\alpha^{(r)}}\right],
\mathbb{E}_{H}\left[\left\Vert \left(I-\frac{\eta_{u}}{b}H\right)e_{1}\right\Vert^{\alpha^{(r)}}\right]\right)<1\right\}.
\end{equation}
} 
Then, the tail-index $\alpha^{(r)}$ is increasing in {\color{black}$p\in\mathcal{P}$.}
In particular, $\alpha^{(r)}\leq\alpha^{(m)}$.
\end{proposition}

Prop.~\ref{cor:mono:p} shows that Markovian stepsizes
lead to heavier tails than cyclic stepsizes.
In Prop.~\ref{prop:compare:iid:cyclic}, we showed
that SGD with i.i.d. stepsizes lead to heavier tail
than the SGD with cyclic stepsizes, 
{\color{black}which has heavier tail than the SGD with constant stepsizes (Prop.~\ref{prop:compare:cyclic:constant}).}
The next result states that Markovian stepsizes may lead to heavier tails than i.i.d. stepsizes
depending on whether $p$ is greater or less than $\frac{1}{2}$.

\begin{proposition}\label{cor:comparison}
Under the setting of Propositions~\ref{prop:compare:cyclic:constant} and \ref{cor:mono:p}, 
we have $\alpha<\alpha^{(r)}<\alpha^{(m)}\mgrev{<\alpha_{c}}$ for any $\frac{1}{2}<p<1$
and we have $\alpha^{(r)}<\alpha<\alpha^{(m)}<\alpha_{c}$ for any $p<\frac{1}{2}$, 
where {\color{black}$\alpha_{c}$,} $\alpha$, $\alpha^{(m)}$, $\alpha^{(r)}$ denote the tail-index
for SGD with {\color{black}constant,} i.i.d., cyclic and Markovian stepsizes respectively.
\end{proposition}

%%%%%%%%%%%%%%%%%%%%%%%%%%%%%%
%\vspace{-0.2in}
\section{Numerical Experiments}\label{sec:numerical}

In this section, we present the numerical experiments including the tail-index estimation of linear regression using uniform stepsize and Markovian stepsize and the performance of uniform, Markovian, cyclic and constant stepsize on deep learning settings.

%%%%%%%%%%%%%%%%%%%%%%%%%%%%%%%%%%%%%%%%%%%%%%%%%%%%%%%%%%
%\subsection{Quadratic Optimization}
%\vspace{-0.1in}
\paragraph{Linear regression (least squares).} In the following experiments, we investigate the relationship between the tail-index and the stepsize choice. We consider the following model:
$w \sim \mathcal{N}\left(0,\sigma^2I\right)$, $x_i\sim \mathcal{N}\left(0,\sigma_x^2I\right)$, and $y_i|w,x_i\sim \mathcal{N}\left(x_i^Tw,\sigma^2_y\right)$,
where $w, x_i \in \mathbb{R}^d$, $y_i\in \mathbb{R}$ for $i=1,\ldots ,n$, and $\sigma, \sigma_x, \sigma_y>0$. In the experiments, we set $d=100$, take $\sigma=3,\sigma_x=1,\sigma_y=3$, and generate $\{x_i,y_i\}_{i=1}^n$ by simulating the statistical model. To estimate the stationary measure, we iterate SGD for $1000$ iterations and we repeat this for $10000$ runs. 
At each run, we take the average of the last $500$ SGD iterates; in the Appendix (Cor.~\ref{cor:clt}, \ref{cor:clt:cyclic} and \ref{cor:clt:Markovian}) we argue that the average follows an $\alpha$-stable distribution under some assumptions. This enables us to use advanced estimators \citep{mohammadi2015estimating} specific to $\alpha$-stable distributions. For constant stepsize schedules, we set the stepsize to be constant $\hat{\eta}$ during the training process. For i.i.d. uniformly distributed stepsize, cyclic stepsize, and Markovian stepsize, the interval of stepsize is set as $[\hat{\eta}-R, \hat{\eta}+R]$. 

\begin{figure}[t]
    \centering
    \subfigure[Different stepsize schedules]{
    \includegraphics[width = 0.31\textwidth]{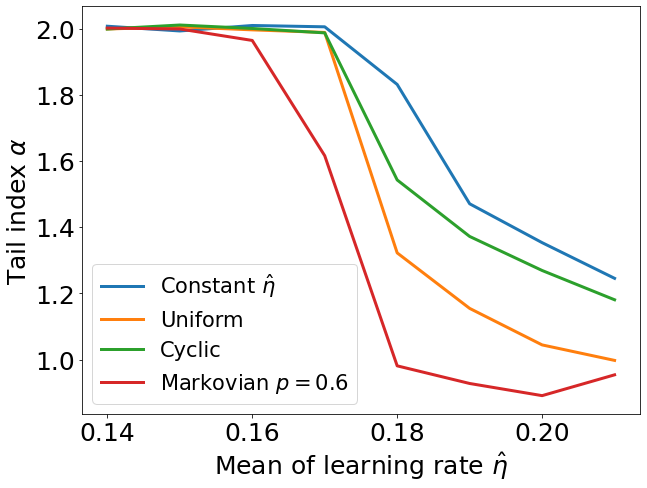}\label{fig:lin-reg-comparison}
    }
    \subfigure[Markovian with parameter $p$]{
    \includegraphics[width=0.31\textwidth]{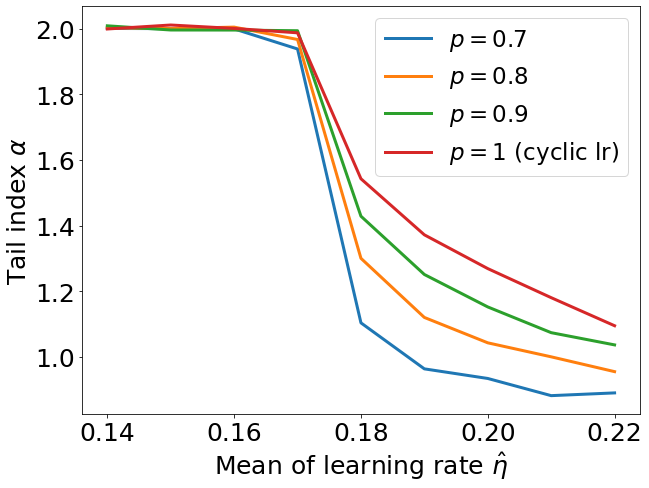}
    \label{fig:mc_p}
    }
    \caption{Center of the stepsize interval ($\hat \eta$) vs. tail-index \mgrev{for linear regression}. \mgrev{(Left panel) Comparison of constant, uniform, cyclic and Markovian stepsizes with $p=0.6$ with batch-size $b=10$, and the stepsize grid parameters $R = 0.05$ and $K=10$. (Right panel) Comparison of Markovian stepsizes with different transition probabilities $p$ when the mean of the learning rate $\hat\eta$ is varied, with batch-size $b=10$, $R=0.05$.}}
\end{figure}

To compare different stepsize (learning rate) schedules, we calculate the tail-index for the linear regression experiment in Fig.~\ref{fig:lin-reg-comparison}. In this experiment, we set batch-size 10 for all stepsize schedules, $R = 0.05$ and $K=10$ for uniform, cyclic, and Markovian stepsize schedules. For Markovian stepsize, we set $p=0.6$. We can conclude from the figure that the tail-index of all these stepsize schedules will decrease when the mean of stepsize $\hat{\eta}$ increases, meanwhile the tail-index of SGD with i.i.d. uniformly distributed stepsize is smaller than the tail-index of cyclic stepsize schedule, which validate our Prop.~\ref{prop:compare:iid:cyclic}. We can also observe that the tail-index of Markovian stepsize is smaller than the cyclic stepsize schedule, which is predicted by our Prop.~\ref{cor:comparison}. 

\begin{figure} 
\centering
\subfigure[i.i.d. stepsize]{
\includegraphics[width=0.31\textwidth]{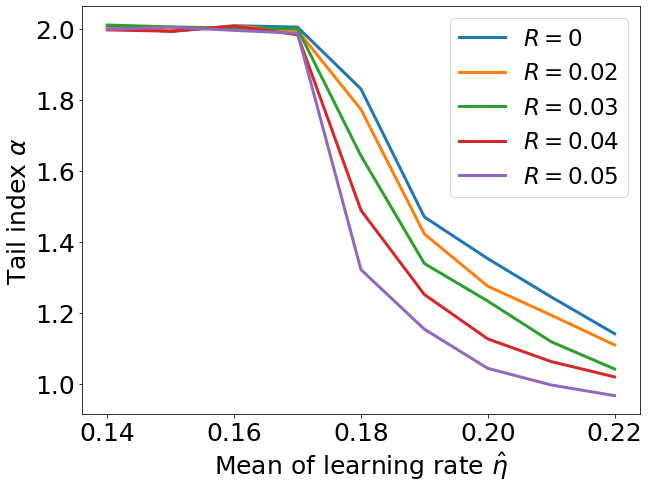}
\label{fig:uniform_delta}
}
\subfigure[Cyclic stepsize]{
\includegraphics[width=0.31\textwidth]{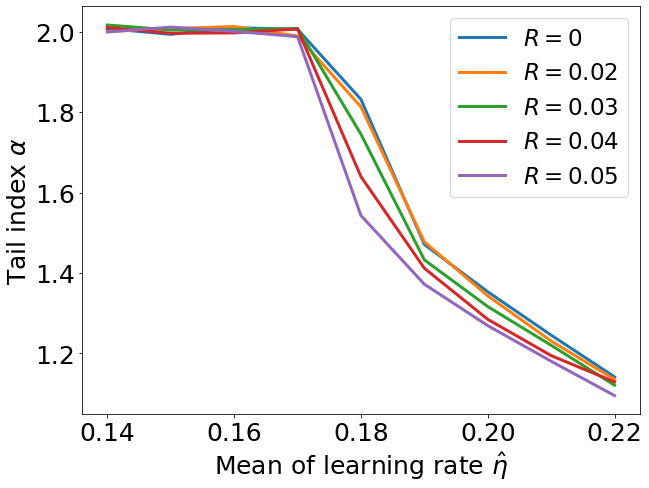}
\label{fig:cyclic_delta}
}
\subfigure[Markovian stepsize]{
\includegraphics[width=0.31\textwidth]{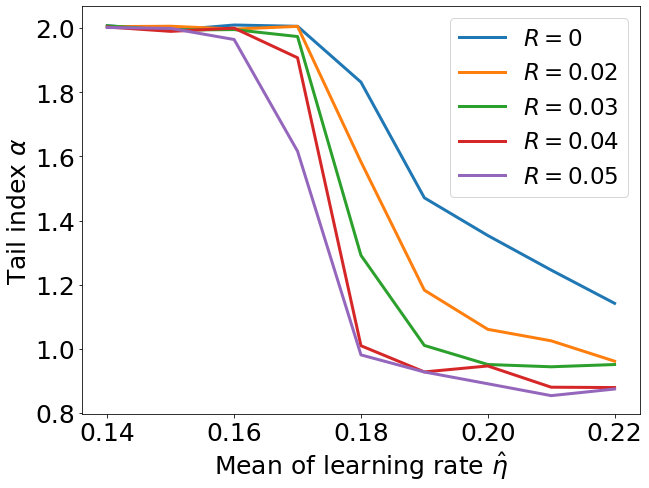}
\label{fig:mc_delta}
}
\caption{Tail-index corresponding to different range values ($R$) as \mgrev{a function of the mean stepsize $\hat\eta$ for linear regression}, \mgrev{for i.i.d. random stepsizes (left panel), for cyclic stepsize (middle panel) and for Markovian stepsizes with $p=0.6$ (right panel). We fix $K=10$ and vary $\delta$ from $0$ to $0.05$.}}
\label{fig:different_R}
%\vspace{-0.2in}
\end{figure}

In the next sets of experiments, we will investigate the influence of different values of parameters on the tail-index for different stepsize schedules. In the first experiment, we test different transition probability $p$ for Markovian stepsize. The result is shown in Fig.~\ref{fig:mc_p}, where the values of $p$ is varied from $0.6$ to $1$ where we set batch-size $b=10$, $R=0.05$. 
When $p=1$, the Markovian stepsize schedule will degenerate to cyclic stepsize. We observe that with a smaller $p$ value, the tail-index of Markovian stepsize will become smaller. This result validates our Prop.~\ref{cor:mono:p}. In another experiment, we test different range $R$ values for i.i.d. uniformly distributed, cyclic and Markovian stepsize schedules where we set the batch-size $b= 10$ and $p=0.6$ for Markovian stepsize. To vary $R=(K-1)\delta/2$, we keep $K=10$ as a constant and vary $\delta$ from $0$ to $0.05$. As shown in Fig.~\ref{fig:different_R}, the tail-index of all schedules will decrease when the range $R$ increases. These results validate our Thm.~\ref{thm:mono:Markovian:range} and Thm.~\ref{thm:mono:Gaussian:range} in the Appendix. Finally, in Fig.~\ref{fig:different_batch}, we test different batch-size values from $5$ to $15$ where we set range $R=0.05$ and for Markovian stepsize $p=0.6$. %{\color{red} how do we set K?}. 
We observe that the tail-index of every stepsize schedule is strictly increasing in batch-size, which is consistent with our theory (Thm.~\ref{thm:mono:Gaussian:cyclic},~\ref{thm:mono:Gaussian:Markov:regeneration} and Thm.~\ref{thm:mono:Gaussian} in the Appendix). 
% {\color{red}Figure 2(c) and Figure 3, chop the end of all plots whhere the behavior is not monotonic?}

% \begin{figure}
% \centering
% \subfigure[Uniform lr]{
% \includegraphics[width=0.3\textwidth]{uniform_linreg_002.png}
% \label{fig:uniform}
% }
% \subfigure[Markovian lr]{
% \includegraphics[width=0.3\textwidth]{Markov_linreg_002.png}
% \label{fig:markov}
% }
% \subfigure[Markovian lr]{
% \includegraphics[width=0.3\textwidth]{Markov_linreg_002_p.png}
% \label{fig:markov p}
% }
% \caption{Tail-index of different learning rate setting.}
% \end{figure}

\begin{figure}[t!]
\centering
\subfigure[Uniform stepsize]{
\includegraphics[width=0.31\textwidth]{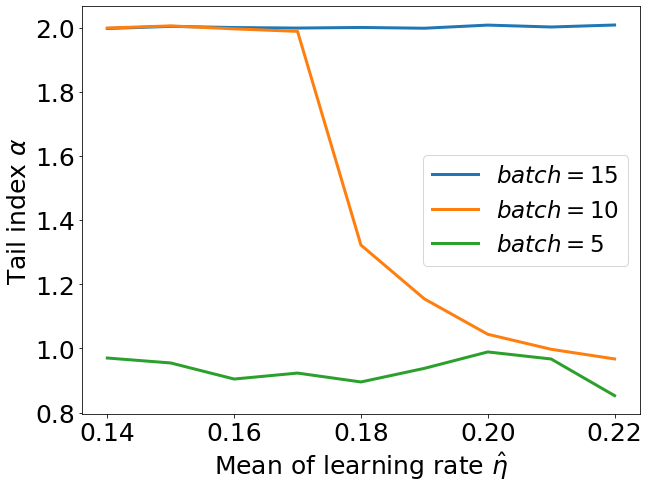}
\label{fig:uniform_batch}
}
\subfigure[Cyclic stepsize]{
\includegraphics[width=0.31\textwidth]{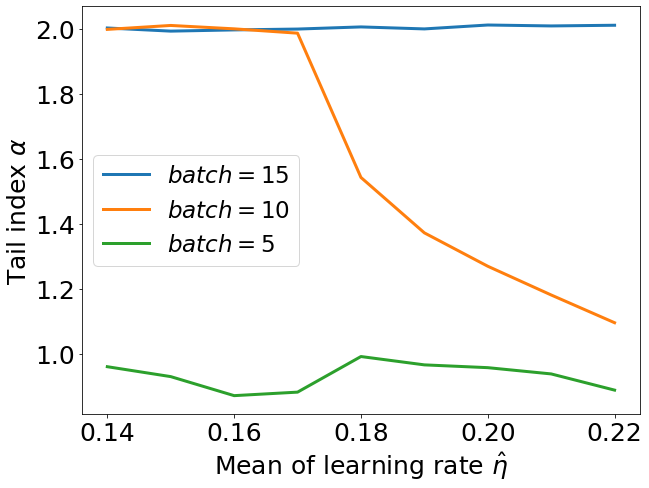}
\label{fig:cyclic_batch}
}
\subfigure[Markovian stepsize]{
\includegraphics[width=0.31\textwidth]{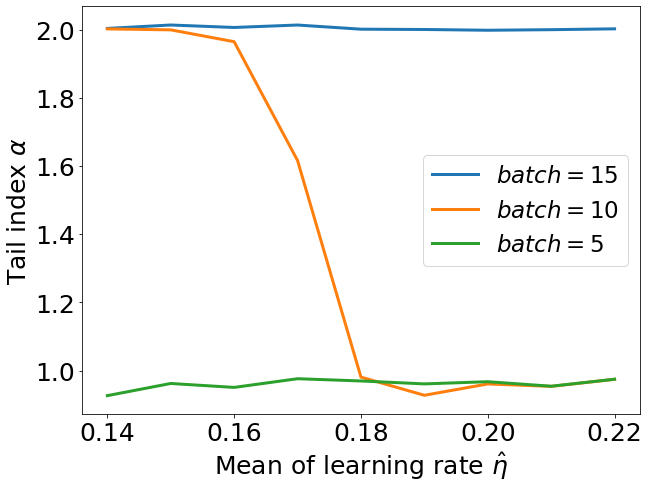}
\label{fig:mc_batch}
}
\caption{Tail-index of batch-size value \mgrev{for linear regression for uniform stepsize (left panel), cyclic stepsize (middle panel) and Markovian stepsize with $p=0.6$ (right panel). We test different batch-size values $b \in \{5,10,15\}$ with range parameter $R=0.05$ and $K=10$}. }
\label{fig:different_batch}
%\vspace{-0.2in}
\end{figure}

%%%%%%%%%%%%%%%%%%%%%%%%%%%%%%%%%%%%%%%%%%%%%%%%%%%%%%%%%%%
%\vspace{-0.1in}
\paragraph{Deep Learning.} In our second set of experiments, we investigate the performance of uniform stepsize, Markovian stepsize, cyclic stepsize, and constant stepsize beyond the linear regression setting. Here we consider the 3-layer fully connected network with the cross entropy loss on the MNIST dataset. We train the models using SGD with batch-size $b=20$. Similar to the linear regression setting, we test different stepsizes centered at $\hat{\eta}$ where we vary $\hat{\eta}$ while keeping the range of stepsize fixed with $R=0.05$ and $K=100$ for cyclic and Markovian stepsize schedules. We display the results of the deep learning experiment in Table~\ref{tab:MNIST_3fcn}. Following the literature \citep{chen2018stability}, we measure generalization in terms of the difference between the training and test loss; the smaller difference is the better generalization. As an alternative but correlated metric, we also consider the difference between test and training accuracy for quantifying the generalization performance. For different center stepsize $\hat{\eta}$, we can observe that uniform, cyclic, and Markovian stepsize schedules lead to a smaller tail-index compared to the constant stepsize. While the tail-index and heavy tail are well correlated according to existing theoretical and numerical results \citep{simsekli2020hausdorff,barsbey2021heavy,raj-quadratic,raj-general,HTP2021}, the relationship is not a perfect straight line. For example, we can see that for $\hat{\eta}=0.06$ case, the smallest tail-index does not lead to the better generalization power in the sense of smaller error difference and accuracy difference. But for all the other cases where $\hat{\eta}=0.08,0.09$, Markovian stepsize schedule with $p=0.5$ has both the smallest tail-index and best generalization performance. 

\mgrev{As a larger-scale deep learning experiment, we next consider the VGG-11 architecture that has 8 convolutional layers and 3 fully-connected layers and we use the CIFAR10 dataset. For the stepsize grid parameters, we use $R=0.03$ and $K=100$ and take batchsize $b=25$. We use SGD for training and compare different stepsize schedules in Table \ref{table-vgg} for different choices of the mean stepsize $\hat{\eta}$, similarly to the previous experiment. We observe that i.i.d. stepsize has the heaviest tails (with the lowest tail index) in all cases, and the constant stepsize has the lightest tails (with the highest tail index) whereas the cyclic and Markovian stepsizes are in between except when $\hat\eta=0.12$. As a general trend, these results are roughly inline with our results given in Prop. \ref{cor:comparison} that compares the tails of different stepsizes schedules under some assumptions. In Figure \ref{table-vgg}, we also see that the heaviest tails led to the best generalization in terms of accuracy, except for the $\hat\eta=0.11$ case. It would be interesting to investigate how the tails are related to generalization and optimization performance further as a part of future work. %There are also some results in optimization literature that says that heavier tails can make the training slower, we cited these results in the paper. Our paper provides a first-time understanding of the heaviness of the tails for different stepsize schedules. Therefore, we believe that our paper can potentially be beneficial to both lines of literature on the optimization and generalization and their relationship to the tail index. 
}

\mgrev{To summarize,} our results suggest that good performance associated to cyclic, randomized and Markovian stepsize can be due to the incurrence of heavier tails compared to constant stepsize in the deep learning settings. While deep learning setting is significantly more complicated than the linear regression setting we considered in our theoretical results, our results offer theoretical support into why alternative stepsizes (randomized, cyclic) can be successful and offers new Markovian stepsize rules that can perform better in some cases.
%Therefore, in the first case, we see that while constant stepsize leads to lighter tail it can still result in similar generalization compared to other stepsize rules. %That being said, Markovian and randomized stepsizes are viable alternatives that can generalize better on
%In this experiment, the relationship between generalization power and tail-index is not clear and may need future investigation.

%{\color{red} Accuracy, and its diff should be displayed as percentage. any other examples where markovian generalizes worse but lead to heavier tail?}
\begin{table}[]
\resizebox{\columnwidth}{!}{%
\begin{tabular}{|c|c|c|c|c|c|c|c|c|}
\hline
Schedule  &$\hat{\eta}$    &Train Err  &Test Err &Error diff &Train Acc  &Test Acc   &Acc diff   &Tail-index\\ \hhline{|=========|}
Constant & \multirow{5}{*}{0.06} & 3.81E-07    & 0.0055 & \textbf{5.54E-03} & 100\%         & 98.39\% & \textbf{1.61\%}    & 1.95 \\ \cline{1-1} \cline{3-9} 
Uniform  &                       & 1.70E-07    & 0.0067 & 6.69E-03 & 100\%         & 98.38\% & 1.62\%    & 1.93 \\ \cline{1-1} \cline{3-9} 
Cyclic   &                       & 3.27E-07    & 0.0065 & 6.50E-03 & 100\%         & 98.33\% & 1.67\%    & 1.98 \\ \cline{1-1} \cline{3-9} 
\mgrev{Markovian $p$=}0.6   &                       & 2.97E-07    & 0.0062 & 6.16E-03 & 100\%         & 98.36\% & 1.64\%    & 1.90 \\ \cline{1-1} \cline{3-9} 
\mgrev{Markovian $p$=}0.5   &                       & 2.79E-07    & 0.0068 & 6.84E-03 & 100\%         & 98.34\% & 1.66\%    & \textbf{1.90} \\ 
% \hhline{|=========|}
% Constant & \multirow{5}{*}{0.07} & 2.12E-07    & 0.0055 & \textbf{5.46E-03} & 1         & 0.9853 & \textbf{0.0147}    & 2.1125  \\ \cline{1-1} \cline{3-9} 
% Uniform  &                       & 1.78E-07    & 0.0075 & 7.54E-03 & 1         & 0.9825 & 0.0175    & 1.9154  \\ \cline{1-1} \cline{3-9} 
% Cyclic   &                       & 0.00154     & 0.0282 & 2.67E-02 & 0.9945    & 0.9739 & 0.0207    & \textbf{1.8149}  \\ \cline{1-1} \cline{3-9} 
% MC 0.6   &                       & 1.80E-07    & 0.0076 & 7.55E-03 & 1         & 0.9833 & 0.0167    & 1.9006 \\ \cline{1-1} \cline{3-9} 
% MC 0.5   &                       & 2.75E-07    & 0.0063 & 6.28E-03 & 1         & 0.9838 & 0.0162    & 1.8903 \\ 
\hhline{|=========|}
Constant & \multirow{5}{*}{0.08} & 0.00194     & 0.0332 & 3.13E-02 & 99.55\%    & 97.64\% & 1.91\%    & 1.95 \\ \cline{1-1} \cline{3-9} 
Uniform  &                       & 0.00221     & 0.0346 & 3.24E-02 & 99.59\%    & 97.63\% & 1.96\%    & 1.94 \\ \cline{1-1} \cline{3-9} 
Cyclic   &                       & 0.01490     & 0.0346 & 1.97E-02 & 94.99\%    & 93.17\% & 1.82\%    & 1.78  \\ \cline{1-1} \cline{3-9} 
\mgrev{Markovian $p$=}0.6   &                       & 0.00055     & 0.0198 & 1.93E-02 & 99.76\%    & 97.80\% & 1.96\%    & 1.89 \\ \cline{1-1} \cline{3-9} 
\mgrev{Markovian $p$=}0.5   &                       & 2.10E-07    & 0.0071 & \textbf{7.14E-03} & 100\%         & 98.35\% & \textbf{1.65\%}    & \textbf{1.69}  \\ \hhline{|=========|}
Constant & \multirow{5}{*}{0.09} & 0.00154     & 0.0366 & 3.50E-02 & 99.61\%   & 97.71\% & 1.90\%    & 1.85 \\ \cline{1-1} \cline{3-9} 
Uniform  &                       & 0.00208     & 0.0283 & 2.62E-02 & 99.49\%    & 97.48\% & 2.01\%    & 1.77 \\ \cline{1-1} \cline{3-9} 
Cyclic   &                       & 0.00294     & 0.0316 & 2.87E-02 & 99.18\%    & 97.22\% & 1.96\%    & 1.62 \\ \cline{1-1} \cline{3-9} 
\mgrev{Markovian $p$=}0.6   &                       & 0.00433     & 0.0266 & 2.22E-02 & 99.04\%    & 97.05\% & 1.99\%    & 1.69 \\ \cline{1-1} \cline{3-9} 
\mgrev{Markovian $p$=}0.5   &                       & 0.00168     & 0.0217 & \textbf{2.00E-02} & 99.38\%    & 97.52\% & \textbf{1.86\%}    & \textbf{1.51}  \\ \hhline{|=========|}
\end{tabular}
}
\caption{3-layer fully connected network on the MNIST dataset. \mgrev{We vary the mean of the stepsize $\hat\eta$ and compare the stepsize schedules, where in the third to nineth columns we report the training error, test error, difference between the test and training errors, training accuracy, test accuracy, difference between the test and training accuracy and the tail-index respectively.}}
%\vspace{-0.2in}
\label{tab:MNIST_3fcn}
\end{table}

%%% VGG TABLE NEXT
\begin{table}[]
\resizebox{\columnwidth}{!}{%
\begin{tabular}{|c|c|c|c|c|c|c|c|c|}
\hline
Schedule  &$\hat{\eta}$    &Train Err  &Test Err &Error diff &Train Acc  &Test Acc   &Acc diff   &Tail-index\\ \hhline{|=========|}
Constant & \multirow{4}{*}{0.10}  & 3.49E-08  & 0.012 & 1.18E-02          & 1         & 85.15\%   & 14.85\%          & 1.83           \\  \cline{1-1} \cline{3-9} 
Uniform  &                       & 3.38E-08  & 0.011 & 1.15E-02          & 1         & 85.19\%   & \textbf{14.81\%} & \textbf{1.80} \\ \cline{1-1} \cline{3-9}
Cyclic   &                       & 3.81E-08  & 0.011  & \textbf{1.14E-02} & 1         & 84.69\%   & 15.31\%          & 1.82          \\ \cline{1-1} \cline{3-9}
Markovian $p$=0.7   &                       & 2.81E-08  & 0.012 & 1.18E-02          & 1         & 85.05\%   & 14.95\%          & 1.83          \\ \cline{1-1} \cline{3-9}
         \hhline{|=========|}
Constant & \multirow{4}{*}{0.11} & 5.59E-08  & 0.011 & \textbf{1.14E-02} & 1         & 85.21\%   & \textbf{14.79\%} & 1.84          \\  \cline{1-1} \cline{3-9} 
Uniform  &                       & 3.39E-08  & 0.012 & 1.20E-02          & 1         & 84.77\%   & 15.23\%          & \textbf{1.79} \\  \cline{1-1} \cline{3-9} 
Cyclic   &                       & 2.74E-08  & 0.012 & 1.17E-02          & 1         & 84.91\%   & 15.09\%          & 1.82         \\  \cline{1-1} \cline{3-9} 
Markovian $p$=0.7   &                       & 3.24E-08  & 0.012 & 1.16E-02          & 1         & 85.13\%   & 14.87\%          & 1.79           \\  \cline{1-1} \cline{3-9} 
         \hhline{|=========|}
Constant & \multirow{4}{*}{0.12} & 3.99E-08  & 0.012 & 1.18E-02          & 1         & 85.33\%   & 14.67\%          & 1.80          \\  \cline{1-1} \cline{3-9} 
Uniform  &                       & 6.85E-08  & 0.012 & 1.21E-02          & 1         & 85.41\%   & \textbf{14.59\%} & \textbf{1.76} \\  \cline{1-1} \cline{3-9} 
Cyclic   &                       & 4.25E-08  & 0.012 & \textbf{1.16E-02} & 1         & 85.15\%   & 14.85\%          & 1.80          \\  \cline{1-1} \cline{3-9} 
Markovian $p$=0.7   &                       & 3.02E-08  & 0.013 & 1.29E-02          & 1         & 84.57\%   & 15.43\%          & 1.81         \\   \cline{1-1} \cline{3-9} \hhline{|=========|}
\end{tabular}
}
\caption{\mgrev{VGG11 network on the CIFAR10 dataset. We vary the mean of the stepsize $\hat\eta$ and compare the stepsize schedules, where in the third to nineth columns we report the training error, test error, difference between the test and training errors, training accuracy, test accuracy, difference between the test and training accuracy and the tail-index respectively.\label{table-vgg}}}
\end{table}
\section{Conclusion}

In this work, we proposed Markovian stepsizes which recovers uniformly random, cyclic and constant stepsizes as special cases. We developed proof techniques where we show that uniformly random, cyclic and Markovian stepsizes can lead to heavier tails in the distribution of SGD iterates. Since smaller tail-indices are correlated with better generalization in deep learning practice, our results shed further light into why random and cyclic stepsizes can perform better than constant stepsize in deep learning. We also showed that our proposed Markovian stepsizes can be a viable alternative to other standard choices in the literature, leading to the best performance in some cases.

\subsubsection*{Acknowledgments}
Mert G\"{u}rb\"{u}zbalaban and Yuanhan Hu acknowledge Rutgers Business School for creating a supportive research atmosphere, most of this work was completed when Yuanhan Hu was a Ph.D. student at the Rutgers Business School. Mert G\"{u}rb\"{u}zbalaban and Yuanhan Hu's research are supported in part by the grants Office of Naval Research Award Number
N00014-21-1-2244, National Science Foundation (NSF)
CCF-1814888, NSF DMS-2053485.  Umut \c{S}im\c{s}ekli's research is supported by the French government under management of Agence Nationale de la Recherche as part of the "Investissements d'avenir" program, reference
ANR-19-P3IA-0001 (PRAIRIE 3IA Institute) and the European Research Council Starting Grant
DYNASTY - 101039676.
Lingjiong Zhu is grateful to the partial support from a Simons Foundation Collaboration Grant
and the grant NSF DMS-2053454 from the National Science Foundation.

\bibliography{heavy}
\bibliographystyle{tmlr}
\newpage
\appendix

\section{Technical Background}

%%%%%%%%%%%%%%%%%%%%%%%%%%%%%%%%%%%%%%%%%%%%%%%%%%%%
\textbf{Wasserstein metric.}
For any $p\geq 1$, define $\mathcal{P}_{p}(\mathbb{R}^{d})$
as the space consisting of all the Borel probability measures $\nu$
on $\mathbb{R}^{d}$ with the finite $p$-th moment
(based on the Euclidean norm).
For any two Borel probability measures $\nu_{1},\nu_{2}\in\mathcal{P}_{p}(\mathbb{R}^{d})$, 
we define the standard $p$-Wasserstein
metric as \citep{villani2008optimal}:
$$\mathcal{W}_{p}(\nu_{1},\nu_{2}):=\left(\inf\mathbb{E}\left[\Vert Z_{1}-Z_{2}\Vert^{p}\right]\right)^{1/p},$$
where the infimum is taken over all joint distributions of the random variables $Z_{1},Z_{2}$ with marginal distributions
$\nu_{1},\nu_{2}$.

%%%%%%%%%%%%%%%%%%%%%%%%%%%%%%%%%
\section{Technical Results}

%%%%%%%%%%%%%%%%%%%%%%%%%%%%%%%%%%%%%%%%%%%%%%%%%%%%%%%
\subsection{Stochastic Gradient Descent with Constant Stepsizes}\label{sec:const}

In this section, let us recall some technical results from \cite{HTP2021}
for the SGD with constant stepsizes. 
When the stepsizes $\eta_{k}\equiv\eta$ are constant, 
the SGD iterates are given by
\begin{equation} 
x_{k+1} = x_k - \eta \tilde \nabla f_{k+1} \left(x_k\right)\,,
\label{eq-stoc-grad-constant}
\end{equation}  
where $\eta>0$ is the stepsize and $\tilde \nabla f_{k}(x) := \frac{1}{b}\sum_{i \in \Omega_{k}}\nabla f_{i}(x)$. 
We first observe that SGD  \eqref{eq-stoc-grad-constant} is an iterated random recursion of the form
\begin{equation} 
x_k = \Psi(x_{k-1},\Omega_k),
\end{equation} 
where the map $\Psi:\mathbb{R}^d\times \mathcal{S}\to \mathbb{R}^d$, $\mathcal{S}$ denotes the set of all subsets of $\{1,2,\dots,n\}$ and $\Omega_k$ is random and i.i.d. 
If we write $\Psi_{\Omega}(x) = \Psi(x,\Omega)$ 
for simplicity where $\Omega$ has the same distribution as $\Omega_k$, 
and assume that the random map $\Psi_{\Omega}$ is Lipschitz on average, i.e.  
$\mathbb{E}[L_{\Omega}] < \infty$ with $L_{\Omega} := \sup\nolimits_{x,y\in\mathbb{R}^d} \frac{\|\Psi_{\Omega}(x) - \Psi_{\Omega,\eta}(y) \|}{\|x-y\|}$,
and is mean-contractive, i.e. if $\mathbb{E}\log(L_{\Omega})<0$ then it can be shown under further technical assumptions that the distribution of the iterates converges to a unique stationary distribution $x_\infty$ geometrically fast \citep{diaconis1999iterated}. 
We recall the following result from \cite{HTP2021} that 
characterize the tail-index for $x_{\infty}$.

%%%%%%%%%%%%%%%%%%%%%%%%%%%%%%%%%

\begin{theorem}[Theorem~1 in \cite{HTP2021}, see also \citet{mirek2011heavy}]\label{thm-mirek} 
Assume stationary solution to $x_k = \Psi_{\Omega_k}(x_{k-1})$ exists and:%. We further assume 

(i) {There exists a random matrix $M(\Omega)$ and a random variable $B(\Omega)>0$ such that for a.e. $\Omega$, 
$|\Psi_\Omega(x)-M(\Omega) x | \leq B(\Omega)$ for every $x$;}

(ii) The conditional law of $\log |M(\Omega)|$ given $M(\Omega)\neq 0$ is non-arithmetic; i.e. \mgrev{its support is not equal to $a\mathbb{Z}$ for any scalar $a$ where $\mathbb{Z}$ is the set of integers}.

(iii) There exists $\alpha_{c}>0$ such that $\mathbb{E} |M(\Omega)|^{\alpha_{c}} = 1$,  $\mathbb{E} |B(\Omega)|^{\alpha_{c}} <\infty$ and 
\begin{equation*}
\mathbb{E} [|M(\Omega)|^{\alpha_{c}} \log^+ |M(\Omega)|]< \infty, 
\end{equation*}
where $\log^+(x) := \max(\log(x),0)$.

Then, it holds that $\lim_{t \to \infty } t^{\alpha_{c}} \mathbb{P}(|x_\infty|>t) = c_{0,c}$ for some constant $c_{0,c}>0$.  
\end{theorem}
%%%%%%%%%%%%%%%%%%%%%%%%%%%%%%%%%%%%%%%%%%%%%%%%%%%%%%%%%%%%%%%%%%

When the objective is quadratic, it is possible
to characterize the tail-index $\alpha_{c}$ in a more explicit way and also go beyond the one-dimensional case. For the quadratic objective, we can rewrite SGD iterations \eqref{eq-stoc-grad-constant} as 
\begin{equation} 
x_{k+1} = \left(I - (\eta/b) H_{k+1}\right) x_k  + q_{k+1}\,,
\end{equation}  
where 
$H_k := \sum_{i\in \Omega_{k} } a_i a_i^T$ and $q_k := \frac{\eta}{b} \sum_{i\in \Omega_{k} } y_i$.
Let us introduce
\begin{equation}
h_{c}(s) := \lim\nolimits_{k\to\infty}\left(\mathbb{E}\| M_k M_{k-1}\dots M_1\|^s\right)^{1/k}\,,
\label{def-hs}
\end{equation}
where
$M_{k}:=I-\frac{\eta}{b}H_{k}$,
which arises in stochastic matrix recursions (see e.g. \cite{buraczewski2014multidimensional}) where $\|\cdot\|$ denotes the matrix 2-norm (i.e. largest singular value of a matrix).  
Since $\mathbb{E}\|M_k\|^s < \infty$ for all $k$ and $s>0$, we have $h_{c}(s) < \infty$.
Let us also define
\begin{equation}
\rho_{c} := \lim\nolimits_{k\to\infty} (2k)^{-1} \log\left(\mbox{largest eigenvalue of } \Pi_k^T \Pi_k\right)\,,
\label{def-rho}
\end{equation}
where $\Pi_k := M_k M_{k-1}\dots M_1$.
In \eqref{def-rho}, the quantity $\rho_{c}$ is
called the top Lyapunov exponent of the stochastic recursion (\ref{eq-stoc-grad-2}). 
Furthermore, if $\rho_{c}$ exists and is negative, it can be shown that
a stationary distribution of the recursion (\ref{eq-stoc-grad-2}) exists.
Indeed, we have the following result from \cite{HTP2021} that characterizes the tail-index
for the stationary distribution.

\begin{theorem}[Theorem~2 in \cite{HTP2021}]\label{thm:main}
Consider the SGD iterations (\ref{eq-stoc-grad-2}).
If $\rho_{c}<0$ and there exists a unique positive $\alpha_{c}$ such that $h_{c}(\alpha_{c})=1$, 
where $h_{c}$ and $\rho_{c}$ are defined in \eqref{def-hs} and \eqref{def-rho},
then \eqref{eq-stoc-grad-2} admits a unique stationary solution $x_{\infty}$
and the SGD iterations converge to $x_{\infty}$ in distribution,
where the distribution of $x_{\infty}$ satisfies
\begin{equation} 
\lim\nolimits_{t\to\infty} t^{\alpha_{c}} \mathbb{P}\left(u^T x_{\infty} > t \right)= e_{\alpha_{c}}(u)\,, \quad u\in\mathbb{S}^{d-1}\,,\label{eq-heavy-tail}
\end{equation}
for some positive and continuous function $e_\alpha$ on 
$\mathbb{S}^{d-1}$. 
\end{theorem}

In general, the tail-index $\alpha_{c}$ does not have a simple formula
since $h_{c}(s)$ function lacks a simple expression. 
A lower bound $\hat{\alpha}_{c}\leq\alpha_{c}$ holds
where $\hat{\alpha}_{c}$ is the unique positive
solution to $\hat{h}_{c}\left(\hat{\alpha}_{c}\right)=1$, 
where $\hat{h}_{c}(s):=\mathbb{E}\left[\left\Vert I-\frac{\eta}{b}H_{1}\right\Vert^{s}\right]$, 
provided that $\hat{\rho}_{c}:=\mathbb{E}\log\left\Vert I-\frac{\eta}{b}H_{1}\right\Vert<0$.

%%%%%%%%%%%%%%%%%%%%%%%%%%%%%%%%%%%%%

%%%%%%%%%%%%%%%%%%%%%%%%%%%%%%%%%%%%%%%%%%%%%%%%%%%%%%%%%%%%%%%%%%%%%%%%%%%%%
\subsection{Stochastic Gradient Descent with i.i.d. Stepsizes}\label{sec:iid}

In this section, we consider the stochastic gradient descent method with i.i.d. stepsizes. 
We first observe that SGD  \eqref{eq-stoc-grad} is an iterated random recursion of the form
\begin{equation} 
x_k = \Psi(x_{k-1},\Omega_k,\eta_{k}),
\end{equation} 
where the map $\Psi:\mathbb{R}^d\times \mathcal{S}\times\mathbb{R}_{+} \to \mathbb{R}^d$, $\mathcal{S}$ denotes the set of all subsets of $\{1,2,\dots,n\}$ and $\Omega_k$ is random and i.i.d. 
When the stepsize $\eta_{k}$ are i.i.d., if we write $\Psi_{\Omega,\eta}(x) = \Psi(x,\Omega,\eta)$ 
for simplicity where $(\Omega,\eta)$ has the same distribution as $(\Omega_k,\eta_{k})$, 
and assume that the random map $\Psi_{\Omega,\eta}$ is Lipschitz on average, i.e.  
$\mathbb{E}[L_{\Omega,\eta}] < \infty$ with $L_{\Omega,\eta} := \sup\nolimits_{x,y\in\mathbb{R}^d} \frac{\|\Psi_{\Omega,\eta}(x) - \Psi_{\Omega,\eta}(y) \|}{\|x-y\|}$,
and is mean-contractive, i.e. if $\mathbb{E}\log(L_{\Omega,\eta})<0$ then it can be shown under further technical assumptions that the distribution of the iterates converges to a unique stationary distribution $x_\infty$ geometrically fast \citep{diaconis1999iterated}. 
We have the following result that characterizes the tail-index under such assumptions for dimension $d=1$, which can be derived from \citet{mirek2011heavy} by adapting it to our setting (see also \citet{buraczewski2016stochastic}. %we refer the readers to \citet{mirek2011heavy} for general $d$).

\begin{theorem}[Adaptation of \citet{mirek2011heavy}]\label{thm:nonlinear:iid}
Assume stationary solution to $$x_k = \Psi_{\Omega_k,\eta_{k}}(x_{k-1})$$ exists and:
(i) There exists a random matrix $M(\Omega,\eta)$ and a random variable $B(\Omega,\eta)>0$ such that for a.e. $\Omega,\eta$, 
$|\Psi_{\Omega,\eta}(x)-M(\Omega,\eta) x | \leq B(\Omega,\eta)$ for every $x$;
(ii) The conditional law of $\log |M(\Omega,\eta)|$ given $M(\Omega,\eta)\neq 0$ is non-arithmetic ;i.e. \mgrev{its support is not equal to $a\mathbb{Z}$ for any scalar $a$ where $\mathbb{Z}$ is the set of integers}.
(iii) There exists $\alpha>0$ such that $\mathbb{E} |M(\Omega,\eta)|^\alpha = 1$,  $\mathbb{E} |B(\Omega,\eta)|^\alpha <\infty$ and 
$\mathbb{E} [|M(\Omega,\eta)|^\alpha \log^+ |M(\Omega,\eta)|]< \infty$,
where $\log^+(x) := \max(\log(x),0)$. Then, it holds that $\lim_{t \to \infty } t^{\alpha} \mathbb{P}(|x_\infty|>t) = c_0$ for some constant $c_0>0$.  
\end{theorem}

Theorem~\ref{thm:nonlinear:iid} shows that heavy tails arises for general losses that has an almost linear growth outside compact sets, however it does not characterize how the tail-index $\alpha$ depends on the stepsize, furthermore it is highly non-trivial how to verify its assumptions in general. Also, it works only in the one dimensional setting; \citet{mirek2011heavy} studies more general $d$ but requires the matrices $M(\Omega,\eta)$ form an orthogonal group which is not satisfied by SGD in general. This motivates us to  study more structured losses in high dimensional settings where more insights can be obtained. 
We next study quadratic $f$ which corresponds to linear regression to obtain finer characterizations. In this case, we have the iterates:
\begin{equation} 
x_{k+1} = \left(I - \frac{\eta_{k+1}}{b} H_{k+1}\right) x_k  + q_{k+1}\,,
\label{eq-stoc-grad-2-iid}
\end{equation} 
where $H_k := \sum_{i\in \Omega_{k} } a_i a_i^T$ are i.i.d. Hessian matrices 
and $q_k := \frac{\eta_{k}}{b} \sum_{i\in \Omega_{k} } y_i$,
and $\eta_{k}$ are i.i.d. with a distribution supported
on an interval $[\eta_{l},\eta_{u}]$, where $\eta_{u}>\eta_{l}>0$.
Under some mild conditions, by following the same arguments as in \cite{HTP2021}, 
$x_{k}$ converges to $x_{\infty}$ in distribution,
where $x_{\infty}$ exhibits the heavy-tail behavior 
with the tail-index $\alpha$ which is the unique positive
value such that $h(\alpha)=1$, where
\begin{equation}\label{h:s:eqn}
h(s) := \lim\nolimits_{k\to\infty}\left(\mathbb{E}\| M_k M_{k-1}\dots M_1\|^s\right)^{1/k},
\end{equation}
provided that
\begin{equation}\label{def-rho-iid}
\rho:= \lim\nolimits_{k\to\infty} (2k)^{-1} \log\left(\mbox{largest eigenvalue of } \Pi_k^T \Pi_k\right)<0,
\end{equation}
where $\Pi_k := M_k M_{k-1}\dots M_1$.
% We have the following result.

Similar to the SGD with constant stepsize case (Theorem~\ref{thm:main}), we have the following result that states that the iterations converge to a stationary distribution with heavy tails.

\begin{theorem}\label{thm:iid}
Consider the SGD iterations with i.i.d. stepsizes \eqref{eq-stoc-grad-2-iid}.
If $\rho<0$ and there exists a unique positive $\alpha$ such that $h(\alpha)=1$, 
where $h$ and $\rho$ are defined in \eqref{h:s:eqn}-\eqref{def-rho-iid},
then \eqref{eq-stoc-grad-2-iid} admits a unique stationary solution $x_{\infty}$
and the SGD iterations with cyclic stepsizes converge to $x_{\infty}$ in distribution,
where the distribution of $x_{\infty}$ satisfies
\begin{equation} 
\lim\nolimits_{t\to\infty} t^{\alpha} \mathbb{P}\left(u^T x_{\infty} > t \right)= e_{\alpha}(u)\,, \quad u\in\mathbb{S}^{d-1}\,,
\end{equation}
for some positive and continuous function $e_{\alpha}$ on 
$\mathbb{S}^{d-1}$.
\end{theorem}

%%%%%%%%%%%%%%%%%%%%%%%%%%%%%%%
Theorem~\ref{thm:iid} says the tail-index $\alpha$ is the unique positive
value such that $h(\alpha)=1$ provided that
$\rho<0$. However, the expressions of $h(s)$ and $\rho$ are not very explicit.
Under Assumption~\textbf{(A3)}, we can simplify the expressions for $h(s)$ and $\rho$ (see Lem.~\ref{lem:alternative:expression} and Lem.~\ref{lem:simplified} in the Appendix).
Moreover, under Assumption~\textbf{(A3)}, we have the following result
which characterizes the dependence of the tail-index $\alpha$ on the batch-size
and the dimension.

%%%%%%%%%%%%%%%%%%%%%%%%%%%%%
% Under Assumption~\textbf{(A3)}, 
% we can further simplify the expressions for $\tilde{h}(s)$ and $\tilde{\rho}$ in Lemma~\ref{lem:alternative:expression} as follows.

\begin{theorem}\label{thm:mono:Gaussian}
Assume \textbf{(A3)} holds and $\rho<0$.
%$\tilde{\rho}<0$ where $\tilde{\rho}$ is defined in \eqref{tilde:rho:defn}.
Then we have: (i) the tail-index $\alpha$ is strictly increasing in batch-size $b$
provided $\alpha\geq 1$. 
(ii) The tail-index $\alpha$ is strictly decreasing in dimension $d$.
\end{theorem}

%We showed in Theorem~\ref{thm:mono:Gaussian} that
%the tail-index $\alpha$ is strictly less than the tail-index $\alpha_{c}$
%with constant stepsize $\mathbb{E}[\eta]$ provided that $\alpha\geq 1$.
%The previous result shows that randomized stepsize $\eta$ leads to heavier stepsize compared to a constant stepsize with value $\mathbb{E}\eta$. 
In Theorem~\ref{thm:mono:Gaussian}, we showed that
that smaller batch-sizes lead to (smaller tail-index) heavier tail provided that $\alpha\geq 1$
and higher dimension leads to (smaller tail-index) heavier tail. 
%We recall that the stepsizes $\eta_{k}$ are i.i.d. supported on an interval $[\eta_{l},\eta_{u}]$ with %$\eta_{u}>\eta_{l}>0$.  Next, we compare the tail-index $\alpha$ with the tail-index
%$\alpha_{u}$ and $\alpha_{l}$ that corresponds
%to the SGD with constant stepsizes $\eta_{u}$ and $\eta_{l}$ respectively.
%
%\begin{proposition}\label{prop:const:boundary}
%Assume \textbf{(A3)} holds. 
%Then we have $\alpha_{u}\leq\alpha\leq\alpha_{l}$ provided
%that $\alpha_{u}\geq 1$.
%\end{proposition}
%
%Proposition~\ref{prop:const:boundary} shows
%that the SGD with constant stepsize $\eta_{u}$ 
%has heavier tail than the SGD with i.i.d stepsizes 
%supported on an interval $[\eta_{l},\eta_{u}]$, 
%which has heavier tail than the SGD with constant stepsize $\eta_{l}$. 
On the other hand, it is also natural to conjecture that the tail-index gets smaller 
if the distribution of $\eta$ is more spread out. 
The formalize our intuition, we assume that
the stepsize is uniformly distributed with mean $\bar{\eta}$
and range $R$, i.e. the stepsize is uniformly 
distributed on the interval $(\bar{\eta}-R,\bar{\eta}+R)$. 
Next, we show that the tail-index decreases
as the range $R$ increases provided the tail-index $\alpha$ is greater than $1$.

\begin{theorem}\label{thm:mono:Gaussian:range}
Assume \textbf{(A3)} holds and $\rho<0$.
Assume $\eta$ is uniformly distributed on $(\bar{\eta}-R,\bar{\eta}+R)$.
Then, the tail-index $\alpha$ is decreasing in the range $R$ provided that $\alpha\geq 1$. 
\end{theorem}

%%%%%%%%%%%%%%%%%%%%%%%%%%%%%%%%%%%%%%%%%%%%%%%%%%%%%%%%%%
Under Assumption \textbf{(A3)}, 
our next result characterizes the tail-index $\alpha$ depending on the choice of the batch-size $b$, the variance $\sigma^2$, which determines the curvature around the minimum and the stepsize; in particular we provide critical threshold such that the stationary distribution will become heavy tailed with an infinite variance.

\begin{proposition}\label{alpha:2}
Assume \textbf{(A3)} holds.
Define 
\begin{equation}
c:=1-2\mathbb{E}[\eta]\sigma^{2}+\frac{\mathbb{E}[\eta^{2}]\sigma^{4}}{b}(d+b+1).
\end{equation}
The following holds: (i) There exists $\delta>0$ such that
for any $1<c<1+\delta$, tail-index $0<\alpha< 2$. (ii) If $c=1$, tail-index $\alpha=2$. (iii) If $c<1$, then tail-index $\alpha>2$.
%Assume \textbf{(A3)} holds.
%Let $\eta_{crit} = \frac{2b}{\sigma^{2}(d+b+1)}$.
%The following holds: (i) There exists $\eta_{max}>\eta_{crit}$ such that
%for any $\eta_{crit}<\frac{\mathbb{E}[\eta^{2}]}{\mathbb{E}[\eta]}<\eta_{max}$, tail-index $0<\alpha< 2$. (ii) If $\frac{\mathbb{E}[\eta^{2}]}{\mathbb{E}[\eta]}= \eta_{crit}$, tail-index $\alpha=2$. (iii) If $\frac{\mathbb{E}[\eta^{2}]}{\mathbb{E}[\eta]}\in (0,\eta_{crit})$, then tail-index $\alpha>2$.
\end{proposition}

Theorem~\ref{thm:iid} is of asymptotic nature which characterizes the stationary distribution $x_{\infty}$ of
SGD iterations 
with a tail-index $\alpha$. 
Next, we provide non-asymptotic moment
bounds for $x_{k}$ at each $k$-th iterate for $p$ such that $h(p)<1$.

\begin{lemma}\label{lem:finite:bound:iid}
Assume \textbf{(A3)} holds.

(i) For any $p\leq 1$ and $h(p)<1$,
\begin{equation}
\mathbb{E}\Vert x_{k}\Vert^{p}
\leq
(h(p))^{k}\mathbb{E}\Vert x_{0}\Vert^{p}
+\frac{1-(h(p))^{k}}{1-h(p)}\mathbb{E}\Vert q_{1}\Vert^{p}.
\end{equation}

(ii) For any $p>1$, $\epsilon>0$ and $(1+\epsilon)h(p)<1$,
\begin{equation}
\mathbb{E}\Vert x_{k}\Vert^{p}
\leq
((1+\epsilon)h(p))^{k}\mathbb{E}\Vert x_{0}\Vert^{p}
+\frac{1-((1+\epsilon)h(p))^{k}}{1-(1+\epsilon)h(p)}
\frac{(1+\epsilon)^{\frac{p}{p-1}}-(1+\epsilon)}{\left((1+\epsilon)^{\frac{1}{p-1}}-1\right)^{p}}
\mathbb{E}\Vert q_{1}\Vert^{p}.
\end{equation}
\end{lemma}
%%%%%%%%%%%%%%%%%%%%%%%%%%%%%%%%%%%%%%%%%%%%%%%%%%%%%%%%%%%

Next, we will study the speed of convergence of the $k$-th iterate $x_{k}$ to its stationary distribution $x_{\infty}$ in the Wasserstein metric $\mathcal{W}_{p}$ for any $p$ such that $h(p)<1$. 

\begin{theorem}\label{thm:convergence:iid}
Assume \textbf{(A3)} holds.
Let $\nu_{k}$, $\nu_{\infty}$ denote the probability laws
of $x_{k}$ and $x_{\infty}$ respectively. 
Then  
\begin{equation}
\mathcal{W}_{p}(\nu_{k},\nu_{\infty})
\leq
\left(h(p)\right)^{k/p}
\mathcal{W}_{p}(\nu_{0},\nu_{\infty}),
\end{equation}
for any $p\geq 1$ and $h(p)<1$,
where the convergence rate $(h(p))^{1/p}\in(0,1)$.
\end{theorem}
%%%%%%%%%%%%%%%%%%%%%%%%%%%%%%
When the tail-index $\alpha>2$, by Lemma~\ref{lem:finite:bound:iid}, the second moments of the iterates $x_k$ are finite, in which case central limit theorem (CLT) says that if the cumulative sum of the iterates $S_{K}:=\sum_{k=1}^{K} x_{k}$ 
is scaled properly, the resulting distribution is Gaussian. In the case where $\alpha<2$, the variance of the iterates is not finite; however in this case, we derive the following generalized CLT (GCLT) which says if the iterates are properly scaled, the limit will be an $\alpha$-stable  distribution. This is stated in a more precise manner as follows.

\begin{corollary}\label{cor:clt}
Assume \textbf{(A3)} holds and the conditions of Theorem~\ref{thm:iid} are satisfied. Then, we have the following:

(i) If  $\alpha \in(0,1) \cup(1,2)$, then there is a sequence $d_{K}=d_{K}(\alpha)$  and a function $C_{\alpha}: \mathbb{S}^{d-1} \mapsto \mathbb{C}$  such that as $K\rightarrow\infty$ the random variables $K^{-\frac{1}{\alpha}}\left(S_{K}-d_{K}\right)$ converge in law to the $\alpha$-stable random variable with characteristic function
$\Upsilon_{\alpha}(tv)=\exp(t^{\alpha}C_{\alpha}(v))$, for $t>0$ and $v \in \mathbb{S}^{d-1}$.

(ii) If $\alpha=1$, then there are functions $\xi, \tau:(0, \infty) \mapsto \mathbb{R}$ and $C_{1}: \mathbb{S}^{d-1} \mapsto \mathbb{C}$ such that as $K\rightarrow\infty$ the random variables $K^{-1} S_{K}-K \xi\left(K^{-1}\right)$ converge in law to the random variable with characteristic function
$\Upsilon_{1}(t v)=\exp \left(t C_{1}(v)+i t\langle v, \tau(t)\rangle\right)$, 
for $t>0$ and $v \in \mathbb{S}^{d-1}$.

(iii) If \(\alpha=2,\) then there is a sequence \(d_{K}=d_{K}(2)\) and a function \(C_{2}: \mathbb{S}^{d-1} \mapsto \mathbb{R}\) such that as $K\rightarrow\infty$
the random variables \((K \log K)^{-\frac{1}{2}}\left(S_{K}-d_{K}\right)\) converge in law to the random variable with
characteristic function
$\Upsilon_{2}(t v)=\exp \left(t^{2} C_{2}(v)\right)$, for $t>0$ and $v \in \mathbb{S}^{d-1}$.

(iv) If \(\alpha \in(0,1),\) then \(d_{K}=0,\) and if \(\alpha \in(1,2],\) then \(d_{K}=K \bar{x},\) where \(\bar{x}=\int_{\mathbb{R}^{d}} x \nu_\infty(d x) .\) 
\end{corollary}

In addition to its evident theoretical interest, Corollary~\ref{cor:clt} has also an important practical implication: estimating the tail-index of a \emph{generic} heavy-tailed distribution is a challenging problem (see e.g. \cite{clauset2009power,goldstein2004problems,bauke-power-law}); however, for the specific case of $\alpha$-stable distributions, accurate and computationally efficient estimators, which \emph{do not} require the knowledge of the functions $C_\alpha$, $\tau$, $\xi$, have been proposed \citep{mohammadi2015estimating}. Thanks to Corollary~\ref{cor:clt}, we will be able to use such estimators in our numerical experiments in Section~\ref{sec:numerical}.

%%%%%%%%%%%%%%%%%%%%%%%%%%%%%%
\subsection{Technical Results for SGD with Cyclic Stepsizes}\label{sec:cyclic:appendix}

In this section, we provide some additional technical results for SGD with cyclic stepsizes.

If we assume that the random map $\Psi^{(m)}$ is Lipschitz on average, i.e.  $\mathbb{E}\left[L^{(m)}\right] < \infty$ with $L^{(m)} := \sup\nolimits_{x,y\in\mathbb{R}^d} \frac{\|\Psi^{(m)}(x) - \Psi^{(m)}(y) \|}{\|x-y\|}$,
and is mean-contractive, i.e. if $\mathbb{E}\log\left(L^{(m)}\right)<0$ then it can be shown under further technical assumptions that the iterates converges to a unique stationary distribution $x_\infty$ geometrically fast \citep{diaconis1999iterated}. %Similar to Theorem~\ref{thm:nonlinear:iid}, 

First, we have the following analogue of Theorem~\ref{thm:nonlinear:Markovian} which 
is a special case of Theorem~\ref{thm:nonlinear:Markovian}.

\begin{theorem}[Adaptation of \citet{mirek2011heavy} ]\label{thm:nonlinear:cyclic}
Assume stationary solution to \eqref{cyclic:dynamics:nonlinear} exists and: %. We further assume 
(i) There exists a random \mgrev{variable} %matrix 
$M^{(m)}$ and a random variable $B^{(m)}>0$ such that a.s. 
$|\Psi^{(m)}(x)-M^{(m)} x | \leq B^{(m)}$ for every $x$;
(ii) The conditional law of $\log |M^{(m)}|$ given $M^{(m)}\neq 0$ is non-arithmetic; i.e. \mgrev{its support is not equal to $a\mathbb{Z}$ for any scalar $a$ where $\mathbb{Z}$ is the set of integers}.
(iii) There exists $\alpha^{(m)}>0$ such that $\mathbb{E} [|M^{(m)}|^{\alpha^{(m)}}] = 1$,  $\mathbb{E} [|B^{(m)}|^{\alpha^{(m)}}] <\infty$ and 
$\mathbb{E}[|M^{(m)}|^{\alpha^{(m)}} \log^+ |M^{(m)}|]< \infty$,
where $\log^+(x) := \max(\log(x),0)$.
Then there exists some constant $c_0^{(m)}>0$ such that 
$\lim_{t \to \infty } t^{\alpha^{(m)}} \mathbb{P}(|x_\infty|>t) = c_0^{(m)}$ .  
\end{theorem}

{\color{black}Next, we consider the setting of the linear regression. We can iterate the SGD from \eqref{eq-stoc-grad-2} to obtain
$x_{(k+1)m}=M_{k+1}^{(m)}x_{km}+q_{k+1}^{(m)}$,
where $M_{k+1}^{(m)}$ is defined in \eqref{M:k:m:defn}
and $q_{k+1}^{(m)}:=\sum_{i=km+1}^{(k+1)m}
\left(I - \frac{\eta_{(k+1)m}}{b} H_{(k+1)m}\right)
\left(I - \frac{\eta_{(k+1)m-1}}{b} H_{(k+1)m-1}\right)\cdots \left(I - \frac{\eta_{i+1}}{b} H_{i+1}\right)q_{i}$. We showed in Theorem~\ref{thm:cyclic}
that $x_{\infty}$ has heavy tails with a tail-index $\alpha^{(m)}$ and further properties of the tail-index $\alpha^{(m)}$ were obtained under Assumption \textbf{(A3)} in Theorem~\ref{thm:mono:Gaussian:cyclic}.}

Under Assumption \textbf{(A3)}, 
our next result characterizes the tail-index $\alpha^{(m)}$ depending on the choice of the batch-size $b$, the variance $\sigma^2$, which determines the curvature around the minimum and the stepsize; in particular we provide critical threshold such that the stationary distribution will become heavy tailed with an infinite variance.

\begin{proposition}\label{alpha:2:cyclic}
Assume \textbf{(A3)} holds.
Define 
\begin{equation}
c^{(m)}:=\prod_{i=1}^{m}\left(1-2\eta_{i}\sigma^{2}+\frac{\eta_{i}^{2}\sigma^{4}}{b}(d+b+1)\right).
\end{equation}
The following holds: (i) There exists $\delta>0$ such that
for any $1<c^{(m)}<1+\delta$, tail-index $0<\alpha^{(m)}< 2$. (ii) If $c^{(m)}=1$, tail-index $\alpha^{(m)}=2$. (iii) If $c^{(m)}<1$, then tail-index $\alpha^{(m)}>2$.
\end{proposition}

In Section~\ref{sec:main:results}, Theorem~\ref{thm:cyclic} is of asymptotic nature which characterizes the stationary distribution $x_{\infty}$ of
SGD iterations 
with a tail-index $\alpha^{(m)}$. 
Next, we provide non-asymptotic moment
bounds for $x_{mk}$ at each $mk$-th iterate
for $p$ such that $h^{(m)}(p)<1$.

\begin{lemma}\label{lem:finite:bound:cyclic}
Assume \textbf{(A3)} holds.

(i) For any $p\leq 1$ and $h^{(m)}(p)<1$,
\begin{equation}
\mathbb{E}\Vert x_{mk}\Vert^{p}
\leq
\left(h^{(m)}(p)\right)^{k}\mathbb{E}\Vert x_{0}\Vert^{p}
+\frac{1-(h^{(m)}(p))^{k}}{1-h^{(m)}(p)}\mathbb{E}\left\Vert q_{1}^{(m)}\right\Vert^{p}.
\end{equation}

(ii) For any $p>1$, $\epsilon>0$ and $(1+\epsilon)h^{(m)}(p)<1$,
\begin{align}
\mathbb{E}\Vert x_{mk}\Vert^{p}
&\leq
\left((1+\epsilon)h^{(m)}(p)\right)^{k}\mathbb{E}\Vert x_{0}\Vert^{p}
+\frac{1-((1+\epsilon)h^{(m)}(p))^{k}}{1-(1+\epsilon)h^{(m)}(p)}
\frac{(1+\epsilon)^{\frac{p}{p-1}}-(1+\epsilon)}{\left((1+\epsilon)^{\frac{1}{p-1}}-1\right)^{p}}
\mathbb{E}\left\Vert q_{1}^{(m)}\right\Vert^{p}.
\end{align}
\end{lemma}

%%%%%%%%%%%%%%%%%%%%%%%%%%%%%%%%%%%%%%%%%%%%%%%%%%%%%%%%%%%

Next, we will study the speed of convergence of the $mk$-th iterate $x_{mk}$ to its stationary distribution $x_{\infty}$ in the Wasserstein metric $\mathcal{W}_{p}$ for any $p$ such that $h^{(m)}(p)<1$. 

\begin{theorem}\label{thm:convergence:cyclic}
Assume \textbf{(A3)} holds.
Let $\nu_{mk}$, $\nu_{\infty}$ denote the probability laws
of $x_{mk}$ and $x_{\infty}$ respectively. 
Then  
\begin{equation}
\mathcal{W}_{p}(\nu_{mk},\nu_{\infty})
\leq
\left(h^{(m)}(p)\right)^{k/p}
\mathcal{W}_{p}(\nu_{0},\nu_{\infty}),
\end{equation}
for any $p\geq 1$ and $h^{(m)}(p)<1$,
where the convergence rate $\left(h^{(m)}(p)\right)^{1/p}\in(0,1)$.
\end{theorem}

%%%%%%%%%%%%%%%%%%%%%%%%%%%%%%%%%%%%%%%%%%%%%%%%%%%%%%%%%%%%%
%Finally, let us remark that when the stepsizes are cyclic, 
%they are of deterministic fashion and if we run
%gradient descent with deterministic gradients, 
%there will be no heavy-tail phenomenon in this setup.

%%%%%%%%%%%%%%%%%%%%%%%%%%%%%%%%%%%%%%%%%%%%%

%%%%%%%%%%%%%%%%%%%%%%%%%%%%%%
Similar as in Corollary~\ref{cor:clt}, 
we have the following generalized CLT (GCLT)
result for
$S_{K}^{(m)}:=\sum_{k=1}^{K} x_{mk}$ 
when it is scaled properly so that the limit will be an alpha-stable  distribution. 

\begin{corollary}\label{cor:clt:cyclic}
Assume \textbf{(A3)} holds and the conditions of Theorem~\ref{thm:cyclic} are satisfied. Then, we have the following:

(i) If  $\alpha^{(m)} \in(0,1) \cup(1,2)$, then there is a sequence $d_{K}=d_{K}(\alpha^{(m)})$  and a function $C_{\alpha^{(m)}}: \mathbb{S}^{d-1} \mapsto \mathbb{C}$  such that as $K\rightarrow\infty$ the random variables $K^{-\frac{1}{\alpha^{(m)}}}\left(S_{K}^{(m)}-d_{K}\right)$ converge in law to the $\alpha^{(m)}$-stable random variable with characteristic function
$\Upsilon_{\alpha^{(m)}}(tv)=\exp(t^{\alpha^{(m)}}C_{\alpha^{(m)}}(v))$, for $t>0$ and $v \in \mathbb{S}^{d-1}$.

(ii) If $\alpha^{(m)}=1$, then there are functions $\xi, \tau:(0, \infty) \mapsto \mathbb{R}$ and $C_{1}: \mathbb{S}^{d-1} \mapsto \mathbb{C}$ such that as $K\rightarrow\infty$ the random variables $K^{-1} S_{K}^{(m)}-K \xi\left(K^{-1}\right)$ converge in law to the random variable with characteristic function
$\Upsilon_{1}(t v)=\exp \left(t C_{1}(v)+i t\langle v, \tau(t)\rangle\right)$, 
for $t>0$ and $v \in \mathbb{S}^{d-1}$.

(iii) If \(\alpha^{(m)}=2,\) then there is a sequence \(d_{K}=d_{K}(2)\) and a function \(C_{2}: \mathbb{S}^{d-1} \mapsto \mathbb{R}\) such that as $K\rightarrow\infty$
the random variables \((K \log K)^{-\frac{1}{2}}\left(S_{K}^{(m)}-d_{K}\right)\) converge in law to the random variable with
characteristic function
$\Upsilon_{2}(t v)=\exp \left(t^{2} C_{2}(v)\right)$, for $t>0$ and $v \in \mathbb{S}^{d-1}$.

(iv) If \(\alpha^{(m)} \in(0,1),\) then \(d_{K}=0,\) and if \(\alpha^{(m)} \in(1,2],\) then \(d_{K}=K \bar{x},\) where \(\bar{x}=\int_{\mathbb{R}^{d}} x \nu_\infty(d x) .\) 
\end{corollary}

For the specific case of $\alpha$-stable distributions, accurate and computationally efficient estimators, which \emph{do not} require the knowledge of the functions $C_\alpha$, $\tau$, $\xi$, have been proposed \citep{mohammadi2015estimating}. Thanks to Corollary~\ref{cor:clt:cyclic}, we will be able to use such estimators in our numerical experiments in Section~\ref{sec:numerical}.

%%%%%%%%%%%%%%%%%%%%%%%%%%%%%%%%%%%%%%%%%%%%%%%%%%%%%%%%%%%%%%%%%%%%%%%%
\subsection{Technical Results for SGD with Markovian Stepsizes}\label{subsec-appendix-markovian}

In this section, we provide some additional technical results for SGD with Markovian stepsizes.
In Section~\ref{sec:main:results}, we restricted our discussions to the finite state space.
In the following section, we provide some technical results for the general state space.

\subsubsection{Markovian Stepsizes with General State Space}\label{sec:general:state:Markov}

When the objective is quadratic, we recall that the iterates of the SGD are given by:
\begin{equation}\label{iterates:Markovian}
x_{k+1}=M_{k+1}x_{k}+q_{k+1}.
\end{equation}
In this case, $M_{k}=I-\frac{\eta_{k}}{b}H_{k}$, where $\eta_{k}$ is a stationary Markov chain with a common distribution $\eta$ supported
on an interval $[\eta_{l},\eta_{u}]$, where $\eta_{u}>\eta_{l}>0$, 
and $H_{k}$ are i.i.d. Hessian matrices. 

To the best of our knowledge, there is no general stochastic linear recursion theory
for Markovian coefficients, except for some special cases, e.g. with heavy-tail coefficient \citep{hay2011multivariate}.
Nevertheless, using a direct approach, we can obtain a lower bound
for the tail-index for the limit of
the SGD with Markovian stepsizes as follows.
Since $\eta_{k}$ is stationary and $H_{k}$ are i.i.d., $M_{k}$ is stationary, we have:
\begin{equation}\label{hat:h:defn}
h^{(r)}(s)\leq\hat{h}^{(g)}(s):=\mathbb{E}\left[\Vert M_{1}\Vert^{s}\right]=\mathbb{E}\left[\left\Vert I-\frac{\eta_{1}}{b}H_{1}\right\Vert^{s}\right],
\qquad\text{for any $s\geq 0$},
\end{equation}
where $\hat{h}^{(g)}(s)$ is an upper bound on $h^{(r)}(s)$ (defined in \eqref{def-hs-Markov}) %\eqref{h:s:eqn})
and we also define
\begin{equation}\label{hat:rho:defn}
\hat{\rho}^{(g)}:=\mathbb{E}\left[\log\Vert M_{1}\Vert\right]=\mathbb{E}\left[\log\left\Vert I-\frac{\eta_{1}}{b}H_{1}\right\Vert\right].
\end{equation}

While having a grasp of the exact value of the tail-index for the stationary distribution of $x_{\infty}$ is difficult when the stepsizes are Markovian, in the next result, based on a technical lemma (Lem.~\ref{lem:finite:bound} in the Appendix) for the moment bounds for $x_{k}$, we can characterize a lower bound $\hat\alpha^{(g)}$ for the tail-index to control how heavy tailed SGD iterates can be, 
in the sense that we have $\mathbb{P}(\Vert x_\infty\Vert>t) < C_{p}/t^{p}$ for some constant $C_{p}$ 
as long as $p<\hat\alpha^{(g)}$. %decays $x_\infty$ polynomially but the decay rate cannot be slower than $\hat{\alpha}$.
% the heavy-tailedness
%{\color{red} Explain What is the value of next corollary?}

% \begin{lemma}\label{lem:finite:bound}
% (i) For any $p\leq 1$ and $\hat{h}(p)<1$,
% \begin{equation}
% \mathbb{E}\Vert x_{k}\Vert^{p}
% \leq
% (\hat{h}(p))^{k}\mathbb{E}\Vert x_{0}\Vert^{p}
% +\frac{1-(\hat{h}(p))^{k}}{1-\hat{h}(p)}\mathbb{E}\Vert q_{1}\Vert^{p}.
% \end{equation}

% (ii) For any $p>1$, $\epsilon>0$ and $(1+\epsilon)\hat{h}(p)<1$,
% \begin{equation}
% \mathbb{E}\Vert x_{k}\Vert^{p}
% \leq
% \left((1+\epsilon)\hat{h}(p)\right)^{k}\mathbb{E}\Vert x_{0}\Vert^{p}
% +\frac{1-((1+\epsilon)\hat{h}(p))^{k}}{1-(1+\epsilon)\hat{h}(p)}
% \frac{(1+\epsilon)^{\frac{p}{p-1}}-(1+\epsilon)}{\left((1+\epsilon)^{\frac{1}{p-1}}-1\right)^{p}}
% \mathbb{E}\Vert q_{1}\Vert^{p}.
% \end{equation}
% \end{lemma}

\begin{proposition}\label{cor:Markovian} 
Let $\hat\alpha^{(g)}$ be the unique positive value such that $\hat{h}^{(g)}(\hat{\alpha}^{(g)})=1$, provided that $\hat{\rho}^{(g)}<0$, where $\hat{h}^{(g)}$ and $\hat{\rho}^{(g)}$ are defined in \eqref{hat:h:defn}-\eqref{hat:rho:defn}.
Then, for any $p\leq 1$ and $\hat{h}^{(g)}(p)<1$,\
\begin{equation}
\mathbb{P}(\Vert x_{\infty}\Vert\geq t)
\leq
\frac{1}{1-\hat{h}^{(g)}(p)}\frac{\mathbb{E}\Vert q_{1}\Vert^{p}}{t^{p}},\qquad\text{for any $t>0$},
\end{equation}
and for any $p>1$, $\epsilon>0$ and $(1+\epsilon)\hat{h}^{(g)}(p)<1$,
\begin{equation}
\mathbb{P}(\Vert x_{\infty}\Vert\geq t)
\leq
\frac{1}{1-(1+\epsilon)\hat{h}^{(g)}(p)}
\frac{(1+\epsilon)^{\frac{p}{p-1}}-(1+\epsilon)}{\left((1+\epsilon)^{\frac{1}{p-1}}-1\right)^{p}}
\frac{\mathbb{E}\Vert q_{1}\Vert^{p}}{t^{p}},\qquad\text{for any $t>0$},
\end{equation}
\end{proposition}

%%%%%%%%%%%%%%%%%%%%%%%%%%%%%%%%%%%%%%%%%%%%%%%%%%%%%%%%%%%

Next, in the following result, we discuss how the tail-index (lower bound) estimate $\hat\alpha^{(g)}$ depends
on the batch-size and how it compares with the tail-index (lower bound) estimate $\hat{\alpha}_{c}$ with constant stepsize $\mathbb{E}[\eta]$. %lower bound for the tail-index.

% \begin{theorem}\label{thm:convergence}
% Let $\nu_{k}$, $\nu_{\infty}$ denote the probability laws
% of $x_{k}$ and $x_{\infty}$ respectively. 
% Then  
% \begin{equation}
% \mathcal{W}_{p}(\nu_{k},\nu_{\infty})
% \leq
% \left(\hat{h}(p)\right)^{k/p}
% \mathcal{W}_{p}(\nu_{0},\nu_{\infty}),
% \end{equation}
% for any $p\geq 1$ and $\hat{h}(p)<1$,
% where the convergence rate $(\hat{h}(p))^{1/p}\in(0,1)$.
% \end{theorem}

\begin{theorem}\label{thm:mono:Markovian}
(i) The lower bound for the tail-index $\hat\alpha^{(g)}$ is strictly increasing in batch-size $b$
provided that $\hat\alpha^{(g)}\geq 1$. (ii) The lower bound for the tail-index $\hat\alpha^{(g)}$ is strictly less than
the lower bound for the tail-index $\hat{\alpha}_{c}$ with constant stepsize $\mathbb{E}[\eta]$
provided that $\hat\alpha^{(g)}\geq 1$.
\end{theorem}

Under Assumption \textbf{(A3)}, 
our next result characterizes the tail-index $\alpha^{(r)}$ depending on the choice of the batch-size $b$, the variance $\sigma^2$, which determines the curvature around the minimum and the stepsize; in particular we provide critical threshold such that the stationary distribution will become heavy tailed with an infinite variance.

\begin{proposition}\label{alpha:2:Markov}
Assume \textbf{(A3)} holds.
Define 
\begin{equation}
c^{(r)}:=\mathbb{E}\left[\prod_{i=1}^{r_{1}}\left(1-2\eta_{i}\sigma^{2}+\frac{\eta_{i}^{2}\sigma^{4}}{b}(d+b+1)\right)\right].
\end{equation}
The following holds: (i) There exists $\delta>0$ such that
for any $1<c^{(r)}<1+\delta$, tail-index $0<\alpha^{(r)}< 2$. (ii) If $c^{(r)}=1$, tail-index $\alpha^{(r)}=2$. (iii) If $c^{(r)}<1$, then tail-index $\alpha^{(r)}>2$.
\end{proposition}

In Section~\ref{sec:main:results}, Theorem~\ref{thm:Markov:regeneration} is of asymptotic nature which characterizes the stationary distribution $x_{\infty}$ of
SGD iterations 
with a tail-index $\alpha^{(r)}$. 
Next, we provide non-asymptotic moment
bounds for the finite iterates
when $\hat{h}^{(g)}(p)<1$, where 
we recall that the definition of $\hat{h}^{(g)}(s)$ from \eqref{hat:h:defn}. 

\begin{lemma}\label{lem:finite:bound}
(i) For any $p\leq 1$ and $\hat{h}^{(g)}(p)<1$,
\begin{equation}
\mathbb{E}\Vert x_{k}\Vert^{p}
\leq
\left(\hat{h}^{(g)}(p)\right)^{k}\mathbb{E}\Vert x_{0}\Vert^{p}
+\frac{1-(\hat{h}^{(g)}(p))^{k}}{1-\hat{h}^{(g)}(p)}\mathbb{E}\Vert q_{1}\Vert^{p}.
\end{equation}

(ii) For any $p>1$, $\epsilon>0$ and $(1+\epsilon)\hat{h}^{(g)}(p)<1$,
\begin{equation}
\mathbb{E}\Vert x_{k}\Vert^{p}
\leq
\left((1+\epsilon)\hat{h}^{(g)}(p)\right)^{k}\mathbb{E}\Vert x_{0}\Vert^{p}
+\frac{1-((1+\epsilon)\hat{h}^{(g)}(p))^{k}}{1-(1+\epsilon)\hat{h}^{(g)}(p)}
\frac{(1+\epsilon)^{\frac{p}{p-1}}-(1+\epsilon)}{\left((1+\epsilon)^{\frac{1}{p-1}}-1\right)^{p}}
\mathbb{E}\Vert q_{1}\Vert^{p}.
\end{equation}
\end{lemma}

Next, we provide
the convergence rate to the stationary distribution in $p$-Wasserstein distance provided that $\hat{h}^{(g)}(p)<1$.

\begin{theorem}\label{thm:convergence}
Let $\nu_{k}$, $\nu_{\infty}$ denote the probability laws
of $x_{k}$ and $x_{\infty}$ respectively. 
Then  
\begin{equation}
\mathcal{W}_{p}(\nu_{k},\nu_{\infty})
\leq
\left(\hat{h}^{(g)}(p)\right)^{k/p}
\mathcal{W}_{p}(\nu_{0},\nu_{\infty}),
\end{equation}
for any $p\geq 1$ and $\hat{h}^{(g)}(p)<1$,
where the convergence rate $(\hat{h}^{(g)}(p))^{1/p}\in(0,1)$.
\end{theorem}

%%%%%%%%%%%%%%%%%%%%%%%%%%%%%%%%%%%%%%%%%%%%%%%%%%%%%%%%%%%%%%%%
\subsubsection{Markovian Stepsizes with Finite State Space}
In this section, we provide additional technical results for SGD with Markovian stepsizes
with finite state space. 
%%%%%%%%%%%%%%%%%%%%%%%%%%%%%%%%%%%%%%%%%%%%%
%Similar to Theorem~\ref{thm:nonlinear:cyclic} and Theorem~\ref{thm-mirek}, we have the following result which is an adaptation of \citet{mirek2011heavy}.
%
%\begin{theorem}[Adaptation of \citet{mirek2011heavy}]\label{thm:nonlinear:Markovian}
%Assume stationary solution to \eqref{Markovian:dynamics:nonlinear} exists and:
%
%(i) There exists a random matrix $M^{(r)}$ and a random variable $B^{(r)}>0$ such that a.s. 
%$|\Psi^{(r)}(x)-M^{(r)} x | \leq B^{(r)}$ for every $x$;
%
%(ii) The conditional law of $\log |M^{(r)}|$ given $M^{(r)}\neq 0$ is non-arithmetic;
%
%(iii) There exists $\alpha^{(r)}>0$ such that $\mathbb{E} \left[|M^{(r)}|^{\alpha^{(r)}}\right] = 1$,  $\mathbb{E} \left[|B^{(r)}|^{\alpha^{(r)}}\right] <\infty$ and 
%\begin{equation*}
%\mathbb{E} \left[\left|M^{(r)}\right|^{\alpha^{(r)}} \log^+ \left|M^{(r)}\right|\right]< \infty,
%\end{equation*}
%where $\log^+(x) := \max(\log(x),0)$.
%
%Then, it holds that $\lim_{t \to \infty } t^{\alpha^{(r)}} \mathbb{P}(|x_\infty|>t) = c_0^{(r)}$ for some constant $c_0^{(r)}>0$.  
%\end{theorem}
%%%%%%%%%%%%%%%%%%%%%%%%%%%%%%%%%%%%%%%%
It is natural to conjecture that the tail-index gets smaller 
if the distribution of $\eta$ is more spread out. 
The formalize our intuition, we assume that
the stepsize is uniformly distributed with mean $\bar{\eta}$.
Without loss of generality, we assume that $K$ is an odd number, 
and the stepsizes are equally spaced with distance $\delta>0$ 
in the sense that the state space of the stepsizes is given by
\begin{equation}\label{state:space}
\left\{\bar{\eta},\bar{\eta}\pm\delta,\bar{\eta}\pm 2\delta,\ldots,\bar{\eta}\pm(K-1)\delta/2\right\}.
\end{equation}
Then, the range of the stepsizes is $(K-1)\delta$, which increases as either $\delta$ or $K$ increases.
The stationary distribution of the simple random walk 
is uniform on the set \eqref{state:space}. The following result shows that if the range of stepsizes increases, the tails gets heavier in the sense that tails admit a smaller lower bound $\hat\alpha^{(g)}$,
which is the unique positive value such that $\hat{h}^{(g)}(\hat\alpha^{(g)})=1$, 
where $\hat{h}^{(g)}(s):=\mathbb{E}\left[\Vert M_{1}\Vert^{s}\right]=\mathbb{E}\left[\left\Vert I-\frac{\eta_{1}}{b}H_{1}\right\Vert^{s}\right]$ (see Prop.~\ref{cor:Markovian} in the Appendix for detailed discussions).

\begin{theorem}\label{thm:mono:Markovian:range}
Assume the stationary distribution of the Markovian stepsizes is uniform
on the set \eqref{state:space}.
Then, the lower bound for the tail-index $\hat\alpha^{(g)}$ is decreasing in the range, i.e. decreasing in $\delta$ and $K$, 
provided that $\hat\alpha^{(g)}\geq 1$. 
\end{theorem}

%%%%%%%%%%%%%%%%%%%%%%%%%%%%%%%%%%%%%%%%%%%%%%%%%%%%%%%%%%%%%%%%%

Next, we assume that the range $\frac{K-1}{2}\delta=R$ is fixed,
so that given $K$, we have $\delta=\frac{2R}{K-1}$.
For simplicity, we assume that $K=2^{n}+1$ for some $n\in\mathbb{N}$
such that the state space of the stepsizes is:
\begin{equation}\label{state:space:2}
\left\{\bar{\eta},\bar{\eta}\pm\left(R2^{-(n-1)}\right),\bar{\eta}\pm 2\left(R2^{-(n-1)}\right),\ldots,\bar{\eta}\pm 2^{n-1}\left(R2^{-(n-1)}\right)\right\}.
\end{equation}

Note that the larger the value of $K=2^{n}+1$, the finer the grid for stepsizes is.
We are interested in studying how the lower bound for the tail-index $\hat\alpha^{(g)}$
depends on $K=2^{n}+1$. We have the following result that shows that
the lower bound for the tail-index $\hat\alpha^{(g)}$ is increasing in the $K=2^{n}+1$.

\begin{theorem}\label{thm:mono:Markovian:range:2}
Assume the stationary distribution of the Markovian stepsizes is uniform
on the set \eqref{state:space:2}.
Then, $\hat\alpha^{(g)}$ is increasing in the $K=2^{n}+1$
provided that $\hat\alpha^{(g)}\geq 1$. 
\end{theorem}

This result shows that the finer the grid for stepsizes is, 
the larger the lower bound for the tail-index so that the tail gets lighter, that is, the lower bound on the tail gets lighter.
In Theorem~\ref{thm:mono:Markovian:range:2}, if we write $\hat{\alpha}^{(g)}_{n}:=\hat{\alpha}^{(g)}$ to emphasize the dependence on $n$, then we showed that $\hat{\alpha}^{(g)}_{n}$ is increasing in $n\in\mathbb{N}$. However, we also showed in Theorem~\ref{thm:mono:Markovian} that for any $n\in\mathbb{N}$, $\hat{\alpha}^{(g)}_{n}$ is less than
the lower bound $\hat{\alpha}_{c}$ for the tail-index for the SGD with the constant stepsize $\bar{\eta}$.

%%%%%%%%%%%%%%%%%%%%%%%%%%%%%%%%%%%%%%%%%%%%%%%%%%%%%%%%%%%%
The following result shows that Markovian stepsizes in fact can lead to heavier tails (in the sense of lower bound for the tail-index $\hat\alpha^{(g)}$ values) compared to cyclic stepsizes.

\begin{proposition}\label{thm:Markovian:cyclic} 
Assume the stationary distribution of the Markovian stepsizes is uniform
on the set \eqref{state:space:m}.
Then, the lower bound for the tail-index $\hat{\alpha}^{(g)}$ is strictly less than
the lower bound for the tail-index $\hat{\alpha}^{(m)}$ for the SGD with cyclic stepsizes.
\end{proposition}

%%%%%%%%%%%%%%%%%%%%%%%%%%%%%%%%%%%%%%%%%%%%%%%%%%%%%%%%%%%%%

Theorem~\ref{thm:Markov:regeneration} in the main text is of asymptotic nature which characterizes the stationary distribution $x_{\infty}$ of
SGD iterations 
with a tail-index $\alpha^{(r)}$. 
Next, we provide non-asymptotic moment
bounds for $x_{r_{k}}$ at each $r_{k}$-th iterate,
and also for the limit $x_{\infty}$.

\begin{lemma}\label{lem:finite:bound:Markov:Gaussian}
Assume \textbf{(A3)} holds.

(i) For any $p\leq 1$ and $h^{(r)}(p)<1$,
\begin{equation}
\mathbb{E}\Vert x_{r_{k}}\Vert^{p}
\leq
\left(h^{(r)}(p)\right)^{k}\mathbb{E}\Vert x_{0}\Vert^{p}
+\frac{1-(h^{(r)}(p))^{k}}{1-h^{(r)}(p)}\mathbb{E}\left\Vert q_{1}^{(r)}\right\Vert^{p}.
\end{equation}

(ii) For any $p>1$, $\epsilon>0$ and $(1+\epsilon)h^{(r)}(p)<1$,
\begin{align}
\mathbb{E}\Vert x_{r_{k}}\Vert^{p}
&\leq
\left((1+\epsilon)h^{(r)}(p)\right)^{k}\mathbb{E}\Vert x_{0}\Vert^{p}
+\frac{1-((1+\epsilon)h^{(r)}(p))^{k}}{1-(1+\epsilon)h^{(r)}(p)}
\frac{(1+\epsilon)^{\frac{p}{p-1}}-(1+\epsilon)}{\left((1+\epsilon)^{\frac{1}{p-1}}-1\right)^{p}}
\mathbb{E}\left\Vert q_{1}^{(r)}\right\Vert^{p}.
\end{align}
\end{lemma}

%%%%%%%%%%%%%%%%%%%%%%%%%%%%%%%%%%%%%%%%%%%%%%%%%%%%%%%%%%%

Next, we will study the speed of convergence of the SGD to its stationary distribution $x_{\infty}$ in the Wasserstein metric $\mathcal{W}_{p}$ for any $p$ such that $h^{(r)}(p)<1$. 

\begin{theorem}\label{thm:convergence:Markov:Gaussian}
Assume \textbf{(A3)} holds.
Let $\nu_{r_{k}}$, $\nu_{\infty}$ denote the probability laws
of $x_{r_{k}}$ and $x_{\infty}$ respectively. 
Then  
\begin{equation}
\mathcal{W}_{p}(\nu_{r_{k}},\nu_{\infty})
\leq
\left(h^{(r)}(p)\right)^{k/p}
\mathcal{W}_{p}(\nu_{0},\nu_{\infty}),
\end{equation}
for any $p\geq 1$ and $h^{(r)}(p)<1$,
where the convergence rate $\left(h^{(r)}(p)\right)^{1/p}\in(0,1)$.
\end{theorem}

%%%%%%%%%%%%%%%%%%%%%%%%%%%%%%
Similar as in Corollary~\ref{cor:clt}, 
we have the following generalized CLT (GCLT)
result for
$S_{K}^{(r)}:=\sum_{k=1}^{K} x_{r_{k}}$ 
when it is scaled properly so that the limit will be an alpha-stable  distribution. 

\begin{corollary}\label{cor:clt:Markovian}
Assume \textbf{(A3)} holds and the conditions of Theorem~\ref{thm:cyclic} are satisfied. Then, we have the following:

(i) If  $\alpha^{(r)} \in(0,1) \cup(1,2)$, then there is a sequence $d_{K}=d_{K}(\alpha^{(r)})$  and a function $C_{\alpha^{(r)}}: \mathbb{S}^{d-1} \mapsto \mathbb{C}$  such that as $K\rightarrow\infty$ the random variables $K^{-\frac{1}{\alpha^{(r)}}}\left(S_{K}^{(r)}-d_{K}\right)$ converge in law to the $\alpha^{(r)}$-stable random variable with characteristic function
$\Upsilon_{\alpha^{(r)}}(tv)=\exp(t^{\alpha^{(r)}}C_{\alpha^{(r)}}(v))$, for $t>0$ and $v \in \mathbb{S}^{d-1}$.

(ii) If $\alpha^{(r)}=1$, then there are functions $\xi, \tau:(0, \infty) \mapsto \mathbb{R}$ and $C_{1}: \mathbb{S}^{d-1} \mapsto \mathbb{C}$ such that as $K\rightarrow\infty$ the random variables $K^{-1} S_{K}^{(r)}-K \xi\left(K^{-1}\right)$ converge in law to the random variable with characteristic function
$\Upsilon_{1}(t v)=\exp \left(t C_{1}(v)+i t\langle v, \tau(t)\rangle\right)$, 
for $t>0$ and $v \in \mathbb{S}^{d-1}$.

(iii) If \(\alpha^{(r)}=2,\) then there is a sequence \(d_{K}=d_{K}(2)\) and a function \(C_{2}: \mathbb{S}^{d-1} \mapsto \mathbb{R}\) such that as $K\rightarrow\infty$
the random variables \((K \log K)^{-\frac{1}{2}}\left(S_{K}^{(r)}-d_{K}\right)\) converge in law to the random variable with
characteristic function
$\Upsilon_{2}(t v)=\exp \left(t^{2} C_{2}(v)\right)$, for $t>0$ and $v \in \mathbb{S}^{d-1}$.

(iv) If \(\alpha^{(r)} \in(0,1),\) then \(d_{K}=0,\) and if \(\alpha^{(r)} \in(1,2],\) then \(d_{K}=K \bar{x},\) where \(\bar{x}=\int_{\mathbb{R}^{d}} x \nu_\infty(d x) .\) 
\end{corollary}

For the specific case of $\alpha$-stable distributions, accurate and computationally efficient estimators, which \emph{do not} require the knowledge of the functions $C_\alpha$, $\tau$, $\xi$, have been proposed \citep{mohammadi2015estimating}. Thanks to Corollary~\ref{cor:clt:Markovian}, we will be able to use such estimators in our numerical experiments in Section~\ref{sec:numerical}.

%%%%%%%%%%%%%%%%%%%%%%%%%%%%%%%%%%%%%%%%%%%%%%%%%%%%%%%%%%%%%%%%%%%%%%%%%%%%%%%%%%%%%%%%%%%%%%%%

We end the discussions of this section by providing some
additional technical results concerning the stationary distribution
of the Markovian stepsizes, and provide a more explicit formula
for the function $h^{(r)}(s)$ that plays a central role
of defining the tail-index $\alpha^{(r)}$. 
%%%%%%%%%%%%%%%%%%%%%%%%%%%%%%%%%%%%%
We recall from \eqref{state:space:m}
that the state space is given by 
\begin{equation*}
{\color{black}\{\eta_{1},\eta_{2},\ldots,\eta_{m},\eta_{m+1}\}
=\{c_{1},c_{2},\ldots,c_{K-1},c_{K},c_{K-1},\ldots,c_{2},c_{1}\},}
\end{equation*}
where $m=2K-2$.
The stepsize goes from $\eta_{1}$ to $\eta_{2}$ with probability $1$
and it goes from $\eta_{K}$ to $\eta_{K-1}$ with probability $1$. 
In between, for any $i=2,3,\ldots,K-1,K+1,\ldots,m$,
the stepsize goes from $\eta_{i}$ to $\eta_{i+1}$ with probability $p$
and from $\eta_{i}$ to $\eta_{i-1}$ with probability $1-p$
with the understanding that $\eta_{m+1}:=\eta_{1}$.
Therefore, $p=1$ reduces to the case of cyclic stepsizes.
The Markov chain exhibits a unique stationary distribution
$\pi_{i}:=\mathbb{P}(\eta_{0}=\eta_{i})$
that is characterized in the following lemma.

\begin{lemma}\label{lem:stationary:dist}
The Markov chain exhibits a unique stationary distribution
$\pi_{i}:=\mathbb{P}(\eta_{0}=\eta_{i})$, 
where
\begin{equation}
\pi_{1}=(1-p)\frac{p-1}{2p-1}\left(\frac{1-p}{p}\right)^{K-2}\pi_{m}+\frac{p^{2}}{2p-1}\pi_{m},
\end{equation}
and for any $2\leq i\leq K-1$, 
\begin{equation}
\pi_{i}=\frac{p-1}{2p-1}\left(\frac{1-p}{p}\right)^{K-i}\pi_{m}+\frac{p}{2p-1}\pi_{m},
\end{equation}
and
\begin{equation}
\pi_{K}=\frac{p(p-1)}{2p-1}\left(\frac{1-p}{p}\right)^{m-K}\pi_{m}+\frac{p^{2}}{2p-1}\pi_{m},
\end{equation}
and for any $K+1\leq i\leq m$, 
\begin{equation}
\pi_{i}=\frac{p-1}{2p-1}\left(\frac{1-p}{p}\right)^{m-i}\pi_{m}+\frac{p}{2p-1}\pi_{m},
\end{equation}
where
\begin{align}
\pi_{m}=\Bigg(\frac{4p^{3}+2(m-3)p^{2}-(m-3)p-1}{(2p-1)^{2}}
+\frac{2p^{3}}{(2p-1)^{2}}\left(\frac{1-p}{p}\right)^{K+1}
+\frac{2p(p-1)^{2}}{(2p-1)^{2}}\left(\frac{1-p}{p}\right)^{m-K}\Bigg)^{-1}.
\end{align}
\end{lemma}

%%%%%%%%%%%%%%%%%%%%%%%%%%%%%%%%%%%%%%%%%%%%%
Next, let us provide an analytic expression
for $h^{(r)}(s)$. 
Under Assumption~\textbf{(A3)}, we define:
\begin{equation}\label{h:r:i:j}
h^{(r)}(s;\eta_{i},\eta_{j}):=\mathbb{E}_{\eta_{0}=\eta_{i}}\left[\prod_{i=1}^{r_{1}(\tau_{j})}\mathbb{E}_{H}\left[\left\Vert \left(I-\frac{\eta_{i}}{b}H\right)e_{1}\right\Vert^{s}\right]\right],
\end{equation}
where $r_{1}(\tau_{j}):=\inf\{k\geq 1:\eta_{k}=\eta_{j}\}$,
and we have the following result.

When the initialization $\eta_{0}$ follows the stationary distribution, 
i.e., $\mathbb{P}(\eta_{0}=\eta_{i})=\pi_{i}$, we conclude that
\begin{equation}\label{h:r:s:formula}
h^{(r)}(s)=\sum_{i=1}^{m}\mathbb{P}(\eta_{0}=\eta_{i})h^{(r)}(s;\eta_{i},\eta_{i})
=\sum_{i=1}^{m}\pi_{i}h^{(r)}(s;\eta_{i},\eta_{i}),
\end{equation}
where $\pi_{i}$ are given in Lemma~\ref{lem:stationary:dist} and $h^{(r)}(s;\eta_{i},\eta_{i})$ is 
defined in \eqref{h:r:i:j}.
In the next proposition, we compute out $h^{(r)}(s;\eta_{i},\eta_{i})$
explicitly and hence we obtain an explicit formula
for $h^{(r)}(s)$ using \eqref{h:r:s:formula} and Lemma~\ref{lem:stationary:dist}.

\begin{proposition}\label{prop:h:eta:i:j}
Under Assumption~\textbf{(A3)}, for any $1\leq i,j\leq m$,
\begin{equation}
h^{(r)}(s;\eta_{i},\eta_{j})
=\left((I-Q^{j})^{-1}p^{j}\right)_{i},
\end{equation}
where $p^{j}:=[p_{1j},p_{2j},\ldots,p_{mj}]^{T}$, 
where for any $i=2,\ldots,K-1,K+1,\ldots,m$
\begin{equation}
p_{ij}:=p\mathbb{E}_{H}\left[\left\Vert \left(I-\frac{\eta_{i+1}}{b}H\right)e_{1}\right\Vert^{s}\right]1_{j=i+1}
+(1-p)\mathbb{E}_{H}\left[\left\Vert \left(I-\frac{\eta_{i-1}}{b}H\right)e_{1}\right\Vert^{s}\right]1_{j=i-1},
\end{equation}
and
\begin{align}
p_{1j}:=\mathbb{E}_{H}\left[\left\Vert \left(I-\frac{\eta_{2}}{b}H\right)e_{1}\right\Vert^{s}\right]1_{j=2},
\qquad
p_{Kj}:=\mathbb{E}_{H}\left[\left\Vert \left(I-\frac{\eta_{K+1}}{b}H\right)e_{1}\right\Vert^{s}\right]1_{j=K+1},
\end{align}
and $Q^{j}:=(Q_{i\ell}^{j})_{1\leq i,\ell\leq m}$
such that for any $i=2,\ldots,K-1,K+1,\ldots,m$
\begin{align}
Q_{i\ell}^{j}&:=p\mathbb{E}_{H}\left[\left\Vert \left(I-\frac{\eta_{i+1}}{b}H\right)e_{1}\right\Vert^{s}\right]1_{j\neq i+1}1_{\ell=i+1}
+(1-p)\mathbb{E}_{H}\left[\left\Vert \left(I-\frac{\eta_{i-1}}{b}H\right)e_{1}\right\Vert^{s}\right]1_{j\neq i-1}1_{\ell=i-1},
\end{align}
and
\begin{align}
Q_{1\ell}^{j}:=1_{j\neq 2}1_{\ell=2},
\qquad
Q_{K\ell}^{j}:=1_{j\neq K+1}1_{\ell=K+1}.
\end{align}
\end{proposition}

%%%%%%%%%%%%%%%%%%%%%%%%%%%%%%%%%%%%%%%%%%%%%%%%%%%%%%%%%%%%%%%%
\subsubsection{Markovian Stepsizes with Two-State Space}

In this section, we study the SGD with Markovian stepsizes with two-state space.
With the general finite state space, we have seen previously that
the tail-index $\alpha^{(r)}$ is the unique positive value such that $h^{(r)}\left(\alpha^{(r)}\right)=1$.
However, the expression for $h^{(r)}(s)$ is quite complicated. 
We are able to characterize $h^{(r)}(s)$ in a more explicit
way for the two-state space case.
First, we recall from Lemma~\ref{lem:alternative:expression:Markov}
that $h^{(r)}(s)=\tilde{h}^{(r)}(s)$ and $\rho^{(r)}=\tilde{\rho}^{(r)}$, 
with $\tilde{h}^{(r)}(s)$ and $\tilde{\rho}^{(r)}$ given in Lemma~\ref{lem:alternative:expression:Markov}.
We have the following result, which plays a central role in order to obtain Proposition~\ref{cor:mono:p}.

\begin{lemma}\label{cor:two:state}
{\color{black}Consider the two-state Markov chain, i.e. $\mathbb{P}(\eta_{1}=\eta_{u}|\eta_{0}=\eta_{l})=p$
and $\mathbb{P}(\eta_{1}=\eta_{l}|\eta_{0}=\eta_{u})=p$ 
and assume that $(1-p)\mathbb{E}_{H}\left[\left\Vert \left(I-\frac{\eta_{l}}{b}H\right)e_{1}\right\Vert^{s}\right]<1$ and $(1-p)\mathbb{E}_{H}\left[\left\Vert \left(I-\frac{\eta_{u}}{b}H\right)e_{1}\right\Vert^{s}\right]<1$.} Then, we have
\begin{align}
\tilde{h}^{(r)}(s)&=
\frac{\mathbb{E}_{H}\left[\left\Vert \left(I-\frac{\eta_{l}}{b}H\right)e_{1}\right\Vert^{s}\right](1-p+(2p-1)\mathbb{E}_{H}\left[\left\Vert \left(I-\frac{\eta_{u}}{b}H\right)e_{1}\right\Vert^{s}\right])}{2(1-(1-p)\mathbb{E}_{H}\left[\left\Vert \left(I-\frac{\eta_{u}}{b}H\right)e_{1}\right\Vert^{s}\right])}
\nonumber
\\
&\qquad
+\frac{\mathbb{E}_{H}\left[\left\Vert \left(I-\frac{\eta_{u}}{b}H\right)e_{1}\right\Vert^{s}\right](1-p+(2p-1)\mathbb{E}_{H}\left[\left\Vert \left(I-\frac{\eta_{l}}{b}H\right)e_{1}\right\Vert^{s}\right])}{2(1-(1-p)\mathbb{E}_{H}\left[\left\Vert \left(I-\frac{\eta_{l}}{b}H\right)e_{1}\right\Vert^{s}\right])},
\end{align}
and
\begin{equation}
\tilde{\rho}^{(r)}
=\mathbb{E}_{H}\left[\log\left\Vert \left(I-\frac{\eta_{l}}{b}H\right)e_{1}\right\Vert\right]
+\mathbb{E}_{H}\left[\log\left\Vert \left(I-\frac{\eta_{u}}{b}H\right)e_{1}\right\Vert\right].
\end{equation}
In particular, when $p=1$, 
we get
$\tilde{h}^{(r)}(s)=\mathbb{E}_{H}\left[\left\Vert \left(I-\frac{\eta_{l}}{b}H\right)e_{1}\right\Vert^{s}\right]
\mathbb{E}_{H}\left[\left\Vert \left(I-\frac{\eta_{u}}{b}H\right)e_{1}\right\Vert^{s}\right]
=h^{(m)}(s)$.
\end{lemma}

In Proposition~\ref{alpha:2:Markov}, we can write $c^{(r)}$ as $c^{(r)}=\tilde{h}^{(r)}(2)$.
Therefore, we immediately obtain the following result by applying Lemma~\ref{cor:two:state}.

\begin{corollary}\label{cor:c:r}
Consider stepsizes following the two-state Markov chain, i.e. $\mathbb{P}(\eta_{1}=\eta_{u}|\eta_{0}=\eta_{l})=p$
and $\mathbb{P}(\eta_{1}=\eta_{l}|\eta_{0}=\eta_{u})=p$. 
In Proposition~\ref{alpha:2:Markov}, we have
\begin{align}
c^{(r)}&=
\frac{\left(1-2\eta_{l}\sigma^{2}+\frac{\eta_{l}^{2}\sigma^{4}}{b}(d+b+1)\right)\left(1-p+(2p-1)\left(1-2\eta_{u}\sigma^{2}+\frac{\eta_{u}^{2}\sigma^{4}}{b}(d+b+1)\right)\right)}{2\left(1-(1-p)\left(1-2\eta_{u}\sigma^{2}+\frac{\eta_{u}^{2}\sigma^{4}}{b}(d+b+1)\right)\right)}
\nonumber
\\
&\quad
+\frac{\left(1-2\eta_{u}\sigma^{2}+\frac{\eta_{u}^{2}\sigma^{4}}{b}(d+b+1)\right)\left(1-p+(2p-1)\left(1-2\eta_{l}\sigma^{2}+\frac{\eta_{l}^{2}\sigma^{4}}{b}(d+b+1)\right)\right)}{2\left(1-(1-p)\left(1-2\eta_{l}\sigma^{2}+\frac{\eta_{l}^{2}\sigma^{4}}{b}(d+b+1)\right)\right)}.
\end{align}
\end{corollary}

We recall from Proposition~\ref{alpha:2:Markov} that
(i) There exists $\delta>0$ such that
for any $1<c^{(r)}<1+\delta$, tail-index $0<\alpha^{(r)}< 2$. (ii) If $c^{(r)}=1$, tail-index $\alpha^{(r)}=2$. (iii) If $c^{(r)}<1$, then tail-index $\alpha^{(r)}>2$.

%%%%%%%%%%%%%%%%%%%%%%%%%%%%%%
\section{Technical Lemmas}\label{sec:tech:lemma}

\begin{lemma}\label{lem:alternative:expression}
Assume \textbf{(A3)} holds. 
Then, we have
\begin{align}
\rho=\tilde{\rho},
\qquad
h(s)=\tilde{h}(s),\quad\text{for every $s\geq 0$},
\end{align}
where
\begin{equation}\label{tilde:rho:defn}
\tilde{\rho}:=\mathbb{E}\left[\log\left\Vert\left(I-\frac{\eta_{1}}{b}H_{1}\right)e_{1}\right\Vert\right],
\end{equation}
and
\begin{equation}
\tilde{h}(s):=\mathbb{E}\left[\Vert M_{1}e_{1}\Vert^{s}\right]
=\mathbb{E}\left[\left\Vert\left(I-\frac{\eta_{1}}{b}H_{1}\right)e_{1}\right\Vert^{s}\right].
\end{equation}
\end{lemma}

\begin{lemma}\label{lem:simplified}
Assume \textbf{(A3)} holds.
For any $s\geq 0$, $h(s)=\tilde{h}(s)$ and $\rho=\tilde{\rho}$, where
\begin{align*}
\tilde{h}(s)
=\mathbb{E}\left[\left(\left(1-\frac{\eta\sigma^{2}}{b}X\right)^{2}
+\frac{\eta^{2}\sigma^{4}}{b^{2}}XY\right)^{s/2}\right],
\end{align*}
and
\begin{align*}
\tilde{\rho}:=\frac{1}{2}
\mathbb{E}\left[\log\left(\left(1-\frac{\eta\sigma^{2}}{b}X\right)^{2}+\frac{\eta^{2}\sigma^{4}}{b^{2}}XY\right)\right],
\end{align*}
where $\eta,X,Y$ are independent and $X$ is chi-square random variable
with degree of freedom $b$ and $Y$ is a chi-square random variable
with degree of freedom $(d-1)$.
\end{lemma}

\begin{lemma}\label{lem:alternative:expression:cyclic}
Assume \textbf{(A3)} holds. For any $s\geq 0$,
\begin{equation}
\left(h^{(m)}(s)\right)^{1/m}
=\tilde{h}^{(m)}(s),
\end{equation}
where
\begin{equation}
\tilde{h}^{(m)}(s):=\left(\prod_{i=1}^{m}\mathbb{E}\left[\left\Vert\left(I-\frac{\eta_{i}}{b}H\right)e_{1}\right\Vert^{s}\right]\right)^{1/m}.
\end{equation}
Moreover
\begin{equation}\label{tilde:rho:defn:m}
\rho^{(m)}=\tilde{\rho}^{(m)}:=\sum_{i=1}^{m}\mathbb{E}\left[\log\left\Vert\left(I-\frac{\eta_{i}}{b}H\right)e_{1}\right\Vert\right].    
\end{equation}
\end{lemma}

\begin{lemma}\label{lem:simplified:cyclic}
Assume \textbf{(A3)} holds.
For any $s\geq 0$, we have $h^{(m)}(s)=\tilde{h}^{(m)}(s)$ and $\rho^{(m)}=\tilde{\rho}^{(m)}$, where
\begin{align*}
&\tilde{h}^{(m)}(s)
=\left(\prod_{i=1}^{m}\mathbb{E}\left[\left(\left(1-\frac{\eta_{i}\sigma^{2}}{b}X\right)^{2}
+\frac{\eta_{i}^{2}\sigma^{4}}{b^{2}}XY\right)^{s/2}\right]\right)^{1/m},
\\
&\tilde{\rho}^{(m)}=\frac{1}{2}\sum_{i=1}^{m}\mathbb{E}\left[\log\left(\left(1-\frac{\eta_{i}\sigma^{2}}{b}X\right)^{2}+\frac{\eta_{i}^{2}\sigma^{4}}{b^{2}}XY\right)\right],
\end{align*}
where $X,Y$ are independent and $X$ is chi-square random variable
with degree of freedom $b$ and $Y$ is a chi-square random variable
with degree of freedom $(d-1)$.
\end{lemma}

\begin{lemma}\label{lem:alternative:expression:Markov}
For any $s\geq 0$,
\begin{equation}
h^{(r)}(s)
=\tilde{h}^{(r)}(s):=\mathbb{E}\left[\prod_{i=1}^{r_{1}}\mathbb{E}_{H}\left[\left\Vert \left(I-\frac{\eta_{i}}{b}H\right)e_{1}\right\Vert^{s}\right]\right],
\end{equation}
and moreover,
\begin{equation}\label{tilde:rho:defn:r}
\rho^{(r)}=\tilde{\rho}^{(r)}:=\mathbb{E}\left[\sum_{i=1}^{r_{1}}\mathbb{E}_{H}\left[\log\left\Vert \left(I-\frac{\eta_{i}}{b}H\right)e_{1}\right\Vert\right]\right],
\end{equation}
where $r_{1}$ is defined in \eqref{defn:regeneration:time}.
\end{lemma}

\begin{lemma}\label{lem:simplified:Markov:regeneration}
Assume \textbf{(A3)} holds.
For any $s\geq 0$, we have $h^{(r)}(s)=\tilde{h}^{(r)}(s)$ and $\rho^{(r)}=\tilde{\rho}^{(r)}$, where
\begin{align*}
&\tilde{h}^{(r)}(s)
=\mathbb{E}\left[\prod_{i=1}^{r_{1}}\mathbb{E}_{X,Y}\left[\left(\left(1-\frac{\eta_{i}\sigma^{2}}{b}X\right)^{2}
+\frac{\eta_{i}^{2}\sigma^{4}}{b^{2}}XY\right)^{s/2}\right]\right],
\\
&\tilde{\rho}^{(r)}
=\frac{1}{2}\mathbb{E}\left[\sum_{i=1}^{r_{1}}\mathbb{E}_{X,Y}\left[\log\left(\left(1-\frac{\eta_{i}\sigma^{2}}{b}X\right)^{2}
+\frac{\eta_{i}^{2}\sigma^{4}}{b^{2}}XY\right)\right]\right],
\end{align*}
where $\mathbb{E}_{X,Y}$ denotes the expectation w.r.t. $X,Y$, where $X,Y$ are independent and $X$ is chi-square random variable
with degree of freedom $b$ and $Y$ is a chi-square random variable
with degree of freedom $(d-1)$ and $X,Y$ are independent of $(\eta_{k})_{k\in\mathbb{N}}$.
\end{lemma}

%%%%%%%%%%%%%%%%%%%%%%%%%%%%%%
\section{Technical Proofs}

\subsection{Proof of results in Section~\ref{sec:main:results}}

%%%%%%%%%%%%%%%%%%%%%%%%%%%%%%%%%%%%%%%%%%%%%%%%%%%%%%%%%%%%%%%%%%
%%%%%%%%%%%%%%%%%%%%%%%%%%%%%%%%%%%%%%%%%%%%%%%%%%%%%%%
\subsection*{Proof of Theorem~\ref{thm:Markov:regeneration:index}}
It follows from the proof of Theorem~4 in \cite{HTP2021}
that for any $s\geq 1$, conditional on $\eta_{i}$, 
$\mathbb{E}_{H}\left[\left\Vert I-\frac{\eta_{i}}{b}H\right\Vert^{s}\right]$
is strictly decreasing in $b$. Therefore, $\hat{h}^{(r)}(s)$ is strictly decreasing in $b$.
It thus follows from the arguments in the proof of Theorem~4 in \cite{HTP2021} that
$\hat{\alpha}^{(r)}$ is strictly increasing in batch-size $b$
provided that $\hat{\alpha}^{(r)}\geq 1$. The proof is complete.
\hfill $\Box$

%%%%%%%%%%%%%%%%%%%%%%%%%%%%%%%%%%%%%%%%%%%%%%%%%%
\subsection*{Proof of Theorem~\ref{thm:mono:Gaussian:Markov:regeneration}}
Given $\rho^{(r)}<0$, the tail-index $\alpha^{(r)}$ is the unique positive value
such that $\tilde{h}^{(r)}(s)\left(\alpha^{(r)}\right)=1$.
It follows from Theorem~4 in \cite{HTP2021} that
conditional on $\eta_{i}$, $\mathbb{E}_{H}\left[\left\Vert\left(I-\frac{\eta_{i}}{b}H\right)e_{1}\right\Vert^{s}\right]$
is strictly decreasing in batch-size $b$ for any $s\geq 1$, 
and it is strictly increasing in dimension $d$. 
Therefore, $\tilde{h}^{(r)}(s)$ is strictly decreasing in batch-size $b$ for any $s\geq 1$, 
and it is strictly increasing in dimension $d$, and the conclusion follows.
\hfill $\Box$

%%%%%%%%%%%%%%%%%%%%%%%%%%%%%%%%%%%%%

%%%%%%%%%%%%%%%%%%%%%%%%%%%%%%%%%%%%%%%%%%%%%%%%%%%%%%%
\subsection*{Proof of Theorem~\ref{thm:cyclic:index}}
It follows from the proof of Theorem~4 in \cite{HTP2021}
that for any $s\geq 1$, the function
\begin{equation*}
\mathbb{E}\left[\left\Vert I-\frac{\eta_{i}}{b}H\right\Vert^{s}\right]
\end{equation*}
is strictly decreasing in $b$. Therefore, $\hat{h}^{(m)}(s)$ is strictly decreasing in $b$.
It thus follows from the arguments in the proof of Theorem~4 in \cite{HTP2021} that
$\hat{\alpha}^{(m)}$ is strictly increasing in batch-size $b$
provided that $\hat{\alpha}^{(m)}\geq 1$. The proof is complete.
\hfill $\Box$

%%%%%%%%%%%%%%%%%%%%%%%%%%%%%%%%%%%%%%%%%%%%%%%%%%
\subsection*{Proof of Theorem~\ref{thm:mono:Gaussian:cyclic}}
Given that $\rho^{(m)}<0$, 
the tail-index $\alpha^{(m)}$ is the unique positive value
such that $\tilde{h}^{(m)}\left(\alpha^{(m)}\right)=1$.
It follows from Theorem~4 in \cite{HTP2021} that
$\mathbb{E}\left[\left\Vert\left(I-\frac{\eta_{i}}{b}H\right)e_{1}\right\Vert^{s}\right]$
is strictly decreasing in batch-size $b$ for any $s\geq 1$, 
and it is strictly increasing in dimension $d$. 
Therefore, $\tilde{h}^{(m)}(s)$ is strictly decreasing in batch-size $b$ for any $s\geq 1$, 
and it is strictly increasing in dimension $d$, and the conclusion follows.
\hfill $\Box$

%%%%%%%%%%%%%%%%%%%%%%%%%%%%%%%%%%%%%%%%%%%%%%%%%%%%%
\subsection{Proofs of Results in Section~\ref{sec:comparison}}

\subsection*{Proof of Proposition~\ref{prop:iid:constant:comparison}}
By Lemma~\ref{lemma-cvx-in-step}, 
for any given positive semi-definite symmetric matrix $H$ fixed, the function $F_H:[0,\infty) \to \mathbb{R}$ defined as
$ F_H(a) :=\left\| \left(I - a H\right)e_{1}\right\|^s$ 
is convex for $s\geq 1$. By tower property and Jensen's inequality, 
\begin{align*}
h(s)&=\mathbb{E}\left[\mathbb{E}\left[\left\Vert\left(I-\frac{\eta}{b}H\right)e_{1}\right\Vert^{s}\bigg|H\right]\right]
\\
&\geq\mathbb{E}\left[\left\Vert\mathbb{E}\left[\left(I-\frac{\eta}{b}H\right)e_{1}\bigg|H\right]\right\Vert^{s}\right]
=\mathbb{E}\left[\left\Vert\left(I-\frac{\mathbb{E}[\eta]}{b}H\right)e_{1}\right\Vert^{s}\right],
\end{align*}
which is the $h$ function with constant stepsize $\mathbb{E}[\eta]$. 
Since $\eta$ is random, the above inequality is strict, hence we conclude
that the tail-index $\alpha$ is strictly less than the tail-index $\alpha_{c}$
with constant stepsize $\mathbb{E}[\eta]$ provided that $\alpha\geq 1$.
The proof is complete.
\hfill $\Box$

%%%%%%%%%%%%%%%%%%%%%%%%%%%%%%%%%%%%%%%%%%%%%%%%%%%%%%%%%%%%%%%%%%%%%%%%%%%%
\subsection*{Proof of Proposition~\ref{prop:compare:iid:cyclic}}
We recall that the tail-index $\alpha^{(m)}$ for the SGD with cyclic stepsizes is the unique
positive value such that $h^{(m)}\left(\alpha^{(m)}\right)=1$.
By the inequality of arithmetic and geometric means, we obtain
\begin{equation}
h^{(m)}(s)\leq\frac{1}{m}\sum_{i=1}^{m}\mathbb{E}\left[\left\Vert\left(I-\frac{\eta_{i}}{b}H\right)e_{1}\right\Vert^{s}\right]=h(s).
\end{equation}
Since $\eta_{i}$ is not constant, the above inequality is strict.
Therefore, we conclude that the tail-index $\alpha$ of SGD with i.i.d. stepsizes is strictly less than
the tail-index $\alpha^{(m)}$ for the SGD with cyclic stepsizes.
The proof is complete.
\hfill $\Box$

%%%%%%%%%%%%%%%%%%%%%%%%%%%%%%%%%%%%%%%%%%%
\subsection*{Proof of Proposition~\ref{prop:compare:cyclic:constant}}
{\color{black}Under the assumption \textbf{(A3)},
we have $h_{c}(\alpha_{c})=1$ and $h^{(m)}\left(\alpha^{(m)}\right)=1$, where by Lemma~\ref{lem:simplified:cyclic}
\begin{align*}
h_{c}(s):=h\left(s;\frac{1}{m}\sum_{i=1}^{m}\eta_{i}\right),
\qquad\text{and}\qquad
h^{(m)}(s)=\left(\prod_{i=1}^{m}h(s;\eta_{i})\right)^{1/m},
\end{align*}
where
\begin{equation*}
h(s;\eta):=\mathbb{E}\left[\left(\left(1-\frac{\eta\sigma^{2}}{b}X\right)^{2}
+\frac{\eta^{2}\sigma^{4}}{b^{2}}XY\right)^{s/2}\right],
\end{equation*}
where $X,Y$ are independent and $X$ is a chi-square random variable
with a degree of freedom $b$ and $Y$ is a chi-square random variable
with a degree of freedom $(d-1)$. 
We can compute that
\begin{align*}
\frac{\partial}{\partial\eta}
h(s;\eta)=\mathbb{E}\left[\frac{s}{2}\left(\left(1-\frac{\eta\sigma^{2}}{b}X\right)^{2}
+\frac{\eta^{2}\sigma^{4}}{b^{2}}XY\right)^{\frac{s}{2}-1}\left(-\frac{2\sigma^{2}}{b}X+\frac{2\eta\sigma^{4}}{b^{2}}X^{2}+\frac{2\eta\sigma^{4}}{b^{2}}XY\right)\right],
\end{align*}
and
\begin{align*}
\frac{\partial^{2}}{\partial\eta^{2}}
h(s;\eta)&=\mathbb{E}\left[\frac{s}{2}\left(\left(1-\frac{\eta\sigma^{2}}{b}X\right)^{2}
+\frac{\eta^{2}\sigma^{4}}{b^{2}}XY\right)^{\frac{s}{2}-1}\left(\frac{2\sigma^{4}}{b^{2}}X^{2}+\frac{2\sigma^{4}}{b^{2}}XY\right)\right]
\\
&\qquad
+\mathbb{E}\left[\frac{s}{2}\left(\frac{s}{2}-1\right)\left(\left(1-\frac{\eta\sigma^{2}}{b}X\right)^{2}
+\frac{\eta^{2}\sigma^{4}}{b^{2}}XY\right)^{\frac{s}{2}-2}\left(-\frac{2\sigma^{2}}{b}X+\frac{2\eta\sigma^{4}}{b^{2}}X^{2}+\frac{2\eta\sigma^{4}}{b^{2}}XY\right)^{2}\right],
\end{align*}
and therefore
\begin{align*}
&h(s;0)\frac{\partial^{2}}{\partial\eta^{2}}
h(s;0)-\left(\frac{\partial}{\partial\eta}
h(s;0)\right)^{2}
\\
&=\mathbb{E}\left[\frac{s}{2}\left(\frac{2\sigma^{4}}{b^{2}}X^{2}+\frac{2\sigma^{4}}{b^{2}}XY\right)\right]
+\mathbb{E}\left[\frac{s}{2}\left(\frac{s}{2}-1\right)\frac{4\sigma^{4}}{b^{2}}X^{2}\right]
-\frac{s^{2}}{4}\frac{4\sigma^{4}}{b^{2}}(\mathbb{E}[X])^{2}
\\
&=\frac{s\sigma^{4}}{b}(d+b+1)
+s(s-2)\frac{\sigma^{4}}{b}(b+2)
-s^{2}\sigma^{4}
=\frac{s\sigma^{4}}{b}\left(d-b+2s-3\right)
>0,
\end{align*}
for any $s>0$ provided that $d\geq b+3$. 
This implies that under the assumption $d\geq b+3$
and the stepsize $\eta>0$ is sufficiently small, $h(s;\eta)$ is log-convex
in $\eta$ and hence by Jensen's inequality, 
$h^{(m)}(s)\geq h_{c}(s)$, which implies
that $\alpha^{(m)}\leq\alpha_{c}$.
This completes the proof.}
\hfill $\Box$

%%%%%%%%%%%%%%%%%%%%%%%%%%%%%%%%%%%%%%%

\subsection*{Proof of Proposition~\ref{cor:mono:p}}
{\color{black}Let us denote
\begin{equation}
x:=\mathbb{E}_{H}\left[\left\Vert \left(I-\frac{\eta_{l}}{b}H\right)e_{1}\right\Vert^{s}\right],
\qquad
y:=\mathbb{E}_{H}\left[\left\Vert \left(I-\frac{\eta_{u}}{b}H\right)e_{1}\right\Vert^{s}\right].
\end{equation}}
We also define:
\begin{equation}
F(p):=\frac{x(1-p+(2p-1)y)}{2(1-(1-p)y)}+\frac{y(1-p+(2p-1)x)}{2(1-(1-p)x)}.
\end{equation}
Then, it follows from Lemma~\ref{cor:two:state}
that $\tilde{h}^{(r)}(s)=F(p)$
{\color{black}provided that $(1-p)x<1$ and $(1-p)y<1$.
For any $p\in\mathcal{P}$ where $\mathcal{P}$ is defined in \eqref{P:set} and $s\in\mathcal{S}$ where
$\mathcal{S}$ is a sufficiently small
interval that contains $\alpha^{(r)}$, 
we have $(1-p)x<1$ and $(1-p)y<1$.}
We can compute that
\begin{align*}
\frac{\partial F}{\partial p}
&=\frac{x(-1+2y)(1-(1-p)y)-x(1-p+(2p-1)y)y}{2(1-(1-p)y)^{2}}
\\
&\qquad\qquad
+\frac{y(-1+2x)(1-(1-p)x)-y(1-p+(2p-1)x)x}{2(1-(1-p)x)^{2}}
\\
&=\frac{-x(1-y)^{2}}{2(1-(1-p)y)^{2}}
+\frac{-y(1-x)^{2}}{2(1-(1-p)x)^{2}}<0,
\end{align*}
so that $\tilde{h}^{(r)}(s)$ is decreasing in {\color{black}$p\in\mathcal{P}$
for any $s\in\mathcal{S}$}
and hence the tail-index $\alpha^{(r)}$ is increasing in $p\in\mathcal{P}$.
{\color{black}Finally, $p=1\in\mathcal{P}$
and $\alpha^{(r)}$ reduces to $\alpha^{(m)}$ when $p=1$
which implies that $\alpha^{(r)}\leq\alpha^{(m)}$.}
The proof is complete.
\hfill $\Box$

%%%%%%%%%%%%%%%%%%%%%%%%%%%%%%%%%%%%%%%%%%%%%%%%%%%%%%

\subsection*{Proof of Proposition~\ref{cor:comparison}}
First of all, we recall that 
$\alpha$ is the tail-index for SGD with i.i.d. stepsizes
which is the unique position value such that $h(\alpha)=1$
and $\alpha^{(m)}$ is the tail-index for SGD with cyclic stepsizes
which is the unique position value such that $h^{(m)}\left(\alpha^{(m)}\right)=1$.
It is easy to see that
\begin{align*}
h^{(m)}(s)&=\mathbb{E}_{H}\left[\left\Vert \left(I-\frac{\eta_{l}}{b}H\right)e_{1}\right\Vert^{s}\right]
\mathbb{E}_{H}\left[\left\Vert \left(I-\frac{\eta_{u}}{b}H\right)e_{1}\right\Vert^{s}\right]
\\
&\leq\left(\frac{\mathbb{E}_{H}\left[\left\Vert \left(I-\frac{\eta_{l}}{b}H\right)e_{1}\right\Vert^{s}\right]
+\mathbb{E}_{H}\left[\left\Vert \left(I-\frac{\eta_{u}}{b}H\right)e_{1}\right\Vert^{s}\right]}{2}\right)^{2}
=(h(s))^{2},
\end{align*}
which implies that $\alpha\leq\alpha^{(m)}$.

Note that $\alpha$ and $\alpha^{(m)}$ are independent of $p$
and by Proposition~\ref{cor:mono:p}, $\alpha^{(r)}$ is increasing in $p$,
and in particular, $\alpha^{(r)}=\alpha^{(m)}$ when $p=1$. 
Moreover, as 
\begin{equation}
p\rightarrow\max\left(1-\frac{1}{\mathbb{E}_{H}\left[\left\Vert \left(I-\frac{\eta_{l}}{b}H\right)e_{1}\right\Vert^{s}\right]},1-\frac{1}{\mathbb{E}_{H}\left[\left\Vert \left(I-\frac{\eta_{u}}{b}H\right)e_{1}\right\Vert^{s}\right]}\right),
\end{equation}
by Lemma~\ref{cor:two:state}, we have $\tilde{h}^{(r)}(s)\rightarrow\infty$, 
and hence we conclude that there exists some critical $p_{c}\in(0,1)$
such that for any $p_{c}<p<1$, we have $\alpha<\alpha^{(r)}<\alpha^{(m)}$
and for any $p<p_{c}$, we have $\alpha^{(r)}<\alpha<\alpha^{(m)}$. 

Indeed one can determine the critical $p_{c}$  explicitly. 
Note that $p_{c}$ is the critical value such that $\alpha=\alpha^{(r)}$, which is equivalent 
to the critical value $p_{c}$ such that $\tilde{h}^{(r)}(\alpha)=1$. 
Hence, $p_{c}$ is determined by the equation:
\begin{align}
&\frac{\mathbb{E}_{H}\left[\left\Vert \left(I-\frac{\eta_{l}}{b}H\right)e_{1}\right\Vert^{\alpha}\right](1-p_{c}+(2p_{c}-1)\mathbb{E}_{H}\left[\left\Vert \left(I-\frac{\eta_{u}}{b}H\right)e_{1}\right\Vert^{\alpha}\right])}{2(1-(1-p_{c})\mathbb{E}_{H}\left[\left\Vert \left(I-\frac{\eta_{u}}{b}H\right)e_{1}\right\Vert^{\alpha}\right])}
\nonumber
\\
&\qquad
+\frac{\mathbb{E}_{H}\left[\left\Vert \left(I-\frac{\eta_{u}}{b}H\right)e_{1}\right\Vert^{\alpha}\right](1-p_{c}+(2p_{c}-1)\mathbb{E}_{H}\left[\left\Vert \left(I-\frac{\eta_{l}}{b}H\right)e_{1}\right\Vert^{\alpha}\right])}{2(1-(1-p_{c})\mathbb{E}_{H}\left[\left\Vert \left(I-\frac{\eta_{l}}{b}H\right)e_{1}\right\Vert^{\alpha}\right])}=1.
\end{align}
After some algebraic computations, one can rewrite the above equation for $p_{c}$ 
as a quadratic equation in $p_{c}$:
\begin{align}
&\left(2(yx^{2}+y^{2}x)-(x+y)^{2}\right)p_{c}^{2}
-\left(3(yx^{2}+y^{2}x)+3(x+y)-4xy-2(x+y)^{2}\right)p_{c}\nonumber
\\
&\qquad\qquad
+3(x+y)-2xy+yx^{2}+y^{2}x-(x+y)^{2}-2=0,\label{quadratic:eqn}
\end{align}
where
\begin{equation}
x:=\mathbb{E}_{H}\left[\left\Vert \left(I-\frac{\eta_{l}}{b}H\right)e_{1}\right\Vert^{\alpha}\right],
\qquad
y:=\mathbb{E}_{H}\left[\left\Vert \left(I-\frac{\eta_{l}}{b}H\right)e_{1}\right\Vert^{\alpha}\right].
\end{equation}
By the definition of $\alpha$, we have
\begin{equation}
h(\alpha)=\frac{1}{2}x+\frac{1}{2}y=1,
\end{equation}
which implies that $x+y=2$ so that 
the quadratic equation~\eqref{quadratic:eqn} can be simplified as:
\begin{align}
&(4xy-4)p_{c}^{2}
-(2xy-2)p_{c}=0,
\end{align}
which yields that
$p_{c}=\frac{1}{2}$.
The proof is complete.
\hfill $\Box$

%%%%%%%%%%%%%%%%%%%%%%%%%%%%%%%%%%%%%%%%%%%%%%%%%%%%%%%%%%%%%%%%%%%%%%%%%%%%%%%%%%%%%%%%%%%%%%%%%%%%%%%%%%%%%%%%%%%%%%%%%%%%%%%%%
%%%%%%%%%%%%%%%%%%%%%%%%%%%%%%%%%%%%%%%%%%%%%%%%%%%%%%%%%%%%%%%%%%%%%%%%%%%%%%%%%%%%%%%%%%%%%%%%%%%%%%%%%%%%%%%%%%%%%%%%%%%%%%%%%
%%%%%%%%%%%%%%%%%%%%%%%%%%%%%%%%%%%%%%%%%%%%%%%%%%%%%%%%%%%%%%%%%%%%%%%%%%%%%%%%%%%%%%%%%%%%%%%%%%%%%%%%%%%%%%%%%%%%%%%%%%%%%%%%%
%%%%%%%%%%%%%%%%%%%%%%%%%%%%%%%%%%%%%%%%%%%%%%%%%%%%%%%%%%%%%%%%%%%%%%%%%%%%%%%%%%%%%%%%%%%%%%%%%%%%%%%%%%%%%%%%%%%%%%%%%%%%%%%%%
\subsection{Proofs of Results in Section~\ref{sec:iid}}
%%%%%%%%%%%%%%%%%%%%%%%%%%%%%%

\subsection*{Proof of Theorem~\ref{thm:iid}}
The proof is similar to the proof of Theorem~2 in \cite{HTP2021} and is omitted here.
\hfill $\Box$

%%%%%%%%%%%%%%%%%%%%%%%%%%%%%%%%%%%%%%%%%%%%%%%%%%%%%%%%%%%%
\subsection*{Proof of Theorem~\ref{thm:mono:Gaussian}}
By following the proof of Theorem~4 in \cite{HTP2021}, it suffices
to show that for any $s\geq 1$, $h(s)$ is decreasing in batch-size $b\in\mathbb{N}$
and for any $s\geq 0$, $h(s)$ is increasing in dimension $d\in\mathbb{N}$.
Under the assumption of the Gaussian input data, by tower property,
\begin{equation}
h(s)=\mathbb{E}[h(s|\eta)],
\qquad
h(s|\eta):=\mathbb{E}\left[\left\Vert\left(I-\frac{\eta}{b}H\right)e_{1}\right\Vert^{s}\bigg|\eta\right].
\end{equation}
In the proof of Theorem~4 in \cite{HTP2021}, it showed that for any given $\eta$,
for any $s\geq 1$, $h(s|\eta)$ is decreasing in batch-size $b\in\mathbb{N}$
and for any $s\geq 0$, $h(s|\eta)$ is increasing in dimension $d\in\mathbb{N}$.
Since $h(s)=\mathbb{E}[h(s|\eta)]$, we conclude that
$h(s)$ is decreasing in batch-size $b\in\mathbb{N}$
and for any $s\geq 0$, $h(s)$ is increasing in dimension $d\in\mathbb{N}$.
Hence, by following the same arguments as in the proof of Theorem~4 in \cite{HTP2021}, we
conclude that the tail-index $\alpha$ is strictly increasing in batch-size $b$
provided that $\alpha\geq 1$ and the tail-index $\alpha$ is strictly decreasing in dimension $d$.
The proof is complete.
\hfill $\Box$

\subsection*{Proof of Theorem~\ref{thm:mono:Gaussian:range}}
When $\eta$ is uniformly distributed on $(\bar{\eta}-R,\bar{\eta}+R)$, 
\begin{equation}
h(s)=\frac{1}{2R}\int_{\bar{\eta}-R}^{\bar{\eta}+R}\mathbb{E}\left[\left\Vert\left(I-\frac{x}{b}H\right)e_{1}\right\Vert^{s}\right]dx.
\end{equation}
It suffices to show that $h(s)$ is increasing in $R$ for any $s\geq 1$. 
We can compute that
\begin{align}
\frac{\partial}{\partial R}h(s)
&=\frac{-1}{2R^{2}}\int_{\bar{\eta}-R}^{\bar{\eta}+R}\mathbb{E}\left[\left\Vert\left(I-\frac{x}{b}H\right)e_{1}\right\Vert^{s}\right]dx
\nonumber
\\
&\qquad
+\frac{1}{2R}\left(\mathbb{E}\left[\left\Vert\left(I-\frac{\bar{\eta}+R}{b}H\right)e_{1}\right\Vert^{s}\right]
+\mathbb{E}\left[\left\Vert\left(I-\frac{\bar{\eta}-R}{b}H\right)e_{1}\right\Vert^{s}\right]\right).
\end{align}
Then, it suffices to show that
\begin{equation}\label{eqn:equiv}
R(f(\bar{\eta}+R)+f(\bar{\eta}-R))\geq\int_{\bar{\eta}-R}^{\bar{\eta}+R}f(x)dx,
\end{equation}
where
\begin{equation}
f(x):=\mathbb{E}\left[\left\Vert\left(I-\frac{x}{b}H\right)e_{1}\right\Vert^{s}\right] 
\end{equation}
is convex in $x$ for any $s\geq 1$ according to Lemma~\ref{lemma-cvx-in-step}.
Note that \eqref{eqn:equiv} is equivalent to
\begin{equation}
F\left(\bar{\eta}+R;\bar{\eta}-R\right)\geq 0,
\end{equation}
where
\begin{equation}
F(x;a):=\frac{x-a}{2}(f(x)+f(a))-\int_{a}^{x}f(y)dy.
\end{equation}
Then we have $F(a;a)=0$ and
\begin{equation}
\frac{\partial}{\partial x}F(x;a)=\frac{f(a)-f(x)+(x-a)f'(x)}{2}\geq 0,
\end{equation}
which holds since $f(x)$ is convex in $x$. This implies that $F(x;a)\geq 0$ for any $x\geq a>0$.
and thus $F(\bar{\eta}+R;\bar{\eta}-R)\geq 0$, which implies \eqref{eqn:equiv}.
This completes the proof.
\hfill $\Box$
%%%%%%%%%%%%%%%%%%%%%%%%%%%%%%%%%%%%%%%%%%%%%%%%%%%%%

\subsection*{Proof of Proposition~\ref{alpha:2}}
We first prove (i). 
Let us first recall from Lemma~\ref{lem:simplified} that
\begin{align*}
&\tilde{h}(s)
=\mathbb{E}\left[\left(\left(1-\frac{\eta\sigma^{2}}{b}X\right)^{2}
+\frac{\eta^{2}\sigma^{4}}{b^{2}}XY\right)^{s/2}\right],
\\
&\tilde{\rho}=\frac{1}{2}\mathbb{E}\left[\log\left(\left(1-\frac{\eta\sigma^{2}}{b}X\right)^{2}+\frac{\eta^{2}\sigma^{4}}{b^{2}}XY\right)\right],
\end{align*}
where $X,Y$ are independent and $X$ is chi-square random variable
with degree of freedom $b$ and $Y$ is a chi-square random variable
with degree of freedom $(d-1)$, and $X,Y$ are independent of $\eta$.
When $c=1-2\mathbb{E}[\eta]\sigma^{2}+\frac{\mathbb{E}[\eta^{2}]\sigma^{4}}{b}(d+b+1)=1$, 
we can compute that
\begin{align}
\tilde{\rho}
&\leq\frac{1}{2}\log
\mathbb{E}\left[1- \frac{2\eta\sigma^2}{b}X
+ \frac{\eta^2 \sigma^4}{b^{2}}(X^{2}+XY)\right]
\label{tilde:rho:inequality}
\\
&=\frac{1}{2}\log\left(1-2\mathbb{E}[\eta]\sigma^{2}+\frac{\mathbb{E}[\eta^{2}]\sigma^{4}}{b}(d+b+1)\right)=0.
\nonumber
\end{align}
Note that since 
$1- \frac{2\eta\sigma^2}{b} X
+ \frac{\eta^2 \sigma^4}{b^{2}}(X^{2}+XY)$
is
random, the inequality in \eqref{tilde:rho:inequality} is a strict inequality
from Jensen's inequality. Thus, when 
 $c=1$, we have $\tilde{\rho}<0$. 
By continuity, there exists some $\delta>0$
such that for any  $1<c<1+\delta$
we have $\tilde{\rho}<0$. Moreover, when $c>1$,
we have
\begin{align*}
\tilde{h}(2)
&=\mathbb{E}\left[1- \frac{2\eta \sigma^2}{b}X
+ \frac{\eta^2 \sigma^4}{b^{2}}(X^{2}+XY)\right]
\\
&=1-2\mathbb{E}[\eta]\sigma^{2}+\frac{\mathbb{E}[\eta^{2}]\sigma^{4}}{b}(d+b+1)=c>1,
\end{align*}
which implies that there exists some $0<\alpha<2$
such that $\tilde{h}\left(\alpha\right)=1$.

Finally, let us prove (ii) and (iii). 
When $c\leq 1$, 
we have $\tilde{h}(2)\leq 1$, which implies that
$\alpha\geq 2$. 
In particular, when $c=1$,   
the tail-index $\alpha=2$.
The proof is complete.
\hfill $\Box$

%%%%%%%%%%%%%%%%%%%%%%%%%%%%%%%%%%%%%%%%%%%%%%%%%%%%%%%%%%%%%%%%%%%%%%%%%%%
\subsection*{Proof of Lemma~\ref{lem:finite:bound:iid}}
We recall that 
\begin{equation}
x_{k}=M_{k}x_{k-1}+q_{k},
\end{equation}
which implies that
\begin{equation}
\Vert x_{k}\Vert
\leq\Vert M_{k}x_{k-1}\Vert
+\Vert q_{k}\Vert.
\end{equation}

(i) For any $p\leq 1$ and $h(p)<1$, by Lemma~\ref{lem:moment:ineq},
\begin{equation}
\Vert x_{k}\Vert^{p}
\leq\Vert M_{k}x_{k-1}\Vert^{p}
+\Vert q_{k}\Vert^{p}.
\end{equation}
Since $M_{k}$ is independent of $x_{k-1}$ and conditional on $x_{k-1}$ the distribution
of $\Vert M_{k}x_{k-1}\Vert$ is the same as $\Vert M_{k}e_{1}\Vert\cdot\Vert x_{k-1}\Vert$, we have
\begin{equation}
\mathbb{E}\Vert x_{k}\Vert^{p}
\leq
\mathbb{E}\Vert M_{k}e_{1}\Vert^{p}\mathbb{E}\Vert x_{k-1}\Vert^{p}
+\mathbb{E}\Vert q_{k}\Vert^{p},
\end{equation}
where $e_{1}$ is the first basis vector in $\mathbb{R}^{d}$, 
so that
\begin{equation}
\mathbb{E}\Vert x_{k}\Vert^{p}
\leq
h(p)\mathbb{E}\Vert x_{k-1}\Vert^{p}
+\mathbb{E}\Vert q_{1}\Vert^{p}.
\end{equation}
By iterating over $k$, we get
\begin{equation}
\mathbb{E}\Vert x_{k}\Vert^{p}
\leq
(h(p))^{k}\mathbb{E}\Vert x_{0}\Vert^{p}
+\frac{1-(h(p))^{k}}{1-h(p)}\mathbb{E}\Vert q_{1}\Vert^{p}.
\end{equation}

(ii) For any $p>1$ and $h(p)<1$,
by Lemma~\ref{lem:moment:ineq}, 
for any $\epsilon>0$, we have
\begin{equation}
\Vert x_{k}\Vert^{p}
\leq
(1+\epsilon)\Vert M_{k}x_{k-1}\Vert^{p}
+\frac{(1+\epsilon)^{\frac{p}{p-1}}-(1+\epsilon)}{\left((1+\epsilon)^{\frac{1}{p-1}}-1\right)^{p}}\Vert q_{k}\Vert^{p},
\end{equation}
which (similar as in (i)) implies that
\begin{equation}
\mathbb{E}\Vert x_{k}\Vert^{p}
\leq
(1+\epsilon)
\mathbb{E}\Vert M_{k}e_{1}\Vert^{p}
\mathbb{E}\Vert x_{k-1}\Vert^{p}
+\frac{(1+\epsilon)^{\frac{p}{p-1}}-(1+\epsilon)}{\left((1+\epsilon)^{\frac{1}{p-1}}-1\right)^{p}}
\mathbb{E}\Vert q_{k}\Vert^{p},
\end{equation}
so that
\begin{equation}
\mathbb{E}\Vert x_{k}\Vert^{p}
\leq
(1+\epsilon)h(p)
\mathbb{E}\Vert x_{k-1}\Vert^{p}
+\frac{(1+\epsilon)^{\frac{p}{p-1}}-(1+\epsilon)}{\left((1+\epsilon)^{\frac{1}{p-1}}-1\right)^{p}}
\mathbb{E}\Vert q_{1}\Vert^{p}.
\end{equation}
We choose $\epsilon>0$
so that $(1+\epsilon)h(p)<1$.
By iterating over $k$, we get
\begin{equation}
\mathbb{E}\Vert x_{k}\Vert^{p}
\leq
((1+\epsilon)h(p))^{k}\mathbb{E}\Vert x_{0}\Vert^{p}
+\frac{1-((1+\epsilon)h(p))^{k}}{1-(1+\epsilon)h(p)}
\frac{(1+\epsilon)^{\frac{p}{p-1}}-(1+\epsilon)}{\left((1+\epsilon)^{\frac{1}{p-1}}-1\right)^{p}}
\mathbb{E}\Vert q_{1}\Vert^{p}.
\end{equation}
The proof is complete.
\hfill $\Box$

%%%%%%%%%%%%%%%%%%%%%%%%%%%%%%%%%%%%%%%%%%%%%%%%%%%%%%%
\subsection*{Proof of Theorem~\ref{thm:convergence:iid}}
For any $\nu_{0},\tilde{\nu}_{0}\in\mathcal{P}_{p}(\mathbb{R}^{d})$, 
there exists a couple $x_{0}\sim\nu_{0}$ and $\tilde{x}_{0}\sim\tilde{\nu}_{0}$ independent
of $(M_{k},q_{k})_{k\in\mathbb{N}}$ and
$\mathcal{W}_{p}^{p}(\nu_{0},\tilde{\nu}_{0})=\mathbb{E}\Vert x_{0}-\tilde{x}_{0}\Vert^{p}$.
We define $x_{k}$ and $\tilde{x}_{k}$ starting from $x_{0}$ and $\tilde{x}_{0}$ respectively,
via the iterates
\begin{align}
&x_{k}=M_{k}x_{k-1}+q_{k},
\\
&\tilde{x}_{k}=M_{k}\tilde{x}_{k-1}+q_{k},
\end{align}
and let $\nu_{k}$ and $\tilde{\nu}_{k}$ denote
the probability laws of $x_{k}$ and $\tilde{x}_{k}$ respectively. 
For any $p\geq 1$, since $\mathbb{E}\Vert M_{k}\Vert^{p}<\infty$
and $\mathbb{E}\Vert q_{k}\Vert^{p}<\infty$, 
we have $\nu_{k},\tilde{\nu}_{k}\in\mathcal{P}_{p}(\mathbb{R}^{d})$
for any $k$. Moreover, we have
\begin{equation}\label{eqn:coupling:iid}
x_{k}-\tilde{x}_{k}=M_{k}(x_{k-1}-\tilde{x}_{k-1}),
\end{equation}
which yields that
\begin{align*}
\mathbb{E}\Vert x_{k}-\tilde{x}_{k}\Vert^{p}
&\leq
\mathbb{E}\left[\Vert M_{k}(x_{k-1}-\tilde{x}_{k-1})\Vert^{p}\right]
\\
&=\mathbb{E}\left[\Vert M_{k}e_{1}\Vert^{p}\Vert x_{k-1}-\tilde{x}_{k-1}\Vert^{p}\right]
\\
&=\mathbb{E}\left[\Vert M_{k}e_{1}\Vert^{p}\right]
\mathbb{E}\left[\Vert x_{k-1}-\tilde{x}_{k-1}\Vert^{p}\right]
=h(p)\mathbb{E}\left[\Vert x_{k-1}-\tilde{x}_{k-1}\Vert^{p}\right],
\end{align*}
where $e_{1}$
is the first basis vector in $\mathbb{R}^{d}$, which by iterating implies that
\begin{equation}
\mathcal{W}_{p}^{p}(\nu_{k},\tilde{\nu}_{k})
\leq
\mathbb{E}\Vert x_{k}-\tilde{x}_{k}\Vert^{p}
\leq
(h(p))^{k}
\mathbb{E}\Vert x_{0}-\tilde{x}_{0}\Vert^{p}
=(h(p))^{k}
\mathcal{W}_{p}^{p}(\nu_{0},\tilde{\nu}_{0}).
\end{equation}
By taking $\tilde{\nu}_{0}=\nu_{\infty}$, the probability
law of the stationary distribution $x_{\infty}$, 
we conclude that
\begin{equation}
\mathcal{W}_{p}(\nu_{k},\nu_{\infty})
\leq
\left(\left(h(p)\right)^{1/q}\right)^{k}
\mathcal{W}_{p}(\nu_{0},\nu_{\infty}).
\end{equation}
The proof is complete.
\hfill $\Box$

%%%%%%%%%%%%%%%%%%%%%%%%%%%%%%%%%%%%%

\subsection*{Proof of Corollary~\ref{cor:clt}}
The result is obtained by a direct application of Theorem~1.15 in \cite{mirek2011heavy} to the recursions \eqref{eq-stoc-grad-2-iid}, where it can be checked in a straightforward manner that the conditions for this theorem hold.
\hfill $\Box$

%%%%%%%%%%%%%%%%%%%%%%%%%%%%%%%%%%%%%%%%%%%%%%%%%%%%%%%%%%%%%%%%
\subsection{Proofs of Results in Section~\ref{sec:cyclic:appendix}}

%%%%%%%%%%%%%%%%%%%%%%%%%%%%%%%%%%%%%%%%%%%%%%%%%%%%%%

%%%%%%%%%%%%%%%%%%%%%%%%%%%%%%%%%%%%%%%%%%%%%%%%%%%%%%%
% \subsection*{Proof of Theorem~\ref{thm:cyclic:index}}
% It follows from the proof of Theorem~4 in \cite{HTP2021}
% that for any $s\geq 1$, 
% \begin{equation*}
% \mathbb{E}\left[\left\Vert I-\frac{\eta_{i}}{b}H\right\Vert^{s}\right]
% \end{equation*}
% is strictly decreasing in $b$. Therefore, $\hat{h}^{(m)}(s)$ is strictly decreasing in $b$.
% It thus follows from the arguments in the proof of Theorem~4 in \cite{HTP2021} that
% $\hat{\alpha}^{(m)}$ is strictly increasing in batch-size $b$
% provided that $\hat{\alpha}^{(m)}\geq 1$. The proof is complete.
% \hfill $\Box$

%%%%%%%%%%%%%%%%%%%%%%%%%%%%%%%%%%%%%%%%%%%%%%%%%%%%%
%%%%%%%%%%%%%%%%%%%%%%%%%%%%%%%%%%%%%%%%%%%%%%%%%%%%%

\subsection*{Proof of Proposition~\ref{alpha:2:cyclic}}
We first prove (i). 
Let us first recall from Lemma~\ref{lem:simplified:cyclic} that
\begin{align*}
&\tilde{h}^{(m)}(s)
=\left(\prod_{i=1}^{m}\mathbb{E}\left[\left(\left(1-\frac{\eta_{i}\sigma^{2}}{b}X\right)^{2}
+\frac{\eta_{i}^{2}\sigma^{4}}{b^{2}}XY\right)^{s/2}\right]\right)^{1/m},
\\
&\tilde{\rho}^{(m)}=\frac{1}{2}\sum_{i=1}^{m}\mathbb{E}\left[\log\left(\left(1-\frac{\eta_{i}\sigma^{2}}{b}X\right)^{2}+\frac{\eta_{i}^{2}\sigma^{4}}{b^{2}}XY\right)\right],
\end{align*}
where $X,Y$ are independent and $X$ is chi-square random variable
with degree of freedom $b$ and $Y$ is a chi-square random variable
with degree of freedom $(d-1)$.
When 
\begin{equation*}
c^{(m)}=\prod_{i=1}^{m}\left(1-2\eta_{i}\sigma^{2}+\frac{\eta_{i}^{2}\sigma^{4}}{b}(d+b+1)\right)=1,
\end{equation*}
we can compute that
\begin{align}
\tilde{\rho}^{(m)}
&\leq\frac{1}{2}\sum_{i=1}^{m}\log
\mathbb{E}\left[1- \frac{2\eta_{i}\sigma^2}{b}X
+ \frac{\eta_{i}^2 \sigma^4}{b^{2}}(X^{2}+XY)\right]
\label{tilde:rho:inequality:m}
\\
&=\frac{1}{2}\sum_{i=1}^{m}\log\left(1-2\eta_{i}\sigma^{2}+\frac{\eta_{i}^{2}\sigma^{4}}{b}(d+b+1)\right)=0.\nonumber
\end{align}
Note that since 
$1- \frac{2\eta_{i}\sigma^2}{b} X
+ \frac{\eta_{i}^2 \sigma^4}{b^{2}}(X^{2}+XY)$ is
random, the inequality in \eqref{tilde:rho:inequality:m} is a strict inequality
from Jensen's inequality. Thus, when 
 $c^{(m)}=1$, we have $\tilde{\rho}^{(m)}<0$. 
By continuity, there exists some $\delta>0$
such that for any  $1<c^{(m)}<1+\delta$
we have $\tilde{\rho}^{(m)}<0$. Moreover, when $c^{(m)}>1$,
we have
\begin{align*}
\left(h^{(m)}(2)\right)^{m}
&=\prod_{i=1}^{m}\mathbb{E}\left[1- \frac{2\eta_{i} \sigma^2}{b}X
+ \frac{\eta_{i}^2 \sigma^4}{b^{2}}(X^{2}+XY)\right]
\\
&=\prod_{i=1}^{m}\left(1-2\eta_{i}\sigma^{2}+\frac{\eta_{i}^{2}\sigma^{4}}{b}(d+b+1)\right)=c^{(m)}>1,
\end{align*}
which implies that there exists some $0<\alpha^{(m)}<2$
such that $h^{(m)}\left(\alpha^{(m)}\right)=1$.

Finally, let us prove (ii) and (iii). 
When $c^{(m)}\leq 1$, 
we have $\tilde{h}^{(m)}(2)\leq 1$, which implies that
$\alpha^{(m)}\geq 2$. 
In particular, when $c^{(m)}=1$,   
the tail-index $\alpha^{(m)}=2$.
The proof is complete.
\hfill $\Box$

\subsection*{Proof of Lemma~\ref{lem:finite:bound:cyclic}}
The proof is similar to the proof of Lemma~\ref{lem:finite:bound:iid} and is hence omitted here.
\hfill $\Box$

%%%%%%%%%%%%%%%%%%%%%%%%%%%%%%%%%%%%%%%%%%%%%%%%%%%%%%%
\subsection*{Proof of Theorem~\ref{thm:convergence:cyclic}}
The proof is similar to the proof of Theorem~\ref{thm:convergence:iid} and is hence omitted here.
\hfill $\Box$

%%%%%%%%%%%%%%%%%%%%%%%%%%%%%%%%%%%%%

\subsection*{Proof of Corollary~\ref{cor:clt:cyclic}}
The proof is similar to the proof of Corollary~\ref{cor:clt} and is hence omitted here.
\hfill $\Box$

%%%%%%%%%%%%%%%%%%%%%%%%%%%%%%%%%%%%%%%%%%%%%%%%%%%%%
\subsection{Proofs of Results in Section~\ref{subsec-appendix-markovian}}

%%%%%%%%%%%%%%%%%%%%%%%%%%%%%%%%%%%%%%%%%%%%%%%%%%%%%
\subsection*{Proof of Proposition~\ref{cor:Markovian}}

For any $p<\hat{\alpha}^{(g)}$, we have $\hat{h}^{(g)}(p)<1$. 
By Lemma~\ref{lem:finite:bound} and Fatou's lemma, we have
that
for any $p\leq 1$ and $\hat{h}^{(g)}(p)<1$,
\begin{equation}
\mathbb{E}\Vert x_{\infty}\Vert^{p}
\leq
\frac{1}{1-\hat{h}^{(g)}(p)}\mathbb{E}\Vert q_{1}\Vert^{p},
\end{equation}
and for any $p>1$, $\epsilon>0$ and $(1+\epsilon)\hat{h}^{(g)}(p)<1$,
\begin{equation}
\mathbb{E}\Vert x_{\infty}\Vert^{p}
\leq
\frac{1}{1-(1+\epsilon)\hat{h}^{(g)}(p)}
\frac{(1+\epsilon)^{\frac{p}{p-1}}-(1+\epsilon)}{\left((1+\epsilon)^{\frac{1}{p-1}}-1\right)^{p}}
\mathbb{E}\Vert q_{1}\Vert^{p}.
\end{equation}
Finally, by applying Chebyshev's inequality inequality, 
we complete the proof.
\hfill $\Box$
%%%%%%%%%%%%%%%%%%%%%%%%%%%%%%%%%%%%%%%
\subsection*{Proof of Theorem~\ref{thm:mono:Markovian}}
By following the proof of Theorem~4 in \cite{HTP2021}, it suffices
to show that for any $s\geq 1$, $\hat{h}^{(g)}(s)$ is decreasing in batch-size $b\in\mathbb{N}$.
By tower property,
\begin{equation}
\hat{h}^{(g)}(s)=\mathbb{E}\left[\hat{h}^{(g)}(s|\eta)\right],
\qquad
\hat{h}^{(g)}(s|\eta):=\mathbb{E}\left[\left\Vert I-\frac{\eta}{b}H\right\Vert^{s}\bigg|\eta\right].
\end{equation}
With slight abuse of notation, we define the function $\hat{h}^{(g)}(b,s|\eta)=\hat{h}^{(g)}(s|\eta)$ to emphasize the dependence on $b$. We have
\begin{equation}
\hat{h}^{(g)}(b,s|\eta)=\mathbb{E}\left[\left\Vert I-\frac{\eta}{b}\sum_{i=1}^{b}a_{i}a_{i}^{T}\right\Vert^{s}\bigg|\eta\right].
\end{equation}
When $s\geq 1$, the function $x\mapsto\Vert x\Vert^{s}$ is convex, 
and by Jensen's inequality, we get for any $b\geq 2$ and $b\in\mathbb{N}$,
\begin{align*}
\hat{h}^{(g)}(b,s|\eta)&=\mathbb{E}\left[\left\Vert\frac{1}{b}\sum_{i=1}^{b}\left(I-\frac{\eta}{b-1}\sum_{j\neq i}a_{j}a_{j}^{T}\right)\right\Vert^{s}\bigg|\eta\right]
\\
&\leq
\mathbb{E}\left[\frac{1}{b}\sum_{i=1}^{b}\left\Vert I-\frac{\eta}{b-1}\sum_{j\neq i}a_{j}a_{j}^{T}\right\Vert^{s}\bigg|\eta\right]
\\
&=\frac{1}{b}\sum_{i=1}^{b}\mathbb{E}\left[\left\Vert I-\frac{\eta}{b-1}\sum_{j\neq i}a_{j}a_{j}^{T}\right\Vert^{s}\bigg|\eta\right]
=\hat{h}^{(g)}(b-1,s|\eta),
\end{align*}
where we used the fact that $a_{i}$ are i.i.d. independent of the distribution of $\eta$.
Indeed, from the condition for equality to hold in Jensen's inequality,
and the fact that $a_{i}$ are i.i.d. random,
the inequality above is a strict inequality.
Hence when $d\in\mathbb{N}$ for any $s\geq 1$, $\hat{h}^{(g)}(b,s|\eta)$ is strictly decreasing in $b$.
Since $\hat{h}^{(g)}(s)=\mathbb{E}[\hat{h}^{(g)}(s|\eta)]$, we conclude that
$\hat{h}^{(g)}(s)$ is decreasing in batch-size $b\in\mathbb{N}$.
Hence, by following the same arguments as in the proof of Theorem~4 in \cite{HTP2021}, we
conclude that the lower bound for the tail-index $\hat{\alpha}^{(g)}$ is strictly increasing in batch-size $b$
provided that $\hat{\alpha}^{(g)}\geq 1$.

Moreover, by adapting the proof of Lemma~\ref{lemma-cvx-in-step} (Lemma~22 in \cite{HTP2021}), 
one can show that for any given positive semi-definite symmetric matrix $H$ fixed, the function $F_H:[0,\infty) \to \mathbb{R}$ defined as
$ F_H(a) :=\left\| \left(I - a H\right)\right\|^s$ 
is convex for $s\geq 1$. The rest of the proof follows from the similar arguments as in the proof of Theorem~\ref{thm:mono:Gaussian}. The proof is complete.
\hfill $\Box$

%%%%%%%%%%%%%%%%%%%%%%%%%%%%%%%%%%%%%%%%%%%%%%%%%%%%%%%%%%%%%%%%%%%%

\subsection*{Proof of Proposition~\ref{alpha:2:Markov}}
We first prove (i). 
Let us first recall from Lemma~\ref{lem:simplified:Markov:regeneration} that
\begin{align*}
&\tilde{h}^{(r)}(s)
=\mathbb{E}\left[\prod_{i=1}^{r_{1}}\mathbb{E}_{X,Y}\left[\left(\left(1-\frac{\eta_{i}\sigma^{2}}{b}X\right)^{2}
+\frac{\eta_{i}^{2}\sigma^{4}}{b^{2}}XY\right)^{s/2}\right]\right],
\\
&\tilde{\rho}^{(r)}
=\frac{1}{2}\mathbb{E}\left[\sum_{i=1}^{r_{1}}\mathbb{E}_{X,Y}\left[\log\left(\left(1-\frac{\eta_{i}\sigma^{2}}{b}X\right)^{2}
+\frac{\eta_{i}^{2}\sigma^{4}}{b^{2}}XY\right)\right]\right],
\end{align*}
where $r_{1}$ is defined in \eqref{defn:regeneration:time}, 
and $X,Y$ are independent and $X$ is chi-square random variable
with degree of freedom $b$ and $Y$ is a chi-square random variable
with degree of freedom $(d-1)$.
When $c^{(r)}=\mathbb{E}\left[\prod_{i=1}^{r_{1}}\left(1-2\eta_{i}\sigma^{2}+\frac{\eta_{i}^{2}\sigma^{4}}{b}(d+b+1)\right)\right]=1$, 
we can compute that
\begin{align}
\tilde{\rho}^{(r)}
&\leq\frac{1}{2}\mathbb{E}\left[\sum_{i=1}^{r_{1}}\log\mathbb{E}_{X,Y}\left[\left(\left(1-\frac{\eta_{i}\sigma^{2}}{b}X\right)^{2}
+\frac{\eta_{i}^{2}\sigma^{4}}{b^{2}}XY\right)\right]\right]
\label{tilde:rho:inequality:r}
\\
&=\frac{1}{2}\mathbb{E}\left[\sum_{i=1}^{r_{1}}\log\left(1-2\eta_{i}\sigma^{2}+\frac{\eta_{i}^{2}\sigma^{4}}{b}(d+b+1)\right)\right]=0.\nonumber
\end{align}
Note that since 
$1- \frac{2\eta_{i}\sigma^2}{b} X
+ \frac{\eta_{i}^2 \sigma^4}{b^{2}}(X^{2}+XY)$
is
random, the inequality in \eqref{tilde:rho:inequality:r} is a strict inequality
from Jensen's inequality. Thus, when 
 $c^{(r)}=1$, we have $\tilde{\rho}^{(r)}<0$. 
By continuity, there exists some $\delta>0$
such that for any  $1<c^{(r)}<1+\delta$
we have $\tilde{\rho}^{(r)}<0$. Moreover, when $c^{(r)}>1$,
we have
\begin{align*}
h^{(r)}(2)
&=\mathbb{E}\left[\prod_{i=1}^{r_{1}}\mathbb{E}\left[1- \frac{2\eta_{i} \sigma^2}{b}X
+ \frac{\eta_{i}^2 \sigma^4}{b^{2}}(X^{2}+XY)\right]\right]
\\
&=\mathbb{E}\left[\prod_{i=1}^{r_{1}}\left(1-2\eta_{i}\sigma^{2}+\frac{\eta_{i}^{2}\sigma^{4}}{b}(d+b+1)\right)\right]=c^{(r)}>1,
\end{align*}
which implies that there exists some $0<\alpha^{(r)}<2$
such that $h^{(r)}\left(\alpha^{(r)}\right)=1$.

Finally, let us prove (ii) and (iii). 
When $c^{(r)}\leq 1$, 
we have $\tilde{h}^{(r)}(2)\leq 1$, which implies that
$\alpha^{(r)}\geq 2$. 
In particular, when $c^{(r)}=1$,   
the tail-index $\alpha^{(r)}=2$.
The proof is complete.
\hfill $\Box$
%%%%%%%%%%%%%%%%%%%%%%%%%%%%%%%%%%%%%%%%%%%%%%%%%%%%%%%
\subsection*{Proof of Lemma~\ref{lem:finite:bound}}
We recall that 
\begin{equation}
x_{k}=M_{k}x_{k-1}+q_{k},
\end{equation}
which implies that
\begin{equation}
\Vert x_{k}\Vert
\leq\Vert M_{k}x_{k-1}\Vert
+\Vert q_{k}\Vert.
\end{equation}

(i) For any $p\leq 1$ and $\hat{h}^{(g)}(p)<1$, by Lemma~\ref{lem:moment:ineq},
\begin{equation}
\Vert x_{k}\Vert^{p}
\leq\Vert M_{k}x_{k-1}\Vert^{p}
+\Vert q_{k}\Vert^{p}.
\end{equation}
Since $M_{k}$ is independent of $x_{k-1}$, we have
\begin{equation}
\mathbb{E}\Vert x_{k}\Vert^{p}
\leq
\mathbb{E}\Vert M_{k}\Vert^{p}\mathbb{E}\Vert x_{k-1}\Vert^{p}
+\mathbb{E}\Vert q_{k}\Vert^{p},
\end{equation}
so that
\begin{equation}
\mathbb{E}\Vert x_{k}\Vert^{p}
\leq
\hat{h}^{(g)}(p)\mathbb{E}\Vert x_{k-1}\Vert^{p}
+\mathbb{E}\Vert q_{1}\Vert^{p}.
\end{equation}
By iterating over $k$, we get
\begin{equation}
\mathbb{E}\Vert x_{k}\Vert^{p}
\leq
(\hat{h}^{(g)}(p))^{k}\mathbb{E}\Vert x_{0}\Vert^{p}
+\frac{1-(\hat{h}^{(g)}(p))^{k}}{1-\hat{h}^{(g)}(p)}\mathbb{E}\Vert q_{1}\Vert^{p}.
\end{equation}

(ii) For any $p>1$ and $\hat{h}^{(g)}(p)<1$,
by Lemma~\ref{lem:moment:ineq}, 
for any $\epsilon>0$, we have
\begin{equation}
\Vert x_{k}\Vert^{p}
\leq
(1+\epsilon)\Vert M_{k}x_{k-1}\Vert^{p}
+\frac{(1+\epsilon)^{\frac{p}{p-1}}-(1+\epsilon)}{\left((1+\epsilon)^{\frac{1}{p-1}}-1\right)^{p}}\Vert q_{k}\Vert^{p},
\end{equation}
which (similar as in (i)) implies that
\begin{equation}
\mathbb{E}\Vert x_{k}\Vert^{p}
\leq
(1+\epsilon)
\mathbb{E}\Vert M_{k}\Vert^{p}
\mathbb{E}\Vert x_{k-1}\Vert^{p}
+\frac{(1+\epsilon)^{\frac{p}{p-1}}-(1+\epsilon)}{\left((1+\epsilon)^{\frac{1}{p-1}}-1\right)^{p}}
\mathbb{E}\Vert q_{k}\Vert^{p},
\end{equation}
so that
\begin{equation}
\mathbb{E}\Vert x_{k}\Vert^{p}
\leq
(1+\epsilon)\hat{h}^{(g)}(p)
\mathbb{E}\Vert x_{k-1}\Vert^{p}
+\frac{(1+\epsilon)^{\frac{p}{p-1}}-(1+\epsilon)}{\left((1+\epsilon)^{\frac{1}{p-1}}-1\right)^{p}}
\mathbb{E}\Vert q_{1}\Vert^{p}.
\end{equation}
We choose $\epsilon>0$
so that $(1+\epsilon)\hat{h}^{(g)}(p)<1$.
By iterating over $k$, we get
\begin{equation}
\mathbb{E}\Vert x_{k}\Vert^{p}
\leq
((1+\epsilon)\hat{h}^{(g)}(p))^{k}\mathbb{E}\Vert x_{0}\Vert^{p}
+\frac{1-((1+\epsilon)\hat{h}^{(g)}(p))^{k}}{1-(1+\epsilon)\hat{h}^{(g)}(p)}
\frac{(1+\epsilon)^{\frac{p}{p-1}}-(1+\epsilon)}{\left((1+\epsilon)^{\frac{1}{p-1}}-1\right)^{p}}
\mathbb{E}\Vert q_{1}\Vert^{p}.
\end{equation}
The proof is complete.
\hfill $\Box$

%%%%%%%%%%%%%%%%%%%%%%%%%%%%%%%%%%%%%%%%%%%%%%%%%%%%%%%
\subsection*{Proof of Theorem~\ref{thm:convergence}}
For any $\nu_{0},\tilde{\nu}_{0}\in\mathcal{P}_{p}(\mathbb{R}^{d})$, 
there exists a couple $x_{0}\sim\nu_{0}$ and $\tilde{x}_{0}\sim\tilde{\nu}_{0}$ independent
of $(M_{k},q_{k})_{k\in\mathbb{N}}$ and
$\mathcal{W}_{p}^{p}(\nu_{0},\tilde{\nu}_{0})=\mathbb{E}\Vert x_{0}-\tilde{x}_{0}\Vert^{p}$.
We define $x_{k}$ and $\tilde{x}_{k}$ starting from $x_{0}$ and $\tilde{x}_{0}$ respectively,
via the iterates
\begin{align}
&x_{k}=M_{k}x_{k-1}+q_{k},
\\
&\tilde{x}_{k}=M_{k}\tilde{x}_{k-1}+q_{k},
\end{align}
and let $\nu_{k}$ and $\tilde{\nu}_{k}$ denote
the probability laws of $x_{k}$ and $\tilde{x}_{k}$ respectively. 
For any $p\geq 1$, since $\mathbb{E}\Vert M_{k}\Vert^{p}<\infty$
and $\mathbb{E}\Vert q_{k}\Vert^{p}<\infty$, 
we have $\nu_{k},\tilde{\nu}_{k}\in\mathcal{P}_{p}(\mathbb{R}^{d})$
for any $k$. Moreover, we have
\begin{equation}\label{eqn:coupling}
x_{k}-\tilde{x}_{k}=M_{k}(x_{k-1}-\tilde{x}_{k-1}),
\end{equation}
which yields that
\begin{align*}
\mathbb{E}\Vert x_{k}-\tilde{x}_{k}\Vert^{p}
&\leq
\mathbb{E}\left[\Vert M_{k}(x_{k-1}-\tilde{x}_{k-1})\Vert^{p}\right]
\\
&\leq\mathbb{E}\left[\Vert M_{k}\Vert^{p}\right]
\mathbb{E}\left[\Vert x_{k-1}-\tilde{x}_{k-1}\Vert^{p}\right]
=\hat{h}^{(g)}(p)\mathbb{E}\left[\Vert x_{k-1}-\tilde{x}_{k-1}\Vert^{p}\right],
\end{align*}
which by iterating implies that
\begin{equation}
\mathcal{W}_{p}^{p}(\nu_{k},\tilde{\nu}_{k})
\leq
\mathbb{E}\Vert x_{k}-\tilde{x}_{k}\Vert^{p}
\leq
(\hat{h}^{(g)}(p))^{k}
\mathbb{E}\Vert x_{0}-\tilde{x}_{0}\Vert^{p}
=(\hat{h}^{(g)}(p))^{k}
\mathcal{W}_{p}^{p}(\nu_{0},\tilde{\nu}_{0}).
\end{equation}
By taking $\tilde{\nu}_{0}=\nu_{\infty}$, the probability
law of the stationary distribution $x_{\infty}$, 
we conclude that
\begin{equation}
\mathcal{W}_{p}(\nu_{k},\nu_{\infty})
\leq
\left(\left(\hat{h}^{(g)}(p)\right)^{1/q}\right)^{k}
\mathcal{W}_{p}(\nu_{0},\nu_{\infty}).
\end{equation}
The proof is complete.
\hfill $\Box$

%%%%%%%%%%%%%%%%%%%%%%%%%%%%%%%%%%%%%%%%%%%%%%%%%%%%%

\subsection*{Proof of Theorem~\ref{thm:mono:Markovian:range}}
When the stationary distribution of the Markovian stepsizes is uniform
on the set \eqref{state:space}, we have
\begin{equation}
\hat{h}^{(g)}(s)=\frac{1}{K}\mathbb{E}\left[\left\Vert I-\frac{\bar{\eta}}{b}H\right\Vert^{s}\right]
+\frac{1}{K}\sum_{j=1}^{\frac{K-1}{2}}
\left(\mathbb{E}\left[\left\Vert I-\frac{\bar{\eta}-j\delta}{b}H\right\Vert^{s}\right]
+\mathbb{E}\left[\left\Vert I-\frac{\bar{\eta}+j\delta}{b}H\right\Vert^{s}\right]\right).
\end{equation}
It suffices to show that for any $s\geq 1$, 
$\hat{h}^{(g)}(s)$ is increasing in $\delta$.
It suffices to show that for any $s\geq 1$ and $j=1,\ldots,\frac{K-1}{2}$, 
\begin{equation}
\hat{h}^{(g)}_{j}(s):=\mathbb{E}\left[\left\Vert I-\frac{\bar{\eta}-j\delta}{b}H\right\Vert^{s}\right]
+\mathbb{E}\left[\left\Vert I-\frac{\bar{\eta}+j\delta}{b}H\right\Vert^{s}\right]
\end{equation}
is increasing in $\delta$.
By adapting the proof of Lemma~\ref{lemma-cvx-in-step} (Lemma~22 in \cite{HTP2021}), 
one can show that the function
\begin{equation}\label{f:x}
f(x):=\mathbb{E}\left[\left\Vert I-\frac{x}{b}H\right\Vert^{s}\right]
\end{equation}
is convex in $x$ for any $s\geq 1$. 
It remains to show that $f(\bar{\eta}-j\delta)+f(\bar{\eta}+j\delta)$ is increasing in $\delta$. 
We claim that
\begin{equation}
F(x;a):=f(x-a)+f(x+a)
\end{equation}
is increasing in $x$ for any $x\geq a>0$. 
To see this, we can compute that $F'(a;a)=0$
and $F''(x;a)=f''(x-a)+f''(x+a)\geq 0$ since $f(x)$ is convex in $x$, 
which implies that $F'(x;a)\geq 0$ for any $x\geq a$
and thus $F(x;a)$ is increasing in $x$ for any $x\geq a>0$. 
Hence, the lower bound for the tail-index $\hat{\alpha}^{(g)}$ is decreasing $\delta$ 
provided that $\hat{\alpha}^{(g)}\geq 1$.

Next, let us show that $\hat{\alpha}^{(g)}$ is increasing in $K$ (where we recall that $K$ is odd without loss of generality)
for any $\hat{\alpha}^{(g)}\geq 1$. Let $\hat{h}^{(g)}(s;K)=\hat{h}^{(g)}(s)$ that emphasizes the dependence on $K$.
Let us show that $\hat{h}^{(g)}(s;K+2)\geq\hat{h}^{(g)}(s;K)$ for any odd $K$ and $s\geq 1$. 
We can compute that
\begin{align*}
\hat{h}^{(g)}(s;K+2)-\hat{h}^{(g)}(s;K)
&=\left(\frac{1}{K+2}-\frac{1}{K}\right)\mathbb{E}\left[\left\Vert I-\frac{\bar{\eta}}{b}H\right\Vert^{s}\right]
\\
&\quad
+\left(\frac{1}{K+2}-\frac{1}{K}\right)\sum_{j=1}^{\frac{K-1}{2}}
\left(\mathbb{E}\left[\left\Vert I-\frac{\bar{\eta}-j\delta}{b}H\right\Vert^{s}\right]
+\mathbb{E}\left[\left\Vert I-\frac{\bar{\eta}+j\delta}{b}H\right\Vert^{s}\right]\right)
\\
&\qquad
+\frac{1}{K+2}
\left(\mathbb{E}\left[\left\Vert I-\frac{\bar{\eta}-\frac{K+1}{2}\delta}{b}H\right\Vert^{s}\right]
+\mathbb{E}\left[\left\Vert I-\frac{\bar{\eta}+\frac{K+1}{2}\delta}{b}H\right\Vert^{s}\right]\right).
\end{align*}
Therefore, it suffices to show that
\begin{align}
&\mathbb{E}\left[\left\Vert I-\frac{\bar{\eta}-\frac{K+1}{2}\delta}{b}H\right\Vert^{s}\right]
+\mathbb{E}\left[\left\Vert I-\frac{\bar{\eta}+\frac{K+1}{2}\delta}{b}H\right\Vert^{s}\right]
\nonumber
\\
&\geq
\frac{2}{K}\mathbb{E}\left[\left\Vert I-\frac{\bar{\eta}}{b}H\right\Vert^{s}\right]
+\frac{2}{K}\sum_{j=1}^{\frac{K-1}{2}}
\left(\mathbb{E}\left[\left\Vert I-\frac{\bar{\eta}-j\delta}{b}H\right\Vert^{s}\right]
+\mathbb{E}\left[\left\Vert I-\frac{\bar{\eta}+j\delta}{b}H\right\Vert^{s}\right]\right).
\label{claim:0}
\end{align}
Since the function $f(x)$ defined in \eqref{f:x} is convex
for any $s\geq 1$, 
for any $j=0,1,2,\ldots,\frac{K-1}{2}$, 
\begin{align}
&\mathbb{E}\left[\left\Vert I-\frac{\bar{\eta}-j\delta}{b}H\right\Vert^{s}\right]
+\mathbb{E}\left[\left\Vert I-\frac{\bar{\eta}+j\delta}{b}H\right\Vert^{s}\right]
\nonumber
\\
&\leq
\mathbb{E}\left[\left\Vert I-\frac{\bar{\eta}-\frac{K+1}{2}\delta}{b}H\right\Vert^{s}\right]
+\mathbb{E}\left[\left\Vert I-\frac{\bar{\eta}+\frac{K+1}{2}\delta}{b}H\right\Vert^{s}\right],
\end{align}
which implies that
\begin{align*}
&\frac{2}{K}\mathbb{E}\left[\left\Vert I-\frac{\bar{\eta}}{b}H\right\Vert^{s}\right]
+\frac{2}{K}\sum_{j=1}^{\frac{K-1}{2}}
\left(\mathbb{E}\left[\left\Vert I-\frac{\bar{\eta}-j\delta}{b}H\right\Vert^{s}\right]
+\mathbb{E}\left[\left\Vert I-\frac{\bar{\eta}+j\delta}{b}H\right\Vert^{s}\right]\right)
\\
&\leq
\left(\frac{1}{K}+\frac{2}{K}\frac{K-1}{2}\right)
\left(\mathbb{E}\left[\left\Vert I-\frac{\bar{\eta}-\frac{K+1}{2}\delta}{b}H\right\Vert^{s}\right]
+\mathbb{E}\left[\left\Vert I-\frac{\bar{\eta}+\frac{K+1}{2}\delta}{b}H\right\Vert^{s}\right]\right)
\\
&=\mathbb{E}\left[\left\Vert I-\frac{\bar{\eta}-\frac{K+1}{2}\delta}{b}H\right\Vert^{s}\right]
+\mathbb{E}\left[\left\Vert I-\frac{\bar{\eta}+\frac{K+1}{2}\delta}{b}H\right\Vert^{s}\right],
\end{align*}
which proves \eqref{claim:0}.
Hence, the lower bound for the tail-index $\hat{\alpha}^{(g)}$ is decreasing $K$ 
provided that $\hat{\alpha}^{(g)}\geq 1$.
The proof is complete.
\hfill $\Box$

%%%%%%%%%%%%%%%%%%%%%%%%%%%%%%%%%%%%%%%%%%%%%%%%%%%%%%

%%%%%%%%%%%%%%%%%%%%%%%%%%%%%%%%%%%%%%%%%%%%%%%%%%%%%

\subsection*{Proof of Theorem~\ref{thm:mono:Markovian:range:2}}
When the stationary distribution of the Markovian stepsizes is uniform
on the set \eqref{state:space:2}, we have
\begin{align*}
\hat{h}^{(g)}(s)&=\frac{1}{2^{n}+1}\mathbb{E}\left[\left\Vert I-\frac{\bar{\eta}}{b}H\right\Vert^{s}\right]
\\
&\qquad+\frac{1}{2^{n}+1}\sum_{j=1}^{2^{n-1}}
\left(\mathbb{E}\left[\left\Vert I-\frac{\bar{\eta}-j\frac{R}{2^{n-1}}}{b}H\right\Vert^{s}\right]
+\mathbb{E}\left[\left\Vert I-\frac{\bar{\eta}+j\frac{R}{2^{n-1}}}{b}H\right\Vert^{s}\right]\right).
\end{align*}
Let us use the notation $\hat{h}^{(g)}(s;n):=\hat{h}^{(g)}(s)$
to emphasize the dependence on $n$. 
We can compute that
\begin{align*}
&\hat{h}^{(g)}(s;n)-\hat{h}^{(g)}(s;n+1)    
\\
&=\left(\frac{1}{2^{n}+1}-\frac{1}{2^{n+1}+1}\right)\mathbb{E}\left[\left\Vert I-\frac{\bar{\eta}}{b}H\right\Vert^{s}\right]
\\
&\qquad
+\frac{1}{2^{n}+1}\sum_{j=1}^{2^{n-1}}
\left(\mathbb{E}\left[\left\Vert I-\frac{\bar{\eta}-j\frac{R}{2^{n-1}}}{b}H\right\Vert^{s}\right]
+\mathbb{E}\left[\left\Vert I-\frac{\bar{\eta}+j\frac{R}{2^{n-1}}}{b}H\right\Vert^{s}\right]\right)
\\
&\qquad
-\frac{1}{2^{n+1}+1}\sum_{j=1}^{2^{n}}
\left(\mathbb{E}\left[\left\Vert I-\frac{\bar{\eta}-j\frac{R}{2^{n}}}{b}H\right\Vert^{s}\right]
+\mathbb{E}\left[\left\Vert I-\frac{\bar{\eta}+j\frac{R}{2^{n}}}{b}H\right\Vert^{s}\right]\right)
\\
&=\left(\frac{1}{2^{n}+1}-\frac{1}{2^{n+1}+1}\right)\mathbb{E}\left[\left\Vert I-\frac{\bar{\eta}}{b}H\right\Vert^{s}\right]
\\
&\qquad
+\left(\frac{1}{2^{n}+1}-\frac{1}{2^{n+1}+1}\right)\sum_{j=1}^{2^{n-1}}
\left(\mathbb{E}\left[\left\Vert I-\frac{\bar{\eta}-j\frac{R}{2^{n-1}}}{b}H\right\Vert^{s}\right]
+\mathbb{E}\left[\left\Vert I-\frac{\bar{\eta}+j\frac{R}{2^{n-1}}}{b}H\right\Vert^{s}\right]\right)
\\
&\qquad
-\frac{1}{2^{n+1}+1}\sum_{j=1}^{2^{n-1}}
\left(\mathbb{E}\left[\left\Vert I-\frac{\bar{\eta}-(2j-1)\frac{R}{2^{n}}}{b}H\right\Vert^{s}\right]
+\mathbb{E}\left[\left\Vert I-\frac{\bar{\eta}+(2j-1)\frac{R}{2^{n}}}{b}H\right\Vert^{s}\right]\right).
\end{align*}
By adapting the proof of Lemma~\ref{lemma-cvx-in-step} (Lemma~22 in \cite{HTP2021}), 
one can show that the function
\begin{equation}
f(x):=\mathbb{E}\left[\left\Vert I-\frac{x}{b}H\right\Vert^{s}\right]
\end{equation}
is convex in $x$ for any $s\geq 1$. 
Therefore, by Jensen's inequality,
\begin{align*}
\mathbb{E}\left[\left\Vert I-\frac{\bar{\eta}-(2j-1)\frac{R}{2^{n}}}{b}H\right\Vert^{s}\right]
\leq
\frac{1}{2}
\mathbb{E}\left[\left\Vert I-\frac{\bar{\eta}-(j-1)\frac{R}{2^{n-1}}}{b}H\right\Vert^{s}\right]
+\frac{1}{2}\mathbb{E}\left[\left\Vert I-\frac{\bar{\eta}-j\frac{R}{2^{n-1}}}{b}H\right\Vert^{s}\right],
\end{align*}
and similarly
\begin{align*}
\mathbb{E}\left[\left\Vert I-\frac{\bar{\eta}+(2j-1)\frac{R}{2^{n}}}{b}H\right\Vert^{s}\right]
\leq
\frac{1}{2}
\mathbb{E}\left[\left\Vert I-\frac{\bar{\eta}+(j-1)\frac{R}{2^{n-1}}}{b}H\right\Vert^{s}\right]
+\frac{1}{2}\mathbb{E}\left[\left\Vert I-\frac{\bar{\eta}+j\frac{R}{2^{n-1}}}{b}H\right\Vert^{s}\right],
\end{align*}
which implies that
\begin{align*}
&\hat{h}^{(g)}(s;n)-\hat{h}^{(g)}(s;n+1)
\\
&\geq\left(\frac{1}{2^{n}+1}-\frac{2}{2^{n+1}+1}\right)\mathbb{E}\left[\left\Vert I-\frac{\bar{\eta}}{b}H\right\Vert^{s}\right]
\\
&\quad
+\left(\frac{1}{2^{n}+1}-\frac{2}{2^{n+1}+1}\right)\sum_{j=1}^{2^{n-1}-1}
\left(\mathbb{E}\left[\left\Vert I-\frac{\bar{\eta}-j\frac{R}{2^{n-1}}}{b}H\right\Vert^{s}\right]
+\mathbb{E}\left[\left\Vert I-\frac{\bar{\eta}+j\frac{R}{2^{n-1}}}{b}H\right\Vert^{s}\right]\right)
\\
&\quad
+\left(\frac{1}{2^{n}+1}-\frac{\frac{3}{2}}{2^{n+1}+1}\right)
\left(\mathbb{E}\left[\left\Vert I-\frac{\bar{\eta}-2^{n-1}\frac{R}{2^{n-1}}}{b}H\right\Vert^{s}\right]
+\mathbb{E}\left[\left\Vert I-\frac{\bar{\eta}+2^{n-1}\frac{R}{2^{n-1}}}{b}H\right\Vert^{s}\right]\right)
\\
&=-\frac{1}{(2^{n}+1)(2^{n+1}+1)}\mathbb{E}\left[\left\Vert I-\frac{\bar{\eta}}{b}H\right\Vert^{s}\right]
\\
&\quad
-\frac{1}{(2^{n}+1)(2^{n+1}+1)}\sum_{j=1}^{2^{n-1}-1}
\left(\mathbb{E}\left[\left\Vert I-\frac{\bar{\eta}-j\frac{R}{2^{n-1}}}{b}H\right\Vert^{s}\right]
+\mathbb{E}\left[\left\Vert I-\frac{\bar{\eta}+j\frac{R}{2^{n-1}}}{b}H\right\Vert^{s}\right]\right)
\\
&\quad
+\left(\frac{1}{2^{n}+1}-\frac{\frac{3}{2}}{2^{n+1}+1}\right)
\left(\mathbb{E}\left[\left\Vert I-\frac{\bar{\eta}-2^{n-1}\frac{R}{2^{n-1}}}{b}H\right\Vert^{s}\right]
+\mathbb{E}\left[\left\Vert I-\frac{\bar{\eta}+2^{n-1}\frac{R}{2^{n-1}}}{b}H\right\Vert^{s}\right]\right).
\end{align*}
Since we proved in the proof of Theorem~\ref{thm:mono:Markovian:range}
that $f(x-a)+f(x+a)$ is increasing in $x$ for any $x\geq a>0$, we have
\begin{align*}
&\hat{h}^{(g)}(s;n)-\hat{h}^{(g)}(s;n+1)
\\
&\geq-\frac{1}{(2^{n}+1)(2^{n+1}+1)}\frac{1}{2}\left(\mathbb{E}\left[\left\Vert I-\frac{\bar{\eta}-2^{n-1}\frac{R}{2^{n-1}}}{b}H\right\Vert^{s}\right]
+\mathbb{E}\left[\left\Vert I-\frac{\bar{\eta}+2^{n-1}\frac{R}{2^{n-1}}}{b}H\right\Vert^{s}\right]\right)
\\
&\qquad
-\frac{1}{(2^{n}+1)(2^{n+1}+1)}
\cdot\sum_{j=1}^{2^{n-1}-1}
\left(\mathbb{E}\left[\left\Vert I-\frac{\bar{\eta}-2^{n-1}\frac{R}{2^{n-1}}}{b}H\right\Vert^{s}\right]
+\mathbb{E}\left[\left\Vert I-\frac{\bar{\eta}+2^{n-1}\frac{R}{2^{n-1}}}{b}H\right\Vert^{s}\right]\right)
\\
&\qquad
+\left(\frac{1}{2^{n}+1}-\frac{\frac{3}{2}}{2^{n+1}+1}\right)
\cdot
\left(\mathbb{E}\left[\left\Vert I-\frac{\bar{\eta}-2^{n-1}\frac{R}{2^{n-1}}}{b}H\right\Vert^{s}\right]
+\mathbb{E}\left[\left\Vert I-\frac{\bar{\eta}+2^{n-1}\frac{R}{2^{n-1}}}{b}H\right\Vert^{s}\right]\right)=0.
\end{align*}
Hence $\hat{h}^{(g)}(s;n)$ is decreasing in $n$ provided that $s\geq 1$
and therefore the lower bound for the tail-index $\hat{\alpha}^{(g)}$ is increasing in $n$
provided that $\hat{\alpha}^{(g)}\geq 1$.
This completes the proof.
\hfill $\Box$

%%%%%%%%%%%%%%%%%%%%%%%%%%%%%%%%%%%%%%%%%%%%%%%%%%%%%%

%%%%%%%%%%%%%%%%%%%%%%%%%%%%%%%%%%%%%%%%%%
\subsection*{Proof of Proposition~\ref{thm:Markovian:cyclic}}
Under the assumption that the stationary distribution of the Markovian stepsizes is uniform
on the set \eqref{state:space:m}, we have
\begin{equation}
\mathbb{P}(\eta=\eta_{i})=\frac{1}{m},\qquad i=1,2,\ldots,m,
\end{equation}
so that
\begin{equation}
\hat{h}^{(g)}(s)=\mathbb{E}\left[\left\Vert I-\frac{\eta}{b}H\right\Vert^{s}\right]=
\frac{1}{m}\sum_{i=1}^{m}\mathbb{E}\left[\left\Vert I-\frac{\eta_{i}}{b}H\right\Vert^{s}\right].
\end{equation}
On the other hand, we recall that the lower bound for the tail-index $\hat{\alpha}^{(m)}$ for the SGD with cyclic stepsizes is the unique
positive value such that $\hat{h}^{(m)}\left(\hat{\alpha}^{(m)}\right)=1$.
By the inequality of arithmetic and geometric means, we obtain
\begin{equation}
\hat{h}^{(m)}(s)\leq\frac{1}{m}\sum_{i=1}^{m}\mathbb{E}\left[\left\Vert I-\frac{\eta_{i}}{b}H\right\Vert^{s}\right]=\hat{h}^{(g)}(s).
\end{equation}
Since $\eta_{i}$ is not constant, the above inequality is strict.
Therefore, we conclude that the lower bound for the tail-index $\hat{\alpha}^{(g)}$ is strictly less than
the lower bound for the tail-index $\hat{\alpha}^{(m)}$ for SGD with cyclic stepsizes.
The proof is complete.
\hfill $\Box$

%%%%%%%%%%%%%%%%%%%%%%%%%%%%%%%%%%%%%%%%%%%%%%%%%%%%%%%%%%%%%%%%%%%%%%%%%%%
\subsection*{Proof of Lemma~\ref{lem:finite:bound:Markov:Gaussian}}
The proof is similar to the proof of Lemma~\ref{lem:finite:bound:iid} and is hence omitted here.
\hfill $\Box$

%%%%%%%%%%%%%%%%%%%%%%%%%%%%%%%%%%%%%%%%%%%%%%%%%%%%%%%
\subsection*{Proof of Theorem~\ref{thm:convergence:Markov:Gaussian}}
The proof is similar to the proof of Theorem~\ref{thm:convergence:iid} and is hence omitted here.
\hfill $\Box$

%%%%%%%%%%%%%%%%%%%%%%%%%%%%%%%%%%%%%

\subsection*{Proof of Corollary~\ref{cor:clt:Markovian}}
The proof is similar to the proof of Corollary~\ref{cor:clt} and is hence omitted here.
\hfill $\Box$

%%%%%%%%%%%%%%%%%%%%%%%%%%%%%%%%%%%%%%%%%%%%%%%%%%%%%%%%%%%%%%%%%%%%%%%%%

\subsection*{Proof of Lemma~\ref{lem:stationary:dist}}
First of all, the Markov chain exhibits a unique stationary distribution
$\pi_{i}:=\mathbb{P}(\eta_{0}=\eta_{i})$ that satisfy the equations:
\begin{align*}
&\pi_{1}=(1-p)\pi_{2}+p\pi_{m},
\qquad
\pi_{2}=\pi_{1}+(1-p)\pi_{3},
\\
&\pi_{3}=p\pi_{2}+(1-p)\pi_{4},
\\
&\cdots\cdots
\\
&\pi_{K-2}=p\pi_{K-3}+(1-p)\pi_{K-1},
\qquad
\pi_{K-1}=p\pi_{K-2},
\\
&\pi_{K}=p\pi_{K-1}+(1-p)\pi_{K+1},
\qquad
\pi_{K+1}=\pi_{K}+(1-p)\pi_{K+2},
\\
&\pi_{K+2}=p\pi_{K+1}+(1-p)\pi_{K+3},
\\
&\cdots\cdots
\\
&\pi_{m-1}=p\pi_{m-2}+(1-p)\pi_{m},
\qquad
\pi_{m}=p\pi_{m-1}.
\end{align*}

Let us solve for $(\pi_{i})_{i=1}^{m}$. 
First, $\pi_{m-1}=\frac{\pi_{m}}{p}$ and for any $K+1\leq i\leq m-2$, we have
\begin{equation}
\pi_{i+1}=p\pi_{i}+(1-p)\pi_{i+2},
\end{equation}
and we can solve the characteristic equation:
\begin{equation}
(1-p)x^{2}-x+p=0,
\end{equation}
to obtain $x=\frac{p}{1-p}$ or $x=1$, 
which implies that for any $K+1\leq i\leq m-2$, 
\begin{equation}
\pi_{i}=d_{1}\left(\frac{p}{1-p}\right)^{i}+d_{2},
\end{equation}
where $d_{1}$ and $d_{2}$ can be determined via the equations:
\begin{align}
&d_{1}\left(\frac{p}{1-p}\right)^{m}+d_{2}=\pi_{m},
\\
&d_{1}\left(\frac{p}{1-p}\right)^{m-1}+d_{2}=\frac{\pi_{m}}{p},
\end{align}
so that
\begin{equation}
d_{1}=\frac{p-1}{2p-1}\left(\frac{1-p}{p}\right)^{m}\pi_{m},
\qquad
d_{2}=\frac{p}{2p-1}\pi_{m}.
\end{equation}
Hence, for any $K+1\leq i\leq m-2$, we have
\begin{equation}
\pi_{i}=\frac{p-1}{2p-1}\left(\frac{1-p}{p}\right)^{m-i}\pi_{m}+\frac{p}{2p-1}\pi_{m}.
\end{equation}
Therefore,
\begin{align*}
\pi_{K}&=\pi_{K+1}-(1-p)\pi_{K+2}
\\
&=\frac{p-1}{2p-1}\left(\frac{1-p}{p}\right)^{m-K-1}\pi_{m}+\frac{p}{2p-1}\pi_{m}
-(1-p)\left(\frac{p-1}{2p-1}\left(\frac{1-p}{p}\right)^{m-K-2}\pi_{m}+\frac{p}{2p-1}\pi_{m}\right)
\\
&=\frac{p(p-1)}{2p-1}\left(\frac{1-p}{p}\right)^{m-K}\pi_{m}+\frac{p^{2}}{2p-1}\pi_{m},
\end{align*}
and
\begin{align*}
\pi_{K-1}&=\frac{\pi_{K}}{p}-\frac{1-p}{p}\pi_{K+1}
\\
&=\frac{p-1}{2p-1}\left(\frac{1-p}{p}\right)^{m-K}\pi_{m}+\frac{p}{2p-1}\pi_{m}
-\frac{p-1}{2p-1}\left(\frac{1-p}{p}\right)^{m-K}\pi_{m}-\frac{1-p}{2p-1}\pi_{m}
\\
&=\pi_{m}.
\end{align*}
Similar as before, we obtain that $\pi_{K-2}=\frac{\pi_{m}}{p}$ and for any $2\leq i\leq K-3$, 
\begin{equation}
\pi_{i}=\frac{p-1}{2p-1}\left(\frac{1-p}{p}\right)^{K-i}\pi_{m}+\frac{p}{2p-1}\pi_{m}.
\end{equation}
Moreover, we can compute that
\begin{align*}
\pi_{1}&=\pi_{2}-(1-p)\pi_{3}
\\
&=\frac{p-1}{2p-1}\left(\frac{1-p}{p}\right)^{K-2}\pi_{m}+\frac{p}{2p-1}\pi_{m}
-(1-p)\left(\frac{p-1}{2p-1}\left(\frac{1-p}{p}\right)^{K-3}\pi_{m}+\frac{p}{2p-1}\pi_{m}\right)
\\
&=(1-p)\frac{p-1}{2p-1}\left(\frac{1-p}{p}\right)^{K-2}\pi_{m}+\frac{p^{2}}{2p-1}\pi_{m}.
\end{align*}
Finally, the constraint $\sum_{i=1}^{m}\pi_{i}=1$ yields that
\begin{align*}
&(1-p)\frac{p-1}{2p-1}\left(\frac{1-p}{p}\right)^{K-2}\pi_{m}+\frac{p^{2}}{2p-1}\pi_{m}
+\sum_{i=2}^{K-1}\left(\frac{p-1}{2p-1}\left(\frac{1-p}{p}\right)^{K-i}\pi_{m}+\frac{p}{2p-1}\pi_{m}\right)
\\
&\qquad
+\frac{p(p-1)}{2p-1}\left(\frac{1-p}{p}\right)^{m-K}\pi_{m}+\frac{p^{2}}{2p-1}\pi_{m}
+\sum_{i=K+1}^{m}\left(\frac{p-1}{2p-1}\left(\frac{1-p}{p}\right)^{m-i}\pi_{m}+\frac{p}{2p-1}\pi_{m}\right)=1,
\end{align*}
which implies that
\begin{align*}
&-\frac{(1-p)^{2}}{2p-1}\left(\frac{1-p}{p}\right)^{K-2}+\frac{2p^{2}}{2p-1}
+\frac{(m-2)p}{2p-1}
+\frac{(1-p)^{2}}{(2p-1)^{2}}\left(\left(\frac{1-p}{p}\right)^{K-2}-1\right)
\\
&\qquad
+\frac{p(p-1)}{2p-1}\left(\frac{1-p}{p}\right)^{m-K}
+\frac{p(p-1)}{(2p-1)^{2}}\left(1-\left(\frac{1-p}{p}\right)^{m-K}\right)=\frac{1}{\pi_{m}},
\end{align*}
so that
\begin{align*}
&\frac{2p^{2}+(m-2)p}{2p-1}
+\frac{2(1-p)^{3}}{(2p-1)^{2}}\left(\frac{1-p}{p}\right)^{K-2}
\\
&\qquad\qquad\qquad
+\frac{2p(p-1)^{2}}{(2p-1)^{2}}\left(\frac{1-p}{p}\right)^{m-K}+\frac{p-1}{(2p-1)^{2}}
=\frac{1}{\pi_{m}},
\end{align*}
which implies that
\begin{align*}
\pi_{m}&=\Bigg(\frac{4p^{3}+2(m-3)p^{2}-(m-3)p-1}{(2p-1)^{2}}
+\frac{2p^{3}}{(2p-1)^{2}}\left(\frac{1-p}{p}\right)^{K+1}
+\frac{2p(p-1)^{2}}{(2p-1)^{2}}\left(\frac{1-p}{p}\right)^{m-K}\Bigg)^{-1}.
\end{align*}
This completes the proof.
\hfill $\Box$

%%%%%%%%%%%%%%%%%%%%%%%%%%%%%%%%%%%%%%%%%%%%%%%%%%%%%%%%%%%%%%%%%%%%%%%%%%

\subsection*{Proof of Proposition~\ref{prop:h:eta:i:j}}
We can compute that
\begin{align*}
&h^{(r)}(s;\eta_{1},\eta_{j})=\mathbb{E}_{H}\left[\left\Vert \left(I-\frac{\eta_{2}}{b}H\right)e_{1}\right\Vert^{s}\right]
\left(1_{j=2}+1_{j\neq 2}h^{(r)}(s;\eta_{2},\eta_{j})\right),
\\
&h^{(r)}(s;\eta_{K},\eta_{j})=\mathbb{E}_{H}\left[\left\Vert \left(I-\frac{\eta_{K+1}}{b}H\right)e_{1}\right\Vert^{s}\right]
\left(1_{j=K+1}+1_{j\neq K+1}h^{(r)}(s;\eta_{K+1},\eta_{j})\right),
\end{align*}
and for any $i=2,\ldots,K-1,K+1,\ldots,m$,
\begin{align*}
h^{(r)}(s;\eta_{i},\eta_{j})
&=p\mathbb{E}_{H}\left[\left\Vert \left(I-\frac{\eta_{i+1}}{b}H\right)e_{1}\right\Vert^{s}\right]
\left(1_{j=i+1}+1_{j\neq i+1}h^{(r)}(s;\eta_{i+1},\eta_{j})\right)
\\
&\qquad
+(1-p)\mathbb{E}_{H}\left[\left\Vert \left(I-\frac{\eta_{i-1}}{b}H\right)e_{1}\right\Vert^{s}\right]
\left(1_{j=i-1}+1_{j\neq i-1}h^{(r)}(s;\eta_{i-1},\eta_{j})\right).
\end{align*}

To simplify the notation, we define:
\begin{equation}
h_{ij}:=h^{(r)}(s;\eta_{i},\eta_{j}),
\qquad
a_{i}:=\mathbb{E}_{H}\left[\left\Vert \left(I-\frac{\eta_{i}}{b}H\right)e_{1}\right\Vert^{s}\right].
\end{equation}
Then, we have 
\begin{align*}
&h_{1j}=a_{2}\left(1_{j=2}+1_{j\neq 2}h_{2j}\right),
\\
&h_{Kj}=a_{K+1}
\left(1_{j=K+1}+1_{j\neq K+1}h_{(K+1)j}\right),
\end{align*}
and for any $i=2,\ldots,K-1,K+1,\ldots,m$,
\begin{align*}
h_{ij}
=pa_{i+1}\left(1_{j=i+1}+1_{j\neq i+1}h_{(i+1)j}\right)
+(1-p)a_{i-1}\left(1_{j=i-1}+1_{j\neq i-1}h_{(i-1)j}\right).
\end{align*}
Let us define the vectors
$h^{j}=[h_{1j},h_{2j},\ldots,h_{mj}]^{T}$,
$p^{j}=[p_{1j},p_{2j},\ldots,p_{mj}]^{T}$, 
where for any $i=2,\ldots,K-1,K+1,\ldots,m$
\begin{equation}
p_{ij}=pa_{i+1}1_{j=i+1}
+(1-p)a_{i-1}1_{j=i-1},
\end{equation}
and
\begin{align*}
p_{1j}=a_{2}1_{j=2},
\qquad
p_{Kj}=a_{K+1}1_{j=K+1},
\end{align*}
and the matrices $Q^{j}=(Q_{i\ell}^{j})_{1\leq i,\ell\leq m}$
such that for any $i=2,\ldots,K-1,K+1,\ldots,m$
\begin{equation}
Q_{i\ell}^{j}=pa_{i+1}1_{j\neq i+1}1_{\ell=i+1}
+(1-p)a_{i-1}1_{j\neq i-1}1_{\ell=i-1},
\end{equation}
and
\begin{align*}
Q_{1\ell}^{j}=1_{j\neq 2}1_{\ell=2},
\qquad
Q_{K\ell}^{j}=1_{j\neq K+1}1_{\ell=K+1}.
\end{align*}
Thus, we have
\begin{equation}
h^{j}=p^{j}+Q^{j}h^{j},
\end{equation}
such that
\begin{equation}
h^{j}=(I-Q^{j})^{-1}p^{j}.
\end{equation}
This completes the proof. 
\hfill $\Box$

\subsection*{Proof of Lemma~\ref{cor:two:state}}
It is easy to compute that:
\begin{equation}
\mathbb{P}(r_{1}=1)=1-p,
\qquad
\mathbb{P}(r_{1}=k)=p^{2}(1-p)^{k-2},\qquad k=2,3,\ldots,
\end{equation}
where $r_{1}$ is defined in \eqref{defn:regeneration:time}.
Conditional on $\eta_{0}=\eta_{l}$, we have
\begin{align}
&\mathbb{E}_{\eta_{0}=\eta_{l}}\left[\prod_{i=1}^{r_{1}}\mathbb{E}_{H}\left[\left\Vert \left(I-\frac{\eta_{i}}{b}H\right)e_{1}\right\Vert^{s}\right]\right]
\nonumber
\\
&=(1-p)\mathbb{E}_{H}\left[\left\Vert \left(I-\frac{\eta_{l}}{b}H\right)e_{1}\right\Vert^{s}\right]
+\sum_{k=2}^{\infty}p^{2}(1-p)^{k-2}\mathbb{E}_{H}\left[\left\Vert \left(I-\frac{\eta_{l}}{b}H\right)e_{1}\right\Vert^{s}\right]
\left(\mathbb{E}_{H}\left[\left\Vert \left(I-\frac{\eta_{u}}{b}H\right)e_{1}\right\Vert^{s}\right]\right)^{k-1}
\nonumber
\\
&=\frac{\mathbb{E}_{H}\left[\left\Vert \left(I-\frac{\eta_{l}}{b}H\right)e_{1}\right\Vert^{s}\right](1-p+(2p-1)\mathbb{E}_{H}\left[\left\Vert \left(I-\frac{\eta_{u}}{b}H\right)e_{1}\right\Vert^{s}\right])}{1-(1-p)\mathbb{E}_{H}\left[\left\Vert \left(I-\frac{\eta_{u}}{b}H\right)e_{1}\right\Vert^{s}\right]},
\end{align}
{\color{black}where we used the assumption that $(1-p)\mathbb{E}_{H}\left[\left\Vert \left(I-\frac{\eta_{u}}{b}H\right)e_{1}\right\Vert^{s}\right]<1$}
and moreover
\begin{align*}
&\mathbb{E}_{\eta_{0}=\eta_{l}}\left[\sum_{i=1}^{r_{1}}\mathbb{E}_{H}\left[\log\left\Vert \left(I-\frac{\eta_{i}}{b}H\right)e_{1}\right\Vert\right]\right]
\\
&=(1-p)\mathbb{E}_{H}\left[\log\left\Vert \left(I-\frac{\eta_{l}}{b}H\right)e_{1}\right\Vert\right]
\nonumber
\\
&\qquad
+\sum_{k=2}^{\infty}p^{2}(1-p)^{k-2}
\left(\mathbb{E}_{H}\left[\log\left\Vert \left(I-\frac{\eta_{l}}{b}H\right)e_{1}\right\Vert\right]
+(k-1)\mathbb{E}_{H}\left[\log\left\Vert \left(I-\frac{\eta_{u}}{b}H\right)e_{1}\right\Vert\right]\right)
\\
&=\mathbb{E}_{H}\left[\log\left\Vert \left(I-\frac{\eta_{l}}{b}H\right)e_{1}\right\Vert\right]
+\mathbb{E}_{H}\left[\log\left\Vert \left(I-\frac{\eta_{u}}{b}H\right)e_{1}\right\Vert\right],
\end{align*}
where we applied Lemma~\ref{lem:mean} to obtain the last equality above.

Similarly, we can compute that
\begin{align}
&\mathbb{E}_{\eta_{0}=\eta_{u}}\left[\prod_{i=1}^{r_{1}}\mathbb{E}_{H}\left[\left\Vert \left(I-\frac{\eta_{i}}{b}H\right)e_{1}\right\Vert^{s}\right]\right]
\nonumber
\\
&=\frac{\mathbb{E}_{H}\left[\left\Vert \left(I-\frac{\eta_{u}}{b}H\right)e_{1}\right\Vert^{s}\right](1-p+(2p-1)\mathbb{E}_{H}\left[\left\Vert \left(I-\frac{\eta_{l}}{b}H\right)e_{1}\right\Vert^{s}\right])}{1-(1-p)\mathbb{E}_{H}\left[\left\Vert \left(I-\frac{\eta_{l}}{b}H\right)e_{1}\right\Vert^{s}\right]},
\end{align}
{\color{black}where we used the assumption that $(1-p)\mathbb{E}_{H}\left[\left\Vert \left(I-\frac{\eta_{l}}{b}H\right)e_{1}\right\Vert^{s}\right]<1$}
and {\color{black}moreover}
\begin{align*}
\mathbb{E}_{\eta_{0}=\eta_{u}}\left[\sum_{i=1}^{r_{1}}\mathbb{E}_{H}\left[\log\left\Vert \left(I-\frac{\eta_{i}}{b}H\right)e_{1}\right\Vert\right]\right]
=\mathbb{E}_{H}\left[\log\left\Vert \left(I-\frac{\eta_{l}}{b}H\right)e_{1}\right\Vert\right]
+\mathbb{E}_{H}\left[\log\left\Vert \left(I-\frac{\eta_{u}}{b}H\right)e_{1}\right\Vert\right].
\end{align*}

Since the Markov chain exhibits a unique stationary distribution $\mathbb{P}(\eta_{0}=\eta_{\ell})=\mathbb{P}(\eta_{0}=\eta_{u})=\frac{1}{2}$,
we conclude that
\begin{align}
\mathbb{E}\left[\prod_{i=1}^{r_{1}}\mathbb{E}_{H}\left[\left\Vert \left(I-\frac{\eta_{i}}{b}H\right)e_{1}\right\Vert^{s}\right]\right]
&=
\frac{\mathbb{E}_{H}\left[\left\Vert \left(I-\frac{\eta_{l}}{b}H\right)e_{1}\right\Vert^{s}\right](1-p+(2p-1)\mathbb{E}_{H}\left[\left\Vert \left(I-\frac{\eta_{u}}{b}H\right)e_{1}\right\Vert^{s}\right])}{2(1-(1-p)\mathbb{E}_{H}\left[\left\Vert \left(I-\frac{\eta_{u}}{b}H\right)e_{1}\right\Vert^{s}\right])}
\nonumber
\\
&\qquad
+\frac{\mathbb{E}_{H}\left[\left\Vert \left(I-\frac{\eta_{u}}{b}H\right)e_{1}\right\Vert^{s}\right](1-p+(2p-1)\mathbb{E}_{H}\left[\left\Vert \left(I-\frac{\eta_{l}}{b}H\right)e_{1}\right\Vert^{s}\right])}{2(1-(1-p)\mathbb{E}_{H}\left[\left\Vert \left(I-\frac{\eta_{l}}{b}H\right)e_{1}\right\Vert^{s}\right])},
\end{align}
and
\begin{align*}
\mathbb{E}\left[\sum_{i=1}^{r_{1}}\mathbb{E}_{H}\left[\log\left\Vert \left(I-\frac{\eta_{i}}{b}H\right)e_{1}\right\Vert\right]\right]
=\mathbb{E}_{H}\left[\log\left\Vert \left(I-\frac{\eta_{l}}{b}H\right)e_{1}\right\Vert\right]
+\mathbb{E}_{H}\left[\log\left\Vert \left(I-\frac{\eta_{u}}{b}H\right)e_{1}\right\Vert\right].
\end{align*}
The proof is complete.
\hfill $\Box$

%%%%%%%%%%%%%%%%%%%%%%%%%%%%%%%%%%%%%%%%%%%%%%%%%%%%%%%%%%%%%%%

%%%%%%%%%%%%%%%%%%%%%%%%%%%%%%%%%%%%%%%%%%%%%%%%%%%%%%%%%%%%%%

\subsection*{Proof of Corollary~\ref{cor:c:r}}
Since $c^{(r)}=\tilde{h}^{(r)}(2)$, it immediately follows from Lemma~\ref{cor:two:state} that
\begin{align}
c^{(r)}&=
\frac{\mathbb{E}_{H}\left[\left\Vert \left(I-\frac{\eta_{l}}{b}H\right)e_{1}\right\Vert^{2}\right](1-p+(2p-1)\mathbb{E}_{H}\left[\left\Vert \left(I-\frac{\eta_{u}}{b}H\right)e_{1}\right\Vert^{2}\right])}{2(1-(1-p)\mathbb{E}_{H}\left[\left\Vert \left(I-\frac{\eta_{u}}{b}H\right)e_{1}\right\Vert^{2}\right])}
\nonumber
\\
&\qquad
+\frac{\mathbb{E}_{H}\left[\left\Vert \left(I-\frac{\eta_{u}}{b}H\right)e_{1}\right\Vert^{2}\right](1-p+(2p-1)\mathbb{E}_{H}\left[\left\Vert \left(I-\frac{\eta_{l}}{b}H\right)e_{1}\right\Vert^{2}\right])}{2(1-(1-p)\mathbb{E}_{H}\left[\left\Vert \left(I-\frac{\eta_{l}}{b}H\right)e_{1}\right\Vert^{2}\right])}.\label{plugging:l:u:into}
\end{align}
Moreover, we can compute that
\begin{align}
\mathbb{E}_{H}\left[\left\Vert \left(I-\frac{\eta_{l}}{b}H\right)e_{1}\right\Vert^{2}\right]
&=\mathbb{E}\left[1- \frac{2\eta_{l} \sigma^2}{b} X
+ \frac{\eta_{l}^2 \sigma^4}{b^{2}}(X^{2}+XY)\right]
\nonumber
\\
&=1-2\eta_{l}\sigma^{2}+\frac{\eta_{l}^{2}\sigma^{4}}{b}(d+b+1),\label{plugging:l}
\end{align}
where $X,Y$ are independent and $X$ is chi-square random variable
with degree of freedom $b$ and $Y$ is a chi-square random variable
with degree of freedom $(d-1)$.
Similarly, we have
\begin{equation}\label{plugging:u}
\mathbb{E}_{H}\left[\left\Vert \left(I-\frac{\eta_{u}}{b}H\right)e_{1}\right\Vert^{2}\right]
=1-2\eta_{u}\sigma^{2}+\frac{\eta_{u}^{2}\sigma^{4}}{b}(d+b+1).
\end{equation}
Finally, by plugging \eqref{plugging:l} and \eqref{plugging:u} into \eqref{plugging:l:u:into}, 
we complete the proof.
\hfill $\Box$

%%%%%%%%%%%%%%%%%%%
\subsection{Proofs of Results in Section~\ref{sec:tech:lemma}}

%%%%%%%%%%%%%%%%%%%%%%%%%%%%%%%%%%

\subsection*{Proof of Lemma~\ref{lem:alternative:expression}}
If we have i.i.d. Guassian data, i.e. $a_{i}\sim\mathcal{N}(0,\sigma^{2}I_{d})$ are Gaussian distributed for every $i$, 
then conditional on the stepsize $\eta_{k}$,
due to spherical symmetry of the isotropic Gaussian distribution, the distribution of $\frac{\|M_k x\|}{\|x\|}$ does not depend on the choice of $x\in\mathbb{R}^d \backslash \{0\}$  and is i.i.d. over $k$ with the same distribution as $\|M_{1}e_{1}\|$ where we chose $x=e_1$, 
where $e_{1}$
is the first basis vector in $\mathbb{R}^{d}$.

To see this, for any $x\in\mathbb{R}^{d}$
with $\Vert x\Vert=1$, 
we can write $x=Re_{1}$ for some
orthonormal matrix $R$, where $e_{1}$
is the first basis vector in $\mathbb{R}^{d}$.
Define $b_{i}:=R^{T}a_{i}$, here 
$a_{i}\sim\mathcal{N}(0,\sigma^{2}I_{d})$, 
and since $R$ is orthonormal, 
$b_{i}$ are also i.i.d. $\mathcal{N}(0,\sigma^{2}I_{d})$
distributed. 
Then, we can compute that
\begin{align*}
\Vert M_{k}x\Vert
=\left\Vert\left(I-\frac{\eta_{k}}{b}\sum_{i\in\Omega_{k}}a_{i}a_{i}^{T}\right)x\right\Vert
&=\left\Vert\left(RR^{T}-\frac{\eta_{k}}{b}\sum_{i\in\Omega_{k}}Rb_{i}b_{i}^{T}R^{T}\right)Re_{1}\right\Vert
\\
&=\left\Vert R\left(I-\frac{\eta_{k}}{b}\sum_{i\in\Omega_{k}}b_{i}b_{i}^{T}\right)R^{T}Re_{1}\right\Vert
\\
&=\left\Vert\left(I-\frac{\eta_{k}}{b}\sum_{i\in\Omega_{k}}b_{i}b_{i}^{T}\right)e_{1}\right\Vert,
\end{align*}
which has the same distribution as $\Vert M_{k}e_{1}\Vert$. 
By following the similar arguments as the proof of Theorem~3 
in \cite{HTP2021}, the conclusion follows.
\hfill $\Box$

%%%%%%%%%%%%%%%%%%%%%%%%%%%%%%%%%%%%%%%%%%%%%%%%%%%%%%%%
\subsection*{Proof of Lemma~\ref{lem:simplified}}
Conditional on the stepsize $\eta$, it follows from Lemma~19 in \cite{HTP2021} that for any $s\geq 0$, 
\begin{equation*}
\mathbb{E}\left[\left\Vert\left(I-\frac{\eta}{b}H\right)e_{1}\right\Vert^{s}\Big|\eta\right]
=\mathbb{E}\left[\left(\left(1-\frac{\eta\sigma^{2}}{b}X\right)^{2}
+\frac{\eta^{2}\sigma^{4}}{b^{2}}XY\right)^{s/2}\Big|\eta\right],
\end{equation*}
and
\begin{align*}
\mathbb{E}\left[\log\left\Vert\left(I-\frac{\eta}{b}H\right)e_{1}\right\Vert\Big|\eta\right]
=\frac{1}{2}
\mathbb{E}\left[\log\left(\left(1-\frac{\eta\sigma^{2}}{b}X\right)^{2}+\frac{\eta^{2}\sigma^{4}}{b^{2}}XY\right)\Big|\eta\right],
\end{align*}
where $X,Y$ are independent and $X$ is chi-square random variable
with degree of freedom $b$ and $Y$ is a chi-square random variable
with degree of freedom $(d-1)$.
Hence, the conclusion follows.
\hfill $\Box$

%%%%%%%%%%%%%%%%%%%%%%%%%%%%%%%%%%%%%%%%%%%%%%%%%%%%%%

\subsection*{Proof of Lemma~\ref{lem:alternative:expression:cyclic}}
We follow the similar arguments
as the proof of Theorem~3  in \cite{HTP2021} and the key observation
is that the distribution
of $\left\Vert M_{1}^{(m)}\right\Vert/\Vert x\Vert=\Vert M_{m}M_{m-1}\cdots M_{1}x\Vert/\Vert x\Vert$
is the same for every $x\in\mathbb{R}^{d}\backslash\{0\}$. 
For any $x\in\mathbb{R}^{d}$
with $\Vert x\Vert=1$, 
we can write $x=Re_{1}$ for some
orthonormal matrix $R$, where $e_{1}$
is the first basis vector in $\mathbb{R}^{d}$.
Define $b_{i}:=R^{T}a_{i}$, here 
$a_{i}\sim\mathcal{N}(0,\sigma^{2}I_{d})$, 
and since $R$ is orthonormal, 
$b_{i}$ are also i.i.d. $\mathcal{N}(0,\sigma^{2}I_{d})$
distributed. 
Then, we can compute that
\begin{align*}
\left\Vert M_{1}^{(m)}\right\Vert
&=\left\Vert M_{m}M_{m-1}\cdots M_{1}x\right\Vert
\\
&=\left\Vert\left(I-\frac{\eta_{m}}{b}\sum_{i\in\Omega_{m}}a_{i}a_{i}^{T}\right)\left(I-\frac{\eta_{m-1}}{b}\sum_{i\in\Omega_{m-1}}a_{i}a_{i}^{T}\right)
\cdots
\left(I-\frac{\eta_{1}}{b}\sum_{i\in\Omega_{1}}a_{i}a_{i}^{T}\right)x\right\Vert
\\
&=\Bigg\Vert R\left(I-\frac{\eta_{m}}{b}\sum_{i\in\Omega_{m}}b_{i}b_{i}^{T}\right)R^{T}R\left(I-\frac{\eta_{m-1}}{b}\sum_{i\in\Omega_{m-1}}b_{i}b_{i}^{T}\right)R^{T}
\cdots
R\left(I-\frac{\eta_{1}}{b}\sum_{i\in\Omega_{1}}b_{i}b_{i}^{T}\right)R^{T}Re_{1}\Bigg\Vert
\\
&=\left\Vert R\left(I-\frac{\eta_{m}}{b}\sum_{i\in\Omega_{m}}b_{i}b_{i}^{T}\right)\left(I-\frac{\eta_{m-1}}{b}\sum_{i\in\Omega_{m-1}}b_{i}b_{i}^{T}\right)
\cdots
\left(I-\frac{\eta_{1}}{b}\sum_{i\in\Omega_{1}}b_{i}b_{i}^{T}\right)e_{1}\right\Vert
\\
&=\left\Vert\left(I-\frac{\eta_{m}}{b}\sum_{i\in\Omega_{m}}b_{i}b_{i}^{T}\right)\left(I-\frac{\eta_{m-1}}{b}\sum_{i\in\Omega_{m-1}}b_{i}b_{i}^{T}\right)
\cdots
\left(I-\frac{\eta_{1}}{b}\sum_{i\in\Omega_{1}}b_{i}b_{i}^{T}\right)e_{1}\right\Vert,
\end{align*}
which has the same distribution as $\Vert M_{m}M_{m-1}\cdots M_{1}x\Vert/\Vert e_{1}\Vert$. 
By following the similar arguments
as the proof of Theorem~3 in \cite{HTP2021}, we obtain:
\begin{equation}
h^{(m)}(s)
=\mathbb{E}\left[\left\Vert\left(I-\frac{\eta_{m}}{b}H_{m}\right)
\left(I-\frac{\eta_{m-1}}{b}H_{m-1}\right)\cdots
\left(I-\frac{\eta_{1}}{b}H_{1}\right)e_{1}\right\Vert^{s}\right].
\end{equation}
By tower property and the fact
that the distribution
of $\Vert M_{m}M_{m-1}\cdots M_{1}x\Vert/\Vert x\Vert$
is the same for every $x\in\mathbb{R}^{d}\backslash\{0\}$
and $(\eta_{i},H_{i})$ are i.i.d., we have
\begin{align*}
h^{(m)}(s)
&=\mathbb{E}\left[\mathbb{E}\left[\left\Vert\left(I-\frac{\eta_{m}}{b}H_{m}\right)
\left(I-\frac{\eta_{m-1}}{b}H_{m-1}\right)\cdots
\left(I-\frac{\eta_{1}}{b}H_{1}\right)e_{1}\right\Vert^{s}\Big|\eta_{1},H_{1}\right]\right]
\\
&=\mathbb{E}\left[\mathbb{E}\left[\left\Vert\left(I-\frac{\eta_{m}}{b}H_{m}\right)
\cdots
\left(I-\frac{\eta_{2}}{b}H_{2}\right)e_{1}\right\Vert^{s}\Big|\eta_{1},H_{1}\right]
\left\Vert\left(I-\frac{\eta_{1}}{b}H_{1}\right)e_{1}\right\Vert^{s}\right]
\\
&=\mathbb{E}\left[\left\Vert\left(I-\frac{\eta_{m}}{b}H_{m}\right)
\cdots
\left(I-\frac{\eta_{2}}{b}H_{2}\right)e_{1}\right\Vert^{s}\right]
\mathbb{E}\left[\left\Vert\left(I-\frac{\eta_{1}}{b}H_{1}\right)e_{1}\right\Vert^{s}\right],
\end{align*}
and therefore inductively we get
\begin{align*}
h^{(m)}(s)
=\mathbb{E}\left[\left\Vert\left(I-\frac{\eta_{m}}{b}H_{m}\right)e_{1}\right\Vert^{s}\right]
\mathbb{E}\left[\left\Vert\left(I-\frac{\eta_{m-1}}{b}H_{m-1}\right)e_{1}\right\Vert^{s}\right]
\cdots
\mathbb{E}\left[\left\Vert\left(I-\frac{\eta_{1}}{b}H_{1}\right)e_{1}\right\Vert^{s}\right].
\end{align*}
Hence, we conclude that
\begin{equation}
\left(h^{(m)}(s)\right)^{1/m}
=\tilde{h}^{(m)}(s),
\end{equation}
where
\begin{equation}
\tilde{h}^{(m)}(s):=\left(\prod_{i=1}^{m}\mathbb{E}\left[\left\Vert\left(I-\frac{\eta_{i}}{b}H\right)e_{1}\right\Vert^{s}\right]\right)^{1/m}.
\end{equation}
Similarly, we can derive that
\begin{equation}
\rho^{(m)}=\tilde{\rho}^{(m)},
\end{equation}
where
\begin{equation}
\tilde{\rho}^{(m)}:=\sum_{i=1}^{m}\mathbb{E}\left[\log\left\Vert\left(I-\frac{\eta_{i}}{b}H\right)e_{1}\right\Vert\right].    
\end{equation}
The proof is complete.
\hfill $\Box$

%%%%%%%%%%%%%%%%%%%%%%%%%%%%%%%%%%%%%%%%%%%%%%%%%%%%%%%%%%%%%%%
\subsection*{Proof of Lemma~\ref{lem:simplified:cyclic}}
It follows from Lemma~19 in \cite{HTP2021} that
\begin{align}
&\mathbb{E}\left[\left\Vert\left(I-\frac{\eta_{i}}{b}H\right)e_{1}\right\Vert^{s}\right]
=\mathbb{E}\left[\left(\left(1-\frac{\eta_{i}\sigma^{2}}{b}X\right)^{2}
+\frac{\eta_{i}^{2}\sigma^{4}}{b^{2}}XY\right)^{s/2}\right],
\\
&\mathbb{E}\left[\log\left\Vert\left(I-\frac{\eta_{i}}{b}H\right)e_{1}\right\Vert\right]
=\frac{1}{2}\mathbb{E}\left[\log\left(\left(1-\frac{\eta_{i}\sigma^{2}}{b}X\right)^{2}
+\frac{\eta_{i}^{2}\sigma^{4}}{b^{2}}XY\right)\right],
\end{align}
where $X,Y$ are independent and $X$ is chi-square random variable
with degree of freedom $b$ and $Y$ is a chi-square random variable
with degree of freedom $(d-1)$. The conclusion follows.
\hfill $\Box$

%%%%%%%%%%%%%%%%%%%%%%%%%%%%%%%%%%%%%%%%%%%%%%%%%%%%%%%%%%%%%%%%%%%%%%%%%%
\subsection*{Proof of Lemma~\ref{lem:alternative:expression:Markov}}
We follow the similar arguments
as the proof of Theorem~3 in \cite{HTP2021} and the key observation
is that conditional on $(\eta_{i})_{i=1}^{r_{1}}$
the distribution
of $\left\Vert M_{1}^{(r)}x\right\Vert/\Vert x\Vert$
is the same for every $x\in\mathbb{R}^{d}\backslash\{0\}$, 
where $r_{1}$ is defined in \eqref{defn:regeneration:time}.
By tower property, we have
\begin{align*}
h^{(r)}(s)
&=\mathbb{E}\left[\mathbb{E}\left[\left\Vert\left(I-\frac{\eta_{r_{1}}}{b}H_{r_{1}}\right)
\left(I-\frac{\eta_{r_{1}-1}}{b}H_{r_{1}-1}\right)\cdots
\left(I-\frac{\eta_{1}}{b}H_{1}\right)e_{1}\right\Vert^{s}\Big|(\eta_{i})_{i=1}^{r_{1}}\right]\right]
\\
&=\mathbb{E}\Bigg[\mathbb{E}_{H_{r_{1}}}\left[\left\Vert\left(I-\frac{\eta_{r_{1}}}{b}H_{r_{1}}\right)e_{1}\right\Vert^{s}\right]
\mathbb{E}_{H_{r_{1}-1}}\left[\left\Vert\left(I-\frac{\eta_{r_{1}-1}}{b}H_{r_{1}-1}\right)e_{1}\right\Vert^{s}\right]
\mathbb{E}_{H_{1}}\left[\left\Vert\left(I-\frac{\eta_{1}}{b}H_{1}\right)e_{1}\right\Vert^{s}\right]\Bigg],
\end{align*}
and therefore inductively we conclude that
\begin{equation}
h^{(r)}(s)
=\tilde{h}^{(r)}(s):=\mathbb{E}\left[\prod_{i=1}^{r_{1}}\mathbb{E}_{H}\left[\left\Vert \left(I-\frac{\eta_{i}}{b}H\right)e_{1}\right\Vert^{s}\right]\right].
\end{equation}
Similarly, we can derive that
$\rho=\rho^{(r)}$,
where
\begin{equation}
\rho^{(r)}:=\mathbb{E}\left[\sum_{i=1}^{r_{1}}\mathbb{E}_{H}\left[\log\left\Vert \left(I-\frac{\eta_{i}}{b}H\right)e_{1}\right\Vert\right]\right].
\end{equation}
The proof is complete.
\hfill $\Box$

%%%%%%%%%%%%%%%%%%%%%%%%%%%%%%%%%%%%%%%%%%%%%%%%%%%%%
\subsection*{Proof of Lemma~\ref{lem:simplified:Markov:regeneration}}
It follows from Lemma~19 in \cite{HTP2021} that conditional on $\eta_{i}$,
\begin{align}
&\mathbb{E}_{H}\left[\left\Vert\left(I-\frac{\eta_{i}}{b}H\right)e_{1}\right\Vert^{s}\right]
=\mathbb{E}_{X,Y}\left[\left(\left(1-\frac{\eta_{i}\sigma^{2}}{b}X\right)^{2}
+\frac{\eta_{i}^{2}\sigma^{4}}{b^{2}}XY\right)^{s/2}\right],
\\
&\mathbb{E}_{H}\left[\log\left\Vert\left(I-\frac{\eta_{i}}{b}H\right)e_{1}\right\Vert\right]
=\frac{1}{2}\mathbb{E}_{X,Y}\left[\log\left(\left(1-\frac{\eta_{i}\sigma^{2}}{b}X\right)^{2}
+\frac{\eta_{i}^{2}\sigma^{4}}{b^{2}}XY\right)\right],
\end{align}
where $X,Y$ are independent and $X$ is chi-square random variable
with degree of freedom $b$ and $Y$ is a chi-square random variable
with degree of freedom $(d-1)$. The conclusion follows.
\hfill $\Box$

%%%%%%%%%%%%%%%%%%%%%%%%%%%%%%%%%%%%%%%%%%%%%%%%%%%%%%%%%%%%%%%%%%%%%%%%%%%%%

%%%%%%%%%%%%%%%%%%%%%%%%%%%%%%%%%%%%%%%%%%%%%%%%%%%%%
\section{Supporting Lemmas}

In this section, we provide a few supporting technical lemmas that are used in the proofs
of the main results in the paper.

\begin{lemma}[Lemma~22 in \cite{HTP2021}]\label{lemma-cvx-in-step} 
For any given positive semi-definite symmetric matrix $H$ fixed, the function $F_H:[0,\infty) \to \mathbb{R}$ defined as
$$ F_H(a) :=\left\| \left(I - a H\right)e_{1}\right\|^s $$
is convex in $a\geq 0$ for any $s\geq 1$.
\end{lemma}

\begin{lemma}[Lemma~23 in \cite{HTP2021}]\label{lem:moment:ineq}
(i) Given $0<p\leq 1$, for any $x,y\geq 0$,
\begin{equation}
(x+y)^{p}\leq x^{p}+y^{p}.
\end{equation}

(ii) Given $p>1$, for any $x,y\geq 0$,
and any $\epsilon>0$, 
\begin{equation}
(x+y)^{p}\leq(1+\epsilon)x^{p}+\frac{(1+\epsilon)^{\frac{p}{p-1}}-(1+\epsilon)}{\left((1+\epsilon)^{\frac{1}{p-1}}-1\right)^{p}}y^{p}.
\end{equation}
\end{lemma}

\begin{lemma}\label{lem:mean}
For any $a>0$, and $k\in\mathbb{N}$, 
\begin{equation*}
\sum_{i=1}^{k}ia^{i}
=\frac{ka^{k+2}-(k+1)a^{k+1}+a}{(a-1)^{2}}.
\end{equation*}
In particular, for any $0<a<1$, 
\begin{equation*}
\sum_{i=1}^{\infty}ia^{i}
=\frac{a}{(a-1)^{2}}.
\end{equation*}
\end{lemma}

\subsection*{Proof of Lemma~\ref{lem:mean}}
We can compute that
\begin{equation*}
\sum_{i=1}^{k}ia^{i}
=a\sum_{i=1}^{k}ia^{i-1}
=a\frac{d}{da}\sum_{i=1}^{k}a^{i}
=a\frac{d}{da}\frac{a^{k+1}-a}{a-1}=\frac{ka^{k+2}-(k+1)a^{k+1}+a}{(a-1)^{2}}.
\end{equation*}
The proof is complete.
\hfill $\Box$

\section{Additional Results}\label{appendix-non-quadratic-results}
\mgrev{
In this section, our purpose is to extend our analysis beyond linear regression, where we will assume that component functions $f_i(x) = f(x,z_i)$ arising in the empirical risk minimization problem \eqref{eq-emp-risk} are twice continuously differentiable, and that $F(x)$ is bounded below so that a minimizer $x_*$ of $F(x)$ exists. In this case, by Taylor's formula, we can write
$$\nabla f_i(x) = \big( \bar{H}_{i}(x_k)\big) (x_k-x_*) + \nabla f_i(x_*) \quad \text{where} \quad  \bar{H}_{i}(x) := \int_{t=0}^{1} \nabla^2 f_i\left(x^*+t(x-x^*)\right)dt$$
%where 
%$$ H_{i,k}(x) := \int_{t=0}^{1} \nabla^2 f_i(x^*+t(x-x^*))dt$$
is an averaged Hessian of the function $f_i$. We then introduce the following \emph{stochastic estimate of the averaged Hessian of $F$}, defined analogously to the stochastic gradient, according to the formula
$$ H_{k+1}(x_k): = \sum_{i\in \Omega_k} \bar{H}_{i}(x_k).$$
With this notation, SGD updates are equivalent to 
\begin{equation} 
x_{k+1} -x_* = \left( M_{k+1}(x_k)\right) (x_k-x_*)  + \tilde{q}_{k+1}\,, \quad  M_{k+1}(x_k):= I - \frac{\eta_{k+1}}{b} H_{k+1}(x_k),
\label{eq-stoc-grad-3}
\end{equation}  
with $\tilde{q}_k := \frac{-\eta_{k}}{b} \sum_{i\in \Omega_{k} } \nabla f_i(x_*)$, $\Omega_k := \{b(k-1)+1, b(k-1)+2, \dots, bk\}$ and $|\Omega_k| = b$. Here, the distribution of the stochastic Hessian estimate $H_{k+1}(x_k)$ depends on the iterate $x_k$; therefore the update \eqref{eq-stoc-grad-3} can be thought as a generalization of the update rule \eqref{eq-stoc-grad-2} that arises for linear regression (where the Hessian's distribution did not depend on $x_k$)}. 

{\color{black}We first consider the case that the stepsizes are cyclic with a cycle length $m$, lying on a grid $(c_1, c_2, \dots,c_K)$. We consider the products}

%We next assume that the functions $f_i$ are strongly convex outside a compact set so that we can control the growth of the products $M_{k+1}(x_k)$.
% \begin{itemize}
% \item [\textbf{(A4)}]
%  We assume that $f_i:\mathbb{R}^d\to \mathbb{R}$ is twice continuously differentiable on $\mathbb{R}^d$ and $\mu$-strongly convex outside a compact set $\mathcal{C}$ for every $i=1,2,\dots,n$, i.e. its Hessian satisfies
% $\nabla^2 F(x)\succeq \mu I$ for all $x\not\in\mathcal{C},$ for some constant $\mu>0$ where $\succeq$ denotes the Loewner ordering between symmetric matrices. 
% \end{itemize}
% For example, if a $C^2$ loss function $\ell(x)$ admits a lower bounded Hessian i.e. then the regularized loss function $F(x)=\ell(x)+ \frac{\lambda}{2}\|x\|^2$ satisfies this property for $\lambda$ large enough. Also if $\textbf{(A4)}$ holds clearly $F(x)$ is bounded below, and a minimizer $x_*$ of $F(x)$ exists. %We also introduce %large enough. %Clearly, any quadratic function with a positive definite Hessian also belongs to this class. Next we introduce
\mgrev{
\begin{equation} 
\overline{\sigma}^{(m)}:= \prod_{j=1}^m \sup_{z\in\mathbb{R}^{d}}\| M_{j}(z)\|, \quad \underline{\sigma}^{(m)}:= \prod_{j=1}^m \left( \liminf_{\|z\|\to\infty} \sigma_{\min} \big(M_{j}(z)\big) \right),
\label{def-sigma-min-max}
\end{equation}
}
\mgrev{which are random quantities (as $M_{k+1}(z)$ is random when $z$ is fixed, due to the randomness in the data) that roughly speaking measure the maximal and minimal growth of $M_{k+1}(x_k)$ in a cycle of length $m$ where $\sigma_{\min}(\cdot)$ denotes the smallest singular value. The following result shows that the distributions can be heavy-tailed at stationarity with cyclic stepsizes provided that the minimal growth is large enough, i.e. if $\mathbb{P}(\underline{\sigma}^{(m)}>1)>0$.}% under some condition%nd $\underline{\sigma}^{(m)}$ can be larger than 1, with a positive
%e consider the empirical risk minimization problem 
 %\eqref{eq-emp-risk} as
%\begin{equation}\label{f:eqn} 
%F(x) = \frac{1}{n}\sum\nolimits_{i=1}^n f_i(x),
%\end{equation}
\mgrev{
\begin{proposition}
\label{prop:ht_strcvx} Let batch-size $b$ be given and fixed. Consider the SGD recursion with cyclic stepsize of period $m$ when $f_i$ are twice continuously differentiable and lower bounded for every $i=1,2,\dots,m$. Assume %$\textbf{(A4)}$ holdsand 
$\mathbb{E}(\log \overline{\sigma}^{(m)})<0$, $\mathbb{E} (\overline{\sigma}^{(m)})<\infty$ and $\mathbb{P}(\underline{\sigma}^{(m)}>1)>0$ where $\underline{\sigma}^{(m)}$ and $\overline{\sigma}^{(m)}$ are defined according to \eqref{def-sigma-min-max}. Then, there exists positive constants $\underline{\alpha},\overline{\alpha}$ such that the tail-index $\alpha$ lies in the interval $[\underline{\alpha},\overline{\alpha}]$, i.e. for every $\delta>0$, $\limsup_{t \to \infty} t^{\underline{\alpha}+\delta} \mathbb{P}\left(\|x^{(\infty)}\|>t\right)>0$, 
and\footnote{We use the convention that $\infty>0$.}
$\limsup_{t \to \infty} t^{\overline{\alpha}-\delta} \mathbb{P}\left(\|x_\infty\|>t\right) < \infty$ where $x_\infty$ is the stationary distribution of the SGD recursion with cyclic stepsize of period $m$. Furthermore, we have $\mathbb{E}\big[(\overline{\sigma}^{(m)})^{\overline{\alpha}}\big] = 1$ and $\mathbb{E}\big[(\underline{\sigma}^{(m)})^{\underline{\alpha}}\big] = 1$. 
%Assume that the loss function $F(x) = \frac{1}{n}\sum\nolimits_{i=1}^n f_i(x)$ that arise in the empirical risk optimization problem \eqref{eq-emp-risk} is strongly convex outside a compact set $\mathcal{C}$ for some $\mu>0$. Consider the SGD iterates with a fixed batch-size $b$. Let 
%$\ell:\mathbb{R}^d\to\mathbb{R}$ is a $\mu$-strongly convex function outside a compact set $\mathcal{C}$ for some $\mu>0$.
\end{proposition}
}
\mgrev{
\begin{proof} If we introduce $z_k = x_k - x_*$, then
from \eqref{eq-stoc-grad-3},
$$ z_{k+1} = \Phi_{k+1}(z_k) \quad \text{where} \quad \Phi_{k+1}(z_k):=\left(M_{k+1}(z_k+x_*)\right)z_k  + \tilde{q}_{k+1}.$$
In particular, the map $\Phi_{k+1}$ admits a linear growth and Lipschitz behavior satisfying
\begin{equation} \underline{s}_{k+1}\|z\| \leq \| \Phi_{k+1}(z) - \Phi_{k+1}(0)\| =  \left\|\left(M_{k+1}(z+x_*)\right)z \right\| \leq \overline{s}_{k+1} \|z\|, 
\label{ineq-lower-upper-bound-for-growth}
\end{equation}
where the first inequality holds for $\|z\|$ large enough, whereas the second inequality holds for every $z$ and
$$\underline{s}_{k+1}:=\liminf_{\|z\|\to\infty} \sigma_{\min} \big(M_{k+1}(z)\big) \quad \text{and} \quad \overline{s}_{k+1} = \sup_{z\in\mathbb{R}^{d}}\| M_{k+1}(z)\|.
$$ 
Then, we follow a similar approach to the proof of Theorem \ref{thm:cyclic} and introduce
$$z_{(k+1)m} = \mathcal{F}_{k+1}(z_{km})\quad \text{where} \quad \mathcal{F}_{k+1}(z_{km})  = \Phi_{(k+1)m} \circ \Phi_{(k+1)m-1} \circ \cdots \circ \Phi_{km+1}(z_{km})$$
is the composition of consecutive $m$ iterations. Then, the composition $\mathcal{F}_{k+1}$ will also be Lipschitz satisfying
$$ \underline{\sigma}^{(m)} \|z\| \leq \| \mathcal{F}_{k+1}(z) - \mathcal{F}_{k+1}(0)\| \leq \overline{\sigma}^{(m)} \|z\|, $$
for $\|z\|$ large enough, 
\mgrev{and the second inequality will be satisfied for every $z$}. Or equivalently, there exists a non-negative random variable $y_{k+1}$ (that depends on the sampled data points at steps $km$ to $(k+1)m$) such that for every $z$ we have
$$ 
\underline{\sigma}^{(m)} \|z\| - y_{k+1} \leq \| \mathcal{F}_{k+1}(z) - \mathcal{F}_{k+1}(0)\| \leq \overline{\sigma}^{(m)} \|z\|. 
$$
Using this inequality, the result follows from \cite[Thm. 1]{hodgkinson2020multiplicative}.
\end{proof}
}
\mgrev{
\begin{remark} Consider the smoothed Lasso loss with $f_i(x) = \frac{1}{2}(a_i^T x - y_i)^2 + \lambda \mbox{pen}(x)$ where the function $x\mapsto\mbox{pen}(x)$ is a smoothed version of the $\ell_1$ loss and $\lambda>0$ is the penalty parameter. We take $\mbox{pen}(x)=\sqrt{\|x\|^2 + 1}$ here, but many other versions are proposed in the literature (see e.g. \cite{Haselimashhadi_2018}). Then, by straightforward calculations it follows that the Hessian matrix $\nabla^2 \mbox{pen}(x)$ is uniformly bounded and satisfies
$-\frac{c_1}{R} I \preceq \nabla^2 \mbox{pen}(x) \preceq \frac{c_1}{R} I$ for a positive constant $c_1$ whenever $\|x\|\geq R$. Under similar assumptions to \textbf{(A1)} and \textbf{(A2)} on the data, it can be checked that when the stepsizes $(\eta_1, \eta_2,\ldots,\eta_m)$ are small enough, the assumptions behind Propositions ~\ref{prop:ht_strcvx} and \ref{prop:ht_strcvx:r} will hold. %Here, the loss $F(x)$ is strongly convex outside a compact set, and a minimizer $x_*$ to $F(x)$ also exists.   
\end{remark}
}
{\color{black}Next, we assume as in \eqref{state:space:m} that the stepsizes follow a Markov chain with the finite state space
\begin{equation}
\{\eta_{1},\eta_{2},\ldots,\eta_{m},\eta_{m+1}\}
=\{c_{1},c_{2},\ldots,c_{K-1},c_{K},c_{K-1},\ldots,c_{2},c_{1}\},
\end{equation}
and let $r_{1}$ be the regeneration time
such that $r_{1}=\inf\{j>0:\eta_{j}=\eta_{0}\}$.
Similar to \eqref{def-sigma-min-max}, we define the products:
\begin{equation}\label{def-sigma-min-max-r}
\overline{\sigma}^{(r)}:= \prod_{j=1}^{r_{1}} \sup_{z\in\mathbb{R}^d}\| M_{j}(z)\|, \quad \underline{\sigma}^{(r)}:= \prod_{j=1}^{r_{1}} \left( \liminf_{\|z\|\to\infty} \sigma_{\min} \big(M_{j}(z)\big) \right).
\end{equation}
By using the similar argument as in the proof of 
Proposition~\ref{prop:ht_strcvx}, we have the following
analogue of Proposition~\ref{prop:ht_strcvx} for the Markovian stepsizes.}

\begin{proposition}\label{prop:ht_strcvx:r}
{\color{black}Let batch-size $b$ be given and fixed. Consider the SGD recursion with Markovian stepsizes with finite state space \eqref{state:space:m} when $f_i$ are twice continuously differentiable and lower bounded for every $i=1,2,\dots,m$. Assume %$\textbf{(A4)}$ holdsand 
$\mathbb{E}(\log \overline{\sigma}^{(r)})<0$, $\mathbb{E} (\overline{\sigma}^{(r)})<\infty$ and $\mathbb{P}(\underline{\sigma}^{(r)}>1)>0$ where $\underline{\sigma}^{(r)}$ and $\overline{\sigma}^{(r)}$ are defined according to \eqref{def-sigma-min-max-r}. Then, there exists positive constants $\underline{\alpha},\overline{\alpha}$ such that the tail-index $\alpha$ lies in the interval $[\underline{\alpha},\overline{\alpha}]$, i.e. for every $\delta>0$, $\limsup_{t \to \infty} t^{\underline{\alpha}+\delta} \mathbb{P}\left(\|x^{(\infty)}\|>t\right)>0$, 
and
$\limsup_{t \to \infty} t^{\overline{\alpha}-\delta} \mathbb{P}\left(\|x_\infty\|>t\right) < \infty$ where $x_\infty$ is the stationary distribution of the SGD recursion with Markovian stepsizes. Furthermore, we have $\mathbb{E}\big[(\overline{\sigma}^{(r)})^{\overline{\alpha}}\big] = 1$ and $\mathbb{E}\big[(\underline{\sigma}^{(r)})^{\underline{\alpha}}\big] = 1$.} 
\end{proposition}

\end{document}